%% file: main.tex
\pdfoutput=1

\documentclass{article}

\usepackage{microtype}
\usepackage{graphicx}
\usepackage{subcaption}
\usepackage{booktabs} %

\usepackage{hyperref}

\usepackage[accepted]{icml2026}

\usepackage{amsmath}
\usepackage{amssymb}
\usepackage{mathtools}
\usepackage{amsthm}
\usepackage{booktabs}
\usepackage[]{multirow}
\usepackage[table]{xcolor}
\usepackage{graphicx}
\usepackage{bm}
\usepackage{pgfplots}
\usepackage{tikz}
\usepgfplotslibrary{groupplots}
\usepgfplotslibrary{fillbetween}
\usetikzlibrary{decorations.pathreplacing}
\usetikzlibrary{arrows.meta}
\usetikzlibrary{patterns}

\usepackage{glossaries} %
\usepackage{xcolor}

\usepackage{algorithm}
\usepackage{algpseudocode}

\definecolor{Blue1}{RGB}{9, 0, 255}
\definecolor{Red1}{RGB}{204, 0, 0}
\definecolor{Green1}{RGB}{0, 92, 78}
\definecolor{DarkGreen}{RGB}{85,168,104}
\definecolor{DarkYellow}{RGB}{240,180,0}
\definecolor{DarkBlue}{RGB}{108,142,191}
\definecolor{DarkBlue2}{RGB}{51,153,255}
\definecolor{DarkBlue}{RGB}{30,64,175}
\definecolor{MediumBlue}{RGB}{59,130,246}
\definecolor{LightBlue}{RGB}{158,200,252}
\definecolor{DarkGreen}{RGB}{22,101,52}
\definecolor{DarkerGreen}{RGB}{0, 0.5, 0}
\definecolor{MediumGreen}{RGB}{34,197,94}
\definecolor{LightGreen}{RGB}{154,234,189}
\definecolor{DarkOrange}{RGB}{154,52,18}
\definecolor{MediumOrange}{RGB}{249,115,22}
\definecolor{LightOrange}{RGB}{255,213,172}
\definecolor{DarkPurple}{RGB}{91,33,182}
\definecolor{MediumPurple}{RGB}{168,85,247}
\definecolor{LightPurple}{RGB}{237,233,254}
\definecolor{DarkYellow}{RGB}{234,179,8}
\definecolor{MediumYellow}{RGB}{255,222,100}
\definecolor{LightYellow}{RGB}{246,228,152}
\definecolor{DarkGrey}{RGB}{65,65,65}       %
\definecolor{MediumGrey}{RGB}{128,128,128}  %
\definecolor{LightGrey}{RGB}{200,200,200}
\definecolor{m}{RGB}{230, 230, 230} %
\definecolor{nice_red}{rgb}{0.8, 0, 0}
\definecolor{icmlblue}{RGB}{31, 119, 180}
\definecolor{icmlorange}{RGB}{255, 127, 14}
\definecolor{icmlred}{HTML}{D62728}

\usepackage{ifthen}

\makeatletter
\newcommand{\midrowcolor}[1]{\global\let\CT@do@color\CT@@do@color\gdef\CT@row@color{\CT@color{#1}}\ignorespaces}
\makeatother

\newboolean{shorttable}
\setboolean{shorttable}{true}

\usepackage[capitalize,noabbrev]{cleveref}

\theoremstyle{plain}

\theoremstyle{definition}

\theoremstyle{remark}

\usepackage[textsize=tiny]{todonotes}

\makeglossaries
\makeatletter
\renewcommand*{\@glotype@main@in}{gls.tex}
\makeatother

\newacronym{dl}{DL}{Deep learning}
\newacronym{ilp}{ILP}{Integer linear program}
\newacronym{ml}{ML}{Machine learning}
\newacronym{mlp}{MLP}{Multi-layer perceptron}
\newacronym{llm}{LLM}{Large language model}
\newacronym{mse}{MSE}{Mean squared error}

\newacronym{nlp}{NLP}{Natural language processing}
\newacronym{pp}{p.p.}{Percentage points}
\newacronym{kl}{KL}{Kullback-Leibler divergence}
\newacronym{tpe}{TPE}{Tree-structured parzen estimation}

\newacronym{wiki}{WikiText2}{WikiText2}
\newacronym{c4}{C4}{Colossal clean crawled corpus}
\newacronym{ptb}{PTB}{Penn treebank}

\newacronym{svd}{SVD}{Singular value decomposition}
\newacronym{ours}{KFAC-SVD}{Kronecker-factored approximate curvature singular value decomposition}
\newacronym{oursearch}{LEMS}{Layer-wise error model search}
\newacronym{kfac}{KFAC}{Kronecker-factored approximate curvature}

\newacronym{comcat}{ComCat}{Compression and customization of attention-based vision models}
\newacronym{fastlrd}{FastLRD}{Fast low rank decomposition}
\newacronym{flarsvd}{FLAR-SVD}{Fast and latency-aware SVD}
\newacronym{mrcs}{MRCS}{Mixed rank compression strategy}
\newacronym{ars}{ARS}{Adaptive rank selection}
\newacronym{atp}{ATP}{A theoretical perspective (on sparsity)}
\newacronym{mdp}{MDP}{Multidimensional vision model pruning with latency constraint}

\newacronym{asvd}{ASVD}{Activation-aware singular value decomposition}
\newacronym{dobisvd}{DOBI-SVD}{Differentiable SVD for LLM compression}
\newacronym{fwsvd}{FWSVD}{Fisher weighted SVD}
\newacronym{gfwsvd}{GFWSVD}{Generalized Fisher weighted SVD}
\newacronym{svdllm}{SVD-LLM}{SVD for LLM compression}
\newacronym{svdllmv2}{SVD-LLMv2}{Optimizing SVD for LLM compression}
\newacronym{bitstack}{BitStack}{BitStack: Any-Size Compression of LLMs in Variable Memory Environments}
\newacronym{basissharing}{BasisSharing}{Basis Sharing: Cross-Layer Parameter Sharing for LLM Compression}

\newacronym{llama3}{Llama 3}{Llama 3}
\newacronym{llama2}{Llama 2}{Llama 2}
\newacronym{llama}{Llama}{Llama}
\newacronym{mistral}{Mistral}{Mistral}
\newacronym{qwen3}{Qwen3}{Qwen3}
\newacronym{vlm}{VLM}{Vision-language models}
\newacronym{moe}{MoE}{Mixture-of-experts}

\icmltitlerunning{Advancing SVD-based LLM Compression via Layer-Wise Error Model Search}

\begin{document}

\twocolumn[
  \icmltitle{
  Advancing SVD-based LLM Compression via Layer-Wise Error Model Search
}

  \icmlsetsymbol{equal}{*}

  \begin{icmlauthorlist}
    \icmlauthor{Moritz Thoma}{tum,bmw}
    \icmlauthor{Maximilian Groezinger}{bmw,equal}
    \icmlauthor{Maximilian Forstenhäusler}{bmw,uk,equal}
    \icmlauthor{Emad Aghajanzadeh}{tum,bmw}
    \icmlauthor{Manoj Rohit Vemparala}{bmw}
    \icmlauthor{Christos Anagnostopoulos}{uk}
    \icmlauthor{Pierpaolo Mori}{bmw}
    \icmlauthor{Nael Fasfous}{bmw}
    \icmlauthor{Alexander Frickenstein}{bmw}
    \icmlauthor{Daniel Mueller-Gritschneder}{vienna}
    \icmlauthor{Ulf Schlichtmann}{tum}
  \end{icmlauthorlist}

  \icmlaffiliation{tum}{Technical University of Munich, Germany}
  \icmlaffiliation{bmw}{BMW Group, Germany}
  \icmlaffiliation{uk}{University of Glasgow, United Kingdom}
  \icmlaffiliation{vienna}{TU Wien, Austria}

  \icmlcorrespondingauthor{Moritz Thoma}{moritz.thoma@tum.de}

  \icmlkeywords{Machine Learning, ICML}

  \vskip 0.3in
]

\printAffiliationsAndNotice{\icmlEqualContribution}

\input{content/00_abstract}

\section{Introduction}
\input{content/01_intro}

\section{Related Work}
\input{content/02_related_work}

\section{Preliminaries}\label{sec:preliminary_fisher}
\label{sec:preliminaries}
\input{content/03_preliminary}

\section{Methodology}
\input{data/plots/method_overview}

\input{content/04_methodolgy}

\section{Evaluation}
\input{content/05_evaluation}

\section{Conclusion and Limitations}
\input{content/06_conclusion}

\section*{AI Acknowledgement}
The creation of this paper was supported by an AI (Claude Code, Anthropic) to help arrange and proof-read the content, and support the implementation of the added related approaches. The authors take responsibility for the research methodology, experimental results, analyses, and manuscript content, including material developed with AI assistance.

\section*{Impact Statement}
This paper presents work whose goal is to advance the field of machine learning.\ There are many potential societal consequences of our work, none of which we feel must be specifically highlighted here.

\bibliography{references}
\bibliographystyle{icml2026}
\newpage
\onecolumn
\printglossary

\newpage
\appendix
\glsresetall %
\onecolumn
\input{content/07_supplementary}

\end{document}

%% file: content/00_abstract.tex
\begin{abstract}
Low-rank \glslink{svd}{SVD}\glsunset{svd}-based compression offers a powerful strategy to reduce the computational costs of \glslink{llm}{LLMs}\glsunset{llm}. However, existing methods face two key limitations: (i) global rank allocation, where uncalibrated error proxies fail to capture complex error propagation, and (ii) decomposition quality, where Fisher-based estimators suffer from severe rank collapse.
In this work, we address these limitations by introducing \textbf{\glslink{oursearch}{Layer-wise Error Modeling Search (LEMS)}\glsunset{oursearch}} and \textbf{\glslink{ours}{KFAC-SVD}\glsunset{ours}}. %
\gls{oursearch} advances rank allocation by introducing a layer-wise error surrogate that integrates local and global layer importance alongside a propagation bias,
enabling effective global rank allocation via an \glslink{ilp}{ILP}\glsunset{ilp} formulation.
\gls{ours} improves decomposition quality by utilizing token-wise statistics, mitigating the rank deficiency observed in prior Fisher-based \gls{svd} approaches.
Across \glslink{mistral}{Mistral}\glsunset{mistral}, \glslink{qwen3}{Qwen3}\glsunset{qwen3}, and \glslink{llama3}{Llama 3}\glsunset{llama3} model families, we show that \gls{oursearch} consistently outperforms existing search strategies, delivering significant zero-shot accuracy gains of up to 4.7 \glslink{pp}{p.p.}\glsunset{pp}\ that generalize to model sizes of 70B parameters, while \gls{ours} achieves an average perplexity improvement of 19\%. %
\linebreak\textit{Project Page \& Code: \href{https://lems-svd.github.io}{https://lems-svd.github.io}}
\end{abstract}

%% file: content/01_intro.tex
The rapid scaling of \glspl{llm}~\cite{yang2025qwen3, grattafiori2024llama3, jiang2023mistral7b} has enabled previously unattainable capabilities, but introduced substantial computational and memory requirements that limit deployment in resource-constrained environments.\
To address these limitations, low-rank approximation using \gls{svd} has emerged as a principled and effective compression strategy~\cite{qinsi2025dobisvd, 2025-svd-llmv2, wang2024svd-llm, wang2025bitstack, chekalina2025generalizedfisherweightedsvdscalable}. By decomposing dense weight matrices into smaller factors, \gls{svd} substantially reduces parameter counts and computational complexity while retaining easy-to-accelerate dense tensor computations. This ensures immediate compatibility with standard inference engines, avoiding the specialized hardware or sparse kernels that are often required by semi-structured or unstructured pruning approaches. Prior work has demonstrated that \gls{svd}-based methods yield significant inference acceleration, particularly in memory-bound regimes, while achieving performance competitive with state-of-the-art pruning techniques~\cite{qinsi2025dobisvd, 2025-svd-llmv2, wang2024svd-llm}. They also demonstrate that low-rank approximation is orthogonal to quantization, allowing the two methods to be combined effectively for maximum efficiency.

However, the full potential of \gls{svd} in modern \glspl{llm} is limited by two fundamental challenges: \textbf{global rank allocation} and \textbf{decomposition quality}. The primary bottleneck is the global allocation problem, which requires determining the optimal rank for each layer under a fixed parameter budget.
Existing strategies fail to solve this optimization effectively, as they rely on uncalibrated local sensitivity proxies~\cite{azizi2024memoryViT, 2025-svd-llmv2} or expensive measurements~\cite{2025_flar-svd, yuan2024asvd} to determine layer importance.\ These methods typically neglect the non-linear \textit{error explosion} inherent in deep networks~\cite{atp_huang2025determining}, creating a critical misalignment between error proxy and actual degradation. Consequently, their ability to identify high-performance rank allocations remains limited. %
Further, high decomposition quality requires distinguishing and preserving the weight directions that are most critical for the downstream task. While Fisher-weighted \gls{svd} theoretically addresses this by aligning reconstruction error with the loss landscape (curvature), existing work~\cite{chekalina2025generalizedfisherweightedsvdscalable} suffers from rank deficiencies under reasonable sample counts, limiting its effectiveness.

In this work, we address both challenges by introducing \textbf{\gls{oursearch}} for global rank allocation and \textbf{\gls{ours}} for robust decomposition. \gls{oursearch} introduces a layer-wise error model that integrates local sensitivity, empirical calibration, and a bias correcting error propagation. By explicitly modeling layer-wise error accumulation, we can formulate the search as an \gls{ilp} and solve it efficiently.
\gls{ours} mitigates the rank deficiency that limits the decomposition quality of prior Fisher-based \gls{svd} by transitioning from sequence-based to token-wise statistics for the Fisher approximation. By accumulating token-level loss curvature information, we effectively prevent rank collapse, yielding a decomposition of significantly higher fidelity.\
Our contributions are threefold:
\vspace{-0.11in} %
\begin{enumerate}
    \item \textbf{\gls{oursearch}.} We propose a layer-wise error model for rank allocation in \gls{svd} that captures both local sensitivity and global error propagation modeling error propagation via a bias. %
    To obtain global rank allocations we search the error model with an \gls{ilp} formulation.
    \vspace{-0.05in}
    \item \textbf{Robust Fisher Decomposition (\gls{ours}).} We leverage a token-wise Kronecker-factor estimation in \gls{svd} to mitigate the rank bottleneck of prior Fisher-based \gls{svd} and improve its decomposition quality.

    \vspace{-0.05in}
    \item \textbf{Empirical Validation.}
    Across \gls{mistral}, \gls{qwen3}, and \gls{llama3}, \textbf{\gls{ours}} improves \glslink{wiki}{\acrshort{wiki}}\glsunset{wiki} perplexity by an average of \textbf{19}\% over baselines. \textbf{\gls{oursearch}} further drives substantial accuracy gains of \textbf{2.1}-\textbf{4.7} \gls{pp} compared to the best existing rank allocation methods. Crucially, it also generalizes to activation-based \gls{svd} and scales to the 70B parameter regime, achieving a \textbf{4.0}~\gls{pp} improvement over uniform rank allocation.
\end{enumerate}

%% file: content/02_related_work.tex
\paragraph{Low Rank Decomposition.} Standard \gls{svd} compression replaces a weight, $\mathbf{W}$, with a rank-$r$ approximation, $\mathbf{W}_r$, that minimizes $\lVert \mathbf{W}-\mathbf{W}_r\rVert_F$. This is optimal for weight reconstruction, but often suboptimal for model performance, as deviation magnitude is a poor proxy for importance.\ To better match the forward computation, %
recent methods optimize an \textbf{activation-based objective}, targeting the layer-output error $\lVert \mathbf{W}\mathbf{X} - \mathbf{W}_r\mathbf{X}\rVert_F$.
\glslink{asvd}{ASVD}\glsunset{asvd}~\cite{yuan2024asvd} weights the rows of the weight matrices by activation importance. Building on this idea, \glslink{svdllm}{SVD-LLM}\glsunset{svdllm}~\cite{wang2024svd-llm} demonstrates that whitening $\mathbf{W}$ using a Cholesky factor of the input activation covariance yields the optimal rank-$r$ approximation for minimizing the Frobenius layer-output error. %
Subsequent work improves numerical stability~\cite{2025-svd-llmv2} and estimation accuracy of the covariance~\cite{2025_flar-svd} while keeping the same output-reconstruction objective.\ Basis-Sharing~\cite{wang2025basis} builds on \gls{svdllm} but reuses low-rank bases across layers to reduce parameters on disk, creating unique memory-computation trade-offs.\ %
\glslink{dobisvd}{DOBI-SVD}\glsunset{dobisvd}~\cite{qinsi2025dobisvd} departs from direct weight whitening by truncating activations and fitting low-rank weights that best preserve the retained features: Because activations are modified during the forward pass, the effective objective differs slightly from \gls{svdllm}-style methods. It also introduces remapping, which is essentially quantization of decomposed components. \glslink{bitstack}{BitStack}\glsunset{bitstack}~\cite{wang2025bitstack} fundamentally alters the decomposition process, separating it from prior \gls{svd}-based methods.

While activation-based methods effectively preserve feature representations, complementary research targets the loss landscape directly.\ These methods model compression impact via a local second-order (Taylor) approximation, yielding \textbf{Fisher-weighted objectives}.\ Fisher-weighted \gls{svd}~\cite{hsu2022fwsvd} uses a diagonal Fisher approximation to weight truncation by parameter importance, while \glslink{gfwsvd}{GFWSVD}\glsunset{gfwsvd}~\cite{chekalina2025generalizedfisherweightedsvdscalable} improves curvature modeling using a Kronecker-factored Fisher, demonstrating that two-sided whitening yields the optimal rank-$r$ solution under this approximation. %

\emph{Our approach.}
Activation-based methods are robust but loss-agnostic, whereas Fisher-weighted compression is loss-aware. %
Current Fisher-based work relies on sequence-gradient fitting, %
sample-limited at \gls{llm} scale. To mitigate, we estimate Kronecker factors from token-wise second moments, making truncation viable and stable in large models.

\paragraph{Searching for Ranks.} Assigning different ranks to different layers can substantially improve quality at a fixed compression budget~\cite{yuan2024asvd, 2025_flar-svd, 2025-svd-llmv2, azizi2024memoryViT}. Existing approaches primarily differ in their search and optimization strategies.

\textbf{(i) Greedy rank allocation.}
Several methods use greedy procedures that progressively increase compression while tracking reconstruction metrics. Both \glslink{mrcs}{MRCS}\glsunset{mrcs}~\cite{azizi2024memoryViT} and \glslink{fastlrd}{FastLRD}\glsunset{fastlrd}~\cite{Luo2024FastLRD} iteratively adjust ranks to control energy loss or weight-based similarity, aiming to keep layer-wise reconstruction error small. %

\textbf{(ii) Balanced reconstruction error.}
A commonly adopted heuristic allocates compression across layers to achieve similar reconstruction errors.\ \gls{asvd} evaluates candidate layer-wise configurations and applies a binary search to find a uniform error threshold that satisfies a global budget. \glslink{flarsvd}{Flar-SVD}\glsunset{flarsvd}~\cite{2025_flar-svd} performs per-layer binary searches to more precisely equalize reconstruction error in vision transformers, while \glslink{svdllmv2}{SVD-LLMv2}\glsunset{svdllmv2}~\cite{2025-svd-llmv2} groups weights by type (attention vs.\ \glslink{mlp}{MLP}\glsunset{mlp}) and redistributes compression within each group based on relative loss contributions inferred from removed singular values.

\textbf{(iii) Learned rank allocation.}
A further class of methods learns rank importance directly.\ \glslink{comcat}{ComCat}\glsunset{comcat}~\cite{xiao2023comcat} introduces trainable signals for per-layer rank allocation, while \glslink{ars}{ARS}\glsunset{ars}~\cite{gao2024adaptive} and the search component of \gls{dobisvd}~\cite{qinsi2025dobisvd} optimize ranks via gradient-updated masks that are controlled by a small hyper-network.

\textbf{(iv) Further sparsity allocation.}
Related search problems appear in structured and semi-structured pruning.\ \glslink{mdp}{MDP}\glsunset{mdp}~\cite{sun2025mdp} formulates structured sparsity allocation as an \gls{ilp}, enabling one-shot global optimization. \glslink{atp}{ATP}\glsunset{atp}~\cite{atp_huang2025determining} analyzes error propagation and shows that lower bounds on reconstruction error increase for later layers when early layers are compressed, motivating a simple linear allocation scheme that increases compression depth-wise.

\emph{Our approach.}
Current rank-search methods often (i) rely on greedy updates with limited global optimality guarantees, (ii) target uniform error distributions rather than minimizing total error, or (iii) require gradient-based optimization with tuned hyper parameters. Inspired by \gls{ilp}-based sparsity allocation~\cite{sun2025mdp} and propagation analyses~\cite{atp_huang2025determining}, we propose to directly \textbf{minimize total error} under a global budget by solving an \gls{ilp} over discrete per-layer rank candidates using a layer-wise proxy model.

\emph{Concurrent work.} Work concurrently released to ours, such as OBD-LLM~\cite{li2026optimalbraindecompositionaccurate}, IO-SVD~\cite{abbasi2026iosvdinputoutputwhitenedsvd}, ZS-SVD~\cite{abbasi2026zero} and Swift-SVD~\cite{qi2026swiftsvd} is discussed and evaluated in \cref{supp:sec:concurrent}.

%% file: content/03_preliminary.tex
Current \gls{svd}-compression methods generally optimize either a task-aware importance metric %
or an activation-preserving reconstruction error.
In \textbf{Fisher-weighted \gls{svd}-compression}, we seek a rank-$r$ approximation, $\widehat{\mathbf{W}}$, for a linear layer weight, $\mathbf{W}\in\mathbb{R}^{d_{\text{out}}\times d_{\text{in}}}$, that minimizes the compression-induced loss increase, $\Delta\mathcal{L}$. By using a second-order Taylor expansion and approximating the Hessian with the Fisher matrix, $\mathbf{F}$, \cite{fisherhessianmartens} we get the objective:
\begin{equation}
\min_{\mathrm{rank}(\widehat{\mathbf{W}})\le r}\; \Delta\bm{\theta}^\top \mathbf{F}\,\Delta\bm{\theta}
\label{eq:fisher_objective}
\end{equation}
where $\Delta\bm{\theta}=\mathrm{vec}(\mathbf{W}-\widehat{\mathbf{W}})$ represents the vectorized change in parameters.\ Since the complete Fisher matrix, $\mathbf{F}\in\mathbb{R}^{(d_{\text{out}}d_{\text{in}})\times(d_{\text{out}}d_{\text{in}})}$, is intractable due to the dimensionality of \gls{llm} layers, i.e., $(d_{\text{out}}d_{\text{in}})\times(d_{\text{out}}d_{\text{in}})$, structured approximations are imperative.\ Thus, we employ Kronecker-factored approximations of the layer-wise Fisher matrix,
\begin{equation}
\mathbf{F}\;\approx\;\mathbf{A}\otimes \mathbf{G},
\qquad
\mathbf{A}\in\mathbb{R}^{d_{\text{in}}\times d_{\text{in}}},\;
\mathbf{G}\in\mathbb{R}^{d_{\text{out}}\times d_{\text{out}}},
\label{eq:kron_fisher}
\end{equation}
where $\mathbf{A}$ and $\mathbf{G}$ are positive semidefinite factors.\
Substituting~\cref{eq:kron_fisher} into~\cref{eq:fisher_objective} yields a Frobenius-norm objective in whitened coordinates:
\begin{equation}
\begin{split}
    & \min_{\mathrm{rank}(\widehat{\mathbf{W}})\le r}\;\Delta\bm{\theta}^\top (\mathbf{A}\otimes \mathbf{G})\,\Delta\bm{\theta} \\
    &\quad =\; \min_{\mathrm{rank}(\widehat{\mathbf{W}})\le r}\;\left\| \mathbf{G}^{1/2}\,(\mathbf{W}-\widehat{\mathbf{W}})\,\mathbf{A}^{1/2} \right\|_F^2.
\end{split}
\label{eq:weighted_frob}
\end{equation}
\cref{eq:weighted_frob} generalizes the \textbf{activation-based SVD} whitening used in \gls{svdllm}~\cite{wang2024svd-llm}. Specifically, setting $\mathbf{G}=\mathbf{I}$ recovers the \gls{svdllm} objective, reducing the problem to minimizing the layer-wise output reconstruction error weighted solely by the input covariance, $\mathbf{A}$.
Then the optimal rank-$r$ approximation for~\cref{eq:weighted_frob} is obtained by truncating the \gls{svd} of $\widetilde{\mathbf{W}} = \mathbf{G}^{1/2}\mathbf{W}\mathbf{A}^{1/2}$ \mbox{\cite{chekalina2025generalizedfisherweightedsvdscalable}}.\
Specifically, let $\widetilde{\mathbf{W}}_r=\mathbf{U}_r\mathbf{\Sigma}_r\mathbf{V}_r^\top$ denote the rank-$r$ truncation of $\widetilde{\mathbf{W}}$, then
\begin{equation}
\widehat{\mathbf{W}} =
\mathbf{G}^{-1/2}\,\mathbf{U}_r\mathbf{\Sigma}_r\mathbf{V}_r^\top\,
\mathbf{A}^{-1/2}.
\label{eq:unwhiten}
\end{equation}
$\widehat{\mathbf{W}}$ is implemented as two consecutive linear layers by symmetrically splitting the singular values,
\begin{equation}
\mathbf{B}=\mathbf{\Sigma}_r^{1/2}\mathbf{V}_r^\top\mathbf{A}^{-1/2},
\qquad
\mathbf{C}=\mathbf{G}^{-1/2}\mathbf{U}_r\mathbf{\Sigma}_r^{1/2},
\label{eq:factors_prelim}
\end{equation}
such that $\widehat{\mathbf{W}}=\mathbf{C}\mathbf{B}$ and the forward pass becomes $y=\mathbf{C}(\mathbf{B}x)$. The bias terms, if present, are kept unchanged.

%% file: data/plots/method_overview.tex
\begin{figure*}[t] %
    \centering

    \begin{subfigure}[b]{0.35\textwidth}
        \centering
        \resizebox{\linewidth}{!}{
            \input{data/plots/method_overview_concept}
        }
        \caption{\textbf{\gls{ours}}: Per token processing to increase the effective sample count.}
        \label{fig:sub1}
    \end{subfigure}
    \hfill %
    \begin{subfigure}[b]{0.3\textwidth}
        \centering
        \resizebox{\linewidth}{!}{
            \input{data/plots/method_overview_score}
        }
        \caption{\textbf{\gls{oursearch} scaling}: Reference based energy rescaling
        to correct relative ordering.
        }
        \label{fig:method_overview_score}
    \end{subfigure}
    \hfill %
    \begin{subfigure}[b]{0.3\textwidth}
        \centering
        \resizebox{\linewidth}{!}{
            \input{data/plots/method_overview_bias}
        }
        \caption{\textbf{\gls{oursearch} bias}: Error bias to account for accumulating error properties.}
        \label{fig:method_overview_bias}
    \end{subfigure}
    \caption{\textbf{Overview of \gls{ours} and \gls{oursearch}.} \textbf{(a)} \gls{ours}
    mitigates the rank bottleneck of prior Fisher-based \gls{svd} by leveraging token-wise statistics (N$\gg$k). \textbf{(b, c)} \gls{oursearch} enables effective allocation by \textbf{(b)} scaling raw spectral (spec.) energy to reflect true empirical importance and \textbf{(c)} adding a depth-dependent bias to model error depth-dependent propagation. See Supp.~\ref{supp:sec:in_depth_search_ablations} for empirical validation.}
    \label{fig:three_plots}
\end{figure*}
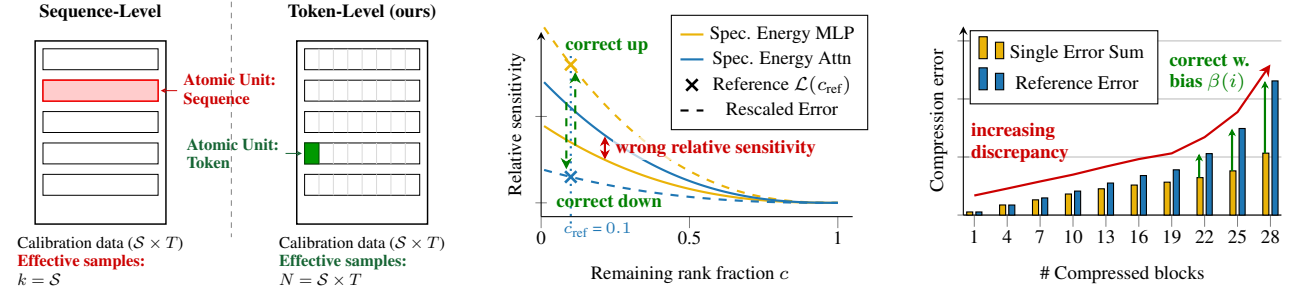

%% file: data/plots/method_overview_concept.tex
\begin{tikzpicture}[
    font=\sffamily,
    >=stealth,
    tensor/.style={draw, thick, fill=white, minimum width=2.5cm, minimum height=3.5cm, inner sep=0pt},
    rowhighlight/.style={fill=red!20, draw=red, thick},
    tokenhighlight/.style={fill=green!60!black, draw=green!30!black,  thin},
    labelstyle/.style={align=center, font=\small},
    gridstyle/.style={draw, gray!50, thin}
]

    \node[font=\bfseries] (titleA) at (0, 2.3) {Sequence-Level};

    \node[tensor] (batchA) at (0, 0) {};
    \node[below=0.1cm of batchA, font=\footnotesize, align=left] {Calibration data ($\mathcal{S} \times T$)\\\textcolor{nice_red}{\textbf{Effective samples:}}\\$k=\mathcal{S}$};

    \foreach \y in {-1.2, -0.6, 0.0, 0.6, 1.2} {
        \draw[fill=white] (-1.1, \y) rectangle (1.1, \y+0.4);
    }

    \draw[rowhighlight] (-1.1, 0.6) rectangle (1.1, 1.0);

    \node[right=0.2cm of batchA, yshift=0.8cm, font=\small, text=nice_red, align=left] (rowLabel) {\textbf{Atomic Unit:}\\\textbf{Sequence}};
    \draw[->, nice_red] (rowLabel.west) -- (1.15, 0.8);

    \begin{scope}[xshift=5cm]

        \node[font=\bfseries] (titleB) at (0, 2.3) {Token-Level (ours)};

        \node[tensor] (batchB) at (0, 0) {};
        \node[below=0.1cm of batchB, font=\footnotesize, align=left] {Calibration data ($\mathcal{S} \times T$)\\\textcolor{DarkGreen}{\textbf{Effective samples:}}\\ $N=\mathcal{S}\times T$};

        \foreach \y in {-1.2, -0.6, 0.0, 0.6, 1.2} {

            \draw[fill=white] (-1.1, \y) rectangle (1.1, \y+0.4);

            \foreach \x in {-0.825, -0.55, -0.275, 0.0, 0.275, 0.55, 0.825} {
                \draw[gray!30] (\x, \y) -- (\x, \y+0.4);
            }
        }

        \draw[tokenhighlight] (-1.1, -0.6) rectangle (-0.825, -0.2);

        \node[left=0.2cm of batchB, yshift=-0.4cm, font=\small, text=DarkGreen, align=left] (tokenLabel) {\textbf{Atomic Unit:}\\\textbf{Token}};
        \draw[->, DarkGreen] (tokenLabel.east) -- (-1.15, -0.4);

    \end{scope}

    \draw[dashed, gray] (2.5, 2.5) -- (2.5, -2.0);

\end{tikzpicture}

%% file: data/plots/method_overview_score.tex
\begin{tikzpicture}
    \tikzset{
        sensitive_color/.style={red!80!black},
        robust_color/.style={blue!70!black},
        dark_green_color/.style={green!50!black},
        anchor_marker/.style={
            mark=x,
            mark options={scale=1.5, line width=1.0pt},
            only marks
        },
        /pgfplots/legend image code/.code={%
            \draw[mark repeat=2,mark phase=2]
                plot coordinates {
                (0cm,0cm)
                (0.15cm,0cm)        %
                (0.3cm,0cm)         %
            };%
        },
    }
    \definecolor{nice_blue}{HTML}{1F77B4}

    \begin{axis}[
        width=1.1\textwidth, height=4.0cm,
        xlabel={\scriptsize Remaining rank fraction $c$ \textcolor{white}{p}},
        xmin=0, xmax=1.05,
        ymin=-0.8, ymax=9.9,
        grid style={dashed, gray!30},
        xtick = {0, 0.5, 1.0},
        xticklabels = {0, 0.5, 1},
        legend style={
            at={(1.05, 0.78)},
            anchor=east,
            fill=none,
            font=\scriptsize,
            row sep=-1pt,
        },
        legend image post style={x=0.2cm},
        ylabel style={yshift=-2.6em},
        ylabel={\scriptsize Relative sensitivity},
        xlabel shift = {yshift=-0.1 cm},
        clip=false,
        yticklabel style = { font=\scriptsize, },
        yticklabel shift = {-0.05cm},
        xticklabel style = { font=\scriptsize, yshift=+0.08cm },
        axis x line* = bottom,
        axis y line = left,
        xtick align=outside,
        yticklabel=\empty, %
    ]

        \addlegendimage{thick, DarkYellow}
        \addlegendentry{Spec. Energy MLP}

        \addlegendimage{thick, icmlblue}
        \addlegendentry{Spec. Energy Attn}

        \addlegendimage{anchor_marker, black}
        \addlegendentry{Reference $\mathcal{L}(c_\text{ref})$}

        \addlegendimage{thick, dashed, black}
        \addlegendentry{Rescaled Error}

        \addplot[
            domain=0.01:1, samples=100, thick, dashed,
            DarkYellow,
            forget plot
        ] {(8 * (1-x)^2.5) / 0.768};

        \addplot[
            domain=0.01:1, samples=100, thick, %
            DarkYellow,
            forget plot
        ] {(3.5 * (1-x)^2.5) / 0.768};

        \addplot[
            domain=0.01:1, samples=100, thick, %
            icmlblue,
            forget plot
        ] {(5.5 * (1-x)^2.5) / 0.768};

        \addplot[
            domain=0.01:1, samples=100, thick,  dashed,
            icmlblue,
            forget plot
        ] {(1.5 * (1-x)^2.5) / 0.768};

        \addplot[DarkYellow, anchor_marker] coordinates {(0.1, 8)};
        \addplot[icmlblue, anchor_marker] coordinates {(0.1, 1.5)};

        \draw[icmlblue, dotted, thick] (axis cs:0.1, -1.2) -- (axis cs:0.1, 8.5);
        \node[icmlblue, anchor=north] at (axis cs:0.19, -0.6) {\tiny $c_{\text{ref}}\text{ = }0.1$};

        \draw[-{Stealth[length=1.5mm, width=1.5mm]}, thick, dashed, dark_green_color, >={Stealth[length=3mm]}]
            (axis cs:0.085, 5.83) -- (axis cs:0.085, 1.94);
        \draw[-{Stealth[length=1.5mm, width=1.5mm]}, thick, dashed, dark_green_color, >={Stealth[length=3mm]}]
            (axis cs:0.115, 3.31) -- (axis cs:0.115, 7.57);

        \node[dark_green_color, anchor=west, align=left, font=\scriptsize] at (axis cs:0.05, 9.1) { %
            \textbf{correct up}
        };
        \node[dark_green_color, anchor=west, align=left, font=\scriptsize] at (axis cs:0.02, 0.1) { %
            \textbf{correct down}
        };
        \node[anchor=west, align=left, font=\scriptsize,fill=white,       %
                fill opacity=0.7, %
                text opacity=1,    %
                inner sep=1pt,
                ] at (axis cs:0.24, 3.15) {
            \textcolor{nice_red}{\textbf{wrong relative sensitivity}}
        };
        \draw[{Stealth[length=1.3mm, width=1.3mm]}-{Stealth[length=1.3mm, width=1.3mm]}, thick, nice_red, >={Stealth[length=3mm]}]     (axis cs:0.215, 2.51) -- (axis cs:0.215, 3.85);
    \end{axis}
\end{tikzpicture}

%% file: data/plots/method_overview_bias.tex
\begin{tikzpicture}
\definecolor{bright_green}{HTML}{2CA02C}
\definecolor{nice_blue}{HTML}{1F77B4}

\begin{axis}[
    ybar, ymin=0,
    xlabel=\scriptsize{\# Compressed blocks},
    xlabel shift = {-0.1 cm},
    xtick=data,
    xticklabel style = { font=\scriptsize, yshift=+0.08cm },
    ylabel style={yshift=-2.6em},
    ylabel={\scriptsize Compression error},
    xtick = {2, 14, 26, 38, 50, 62, 74, 86, 98, 110},
    xticklabels = {1, 4, 7, 10, 13, 16, 19, 22, 25, 28},
    axis x line* = bottom,
    axis y line = left,
    height = 4.0cm,
    width = 1.1\textwidth,
    ymin = 0.0, ymax = 6.50,
    xmin=-2.0, xmax=114.0,
    bar width=2.2px,
    yticklabel style = { font=\scriptsize, },
    yticklabel shift = {-0.05cm},
    ylabel shift = {-2.9 cm},
    ymajorgrids,
    ybar=0.04cm,
    legend style={at={(0.02, 0.98)}, anchor=north west, font=\scriptsize},
    yticklabel=\empty, %
    ]

\addplot[fill=DarkYellow] coordinates {
(2, 0.10692787170410156)
(14, 0.3477010726928711)
(26, 0.5245985984802246)
(38, 0.7252974510192871)
(50, 0.9055156707763672)
(62, 1.0340080261230469)
(74, 1.1332406999999999) %
(86, 1.2901582717895508)
(98, 1.5226831436157227)
(110, 2.1331396102905273)
};
\addlegendentry{\scriptsize Single Error Sum};

\addplot[fill=icmlblue] coordinates {
(2, 0.10692787170410156)
(14, 0.34619140625)
(26, 0.5905685424804688)
(38, 0.8261260986328125)
(50, 1.1029815673828125)
(62, 1.3616180419921875)
(74, 1.55902099609375)
(86, 2.116943359375)
(98, 2.982177734375)
(110, 4.621826171875)
};
\addlegendentry{\scriptsize Reference Error};

\addplot[
    sharp plot,
    thick,
    nice_red,
    mark=none,         %
    -{Stealth[length=1.5mm, width=1.5mm]},                %
    >=Stealth,         %
    yshift=6pt         %
] coordinates {
(2, 0.10692787170410156)
(14, 0.34619140625)
(26, 0.5905685424804688)
(38, 0.8261260986328125)
(50, 1.1029815673828125)
(62, 1.3616180419921875)
(74, 1.55902099609375)
(86, 2.116943359375)
(98, 2.982177734375)
(110, 4.621826171875)
};

\draw[-{Stealth[length=1.0mm, width=1.0mm]}, thick, green!50!black] (axis cs:108, 2.133) -- (axis cs:108, 4.622);
\draw[-{Stealth[length=1.0mm, width=1.0mm]}, thick, green!50!black] (axis cs:96, 1.523) -- (axis cs:96, 2.982);
\draw[-{Stealth[length=1.0mm, width=1.0mm]}, thick, green!50!black] (axis cs:84, 1.290) -- (axis cs:84, 2.117);

\node[nice_red, font=\scriptsize, align=left, anchor=east] at (axis cs: 40, 2.5) {\textbf{increasing}\\\textbf{discrepancy}};
\node[green!50!black, font=\scriptsize, align=left, anchor=east] at (axis cs: 106, 4.9) {\textbf{correct w.} \\ \textbf{bias $\beta(i)$}};
\end{axis}
\end{tikzpicture}

%% file: content/04_methodolgy.tex
We present \textbf{\gls{ours}} that uses a token-wise estimation to improve decomposition in Fisher-\gls{svd}, and \textbf{\gls{oursearch}}, an error-model-driven strategy for global rank allocation.

\subsection{Decoupled Token-Wise Estimation (\gls{ours})}
\paragraph{Motivation.} Once the Kronecker factors, $\mathbf{A}$ and $\mathbf{G}$, are fixed, the Fisher-weighted truncation in \cref{eq:weighted_frob} has a standard closed-form solution and is identical across
methods.\ \gls{gfwsvd}~\cite{chekalina2025generalizedfisherweightedsvdscalable} determines the factors $\mathbf{A}$ and $\mathbf{G}$ using $k$ per-sequence gradients (individual ones or sums over micro-batches). Let $\mathbf{G}_i^{(\mathbf{W})}=\nabla_{\mathbf{W}}\ell_i\in\mathbb{R}^{d_{\text{out}}\times d_{\text{in}}}$ denote the weight gradient for sequence $i$ and $\bm{g}_i=\mathrm{vec}(\mathbf{G}_i^{(\mathbf{W})})$ its vectorized form, then the empirical Fisher $\widehat{\mathbf{F}}$ is formalized as %
\begin{equation}
\widehat{\mathbf{F}}
\;=\;
\frac{1}{k}\sum_{i=1}^k \bm{g}_i \bm{g}_i^\top.
\label{eq:emp_fisher_outer}
\end{equation}
By fitting a rank-1 approximation to $\widehat{\mathbf{F}}$ in a rearranged space using Lanczos~\cite{Lanczos1950}%
, \gls{gfwsvd} creates a Kronecker product $\mathbf{A}\otimes \mathbf{G}$. %
However, as~\cref{eq:emp_fisher_outer} is a sum of $k$ outer products, it obeys the strict rank bound
\begin{equation}
\text{rank}(\widehat{\mathbf{F}})\;\le\; k,
\label{eq:rank_bound}
\end{equation} inducing a \textbf{rank and conditioning bottleneck}.
In modern \gls{llm} layers, the parameter dimension $d_{\text{out}}d_{\text{in}}$ is orders of magnitude larger than feasible $k$ (e.g., $k\!\le\!256$), so curvature is observed through a low-dimensional subspace. Empirically, this manifests as spectral collapse of the fitted factors (low effective rank/poor conditioning), and increased Fisher degradation after rank-$r$ truncations under \cref{eq:weighted_frob}.

\paragraph{Method.} To avoid the rank bottleneck of \gls{gfwsvd}'s per-sequence gradient-based estimator, we estimate Kronecker factors from \emph{token-wise} second moments.
For a linear layer with input $\bm{x}$ and output gradient $\bm{\delta} = \partial\mathcal{L}/\partial \bm{y}$, the per-token weight gradient is $\nabla_{\mathbf{W}}\mathcal{L} = \bm{\delta} \bm{x}^\top$. Following \cite{kfac-martens15}, we approximate the expectation of the Kronecker product by the Kronecker product of expectations:
\begin{equation}
\mathbf{F} \kern-0.1em=\kern-0.1em \mathbb{E}[(\bm{x}\kern-0.05em\otimes\kern-0.05em\bm{\delta})(\bm{x}\otimes\bm{\delta})^\top] \kern-0.1em\approx\kern-0.05em \mathbb{E}[\bm{xx}^\top] \otimes \mathbb{E}[\bm{\delta\delta}^\top] \kern-0.1em=\kern-0.1em \mathbf{A} \otimes \mathbf{G}.
\label{eq:kfac_factorization}
\end{equation}

Unlike fitting from a small set of full weight gradients $g_i$, KFAC estimates $\mathbf{A}$ and $\mathbf{G}$ from $N$ token instances.
With $\mathcal{S}$ sequences of length $T$, the effective sample count is $N\approx \mathcal{S}T$, %
e.g., $256\times 2048 \approx 5.2\times 10^5$.
We compute
\begin{equation}
\widehat{\mathbf{G}}
=
\frac{1}{N}\sum_{i=1}^{N} \bm{\delta}_i \bm{\delta}_i^\top,
\qquad
\widehat{\mathbf{A}}
=
\frac{1}{N}\sum_{i=1}^{N} \bm{x}_i \bm{x}_i^\top.
\label{eq:token_covariances}
\end{equation}
These estimators satisfy
\begin{equation}
\kern-0.1em\text{rank}(\widehat{\mathbf{A}})\le \min(d_{\text{in}},N),
\;\;
\text{rank}(\widehat{\mathbf{G}})\le \min(d_{\text{out}},N),
\label{eq:rank_bounds_kfac}
\end{equation}
therefore avoiding the $k$-limited rank bottleneck of~\cref{eq:rank_bound} and maintaining more diverse curvature directions.\
Although this token-wise approximation omits cross-token second moments (e.g., induced by weight sharing~\cite{runa-kfacreduce}), this trade-off proves favorable in practice, as it eliminates spectral collapse, i.e., the dominant failure mode limiting \gls{gfwsvd}'s performance.
The full algorithm is provided in the supplementary~\cref{supp:sec:algorithms}.

\subsection{Global Rank Allocation via \glslink{oursearch}{Layer-wise Error Modeling (LEMS)}\glsunset{oursearch}}\label{sec:erorr_modelling}

\paragraph{Motivation.}

Effective rank allocation requires accurately predicting the contribution of a given rank choice, $r_i$, for a linear layer weight, e.g., $\mathbf{W}_\text{Q}$, $\mathbf{W}_{\text{down}}$, to the global model degradation. This problem poses three distinct challenges:
\begin{enumerate}
\item \textbf{Granularity (Shape):} Solving the allocation problem requires loss estimators for \emph{all} candidate ranks across weights; however, evaluating these via forward passes is computationally intractable.
\item \textbf{Sensitivity (Scale):} Standard proxies derived from \gls{svd} spectra, e.g., tail energy, are informative but uncalibrated. A 10\% spectral energy loss in a \gls{mlp} projection may impact the output differently than a 10\% loss in an attention projection (see \cref{fig:method_overview_score}).
\vspace{-0.05in}
\item \textbf{Propagation (Bias):} Losses are not strictly additive. Errors introduced in early layers propagate and amplify through the depth of the network. Simply summing layer-wise errors systematically underestimates global degradation for compressed models (see \cref{fig:method_overview_bias}).
\end{enumerate}
\vspace{-0.05in}
Existing approaches fail to address these challenges simultaneously. Methods relying on spectral energy~\cite{2025-svd-llmv2,azizi2024memoryViT} are uncalibrated (sensitivity) and ignore propagation.\ Similarly, measurement approaches~\cite{yuan2024asvd, 2025_flar-svd} are not only intractable for fine-grained search (granularity) but also fail to model propagation. The only exception addressing propagation, \cite{atp_huang2025determining}, sacrifices sensitivity by enforcing uniform block-wise compression rates.

\paragraph{Method.}
To systematically address these challenges, we rely on two core components: (i) a decomposable cost model that disentangles complex loss contributions via spectral shape (for granularity), a calibration point (for scale), and a depth-dependent bias (for propagation); and (ii) an \gls{ilp} that identifies the globally optimal allocation under this proxy. This enables rank selection to be formulated as a constrained combinatorial optimization problem over the set of linear layers, $\mathbf{W}=\{\mathbf{W}_1,...,\mathbf{W}_{N_L}\}$, distributed across $B$ transformer blocks. The objective is to minimize the estimated global degradation subject to a parameter budget $P_\text{target}$, which is determined by the target retention rate $c_\text{target}$.

\textbf{(i) Reliable Error Model.}
Our goal is to disentangle the complex, non-linear loss landscape into tractably additive terms. We approximate the coupled end-to-end loss using a \emph{layer-additive} surrogate, $L_i(c)$. We define this cost as the product of three calibrated terms, mapping a retained parameter ratio, $c$, to an estimated global degradation:
\begin{equation}
L_i(c) = \underbrace{S_i(c)}_{\text{Shape}} \cdot \underbrace{\alpha_i}_{\text{Scale}} \cdot \underbrace{(1+\beta(d_i))}_{\text{Bias}}.
\end{equation}
Here,
$d_i\in\{1,…,B\}$ represents the block depth where layer $i$ resides.

\textbf{(a) Analytical Sensitivity (Shape).}
Whether using Fisher-weighted or activation-based \gls{svd}, the singular value spectrum $\{\sigma_{i,k}\}$ quantifies the local information retained at any specific rank.
We exploit this to handle the granularity challenge. Let $\widetilde{\mathbf{W}}_i$ be the whitened weight matrix (as defined in \cref{sec:preliminary_fisher}). We define the normalized tail energy $S_i(c)$ for a parameter ratio $c=r_i \frac{d_{\text{in}}+d_{\text{out}}}{d_{\text{in}}d_{\text{out}}}$ (ratio of retained parameters relative to full matrix) as:
\begin{equation} S_i(c) =
\frac{\sum_{k> r_i}\sigma_{i,k}^2}{\sum_{k}\sigma_{i,k}^2}
\;=\;
\frac{\|\widetilde{\mathbf{W}}_i-\widetilde{\mathbf{W}}_{i,r_i}\|_F^2}{\|\widetilde{\mathbf{W}}_i\|_F^2}.
\label{eq:Si}
\end{equation}
$S_i(c)$ yields a computationally cheap, high-resolution curve describing the \emph{relative shape} of degradation. It quantifies the rate at which reconstruction error grows as rank decreases.

\textbf{(b) Empirical Calibration (Scale).}
The spectral measure $S_i(c)$ captures the \emph{relative} quality of ranks within a layer, but its magnitude is hard to compare between layer (types) \cite{2025-svd-llmv2}.
To fix the scale, we measure the model degradation (\glslink{kl}{KL}\glsunset{kl} divergence between original and compressed model output) at a single reference point $c_{\text{ref}}$ (e.g., 0.1) for each layer in isolation. We derive a scaling factor, $\alpha_i$, to map the spectral shape to the task loss:
\begin{equation}
l_i(c)
= S_i(c) \cdot \underbrace{\left( \frac{\mathcal{L}_{\mathrm{KL},i}(c_{\text{ref}})}{S_i(c_{\text{ref}})+\varepsilon} \right)}_{\alpha_i}.
\label{eq:li}
\end{equation}
This combines the best of both: the dense granularity of the analytical spectrum and the ground-truth accuracy of empirical probing, as illustrated in \cref{fig:method_overview_score}.

\textbf{(c) Interaction Correction (Bias).}
Summing local costs, $l_i(c)$, ignores that early errors can amplify during propagation \cite{atp_huang2025determining}. As illustrated in \cref{fig:method_overview_bias}, this mismatch grows with depth, motivating a depth-dependent prior.
However, rigid block-level constraints or Lagrangian relaxations \cite{bragin_surrogate-lagrangian_2015} are ill-suited here, as they conflate individual layer importance through score aggregation and complicate discrete (non-smooth) \gls{ilp} penalty multiplier tuning. Instead, we apply a \emph{soft} bias that modulates layer costs by block index, $d_i$, modeling cumulative downstream impact as a harmonic tail sum:
\begin{equation}
\beta(d_i) = \kappa \cdot \frac{\sum_{k=d_i}^{B} k^{-\gamma}}{Z}, \quad Z=\sum_{k=1}^{B} k^{-\gamma}.
\end{equation}
This profile assigns larger penalties to early blocks and decays smoothly with depth. Intuitively, $\beta(\cdot)$ provides a flexible relaxation of the depth-dependent prior without imposing rigid monotonicity.\ $\kappa$ and $\gamma$ are calibrated via lightweight \gls{tpe}, adapting them to architectural properties. See Supplement~\ref{sup:search_bias_ablations} for additional details).

\textbf{(ii) \acrfull{ilp}.}
Given the calibrated costs for all layers, $L_i(c)$, we select exactly one rank configuration $j \in \mathcal{J}_i$ for each linear layer $i \in \{1, \dots, N_L\}$ to minimize the total cost under a budget constraint.
Let $p_{ij} = P_{\text{full}}^{(i)} \cdot c_{ij}$ denote the retained parameters and let $x_{ij} \in \{0, 1\}$ indicate whether the configuration is selected. Then the optimization problem is formulated as:
\begin{align}
\min_{{x_{ij}}} \quad & \sum_{i=1}^{N_L} \sum_{j \in \mathcal{J}_i} L_i(c_{ij}) \cdot x_{ij} \label{eq:ilp_obj} \\
\textrm{s.t.} \quad & \sum_{j \in \mathcal{J}_i} x_{ij} = 1, \quad \forall i \in {1, \dots, N_L} \nonumber \\
& \sum_{i=1}^{N_L} \sum_{j \in \mathcal{J}i} p_{ij} \cdot x_{ij} \le P_{\text{target}}. \nonumber
\end{align}
This allows to identify the globally optimal allocation for our proxy objective in seconds using off-the-shelf solvers.

%% file: content/05_evaluation.tex
We evaluate our proposed methods across diverse \gls{llm} families, \gls{svd} baselines and perform exhaustive ablation studies. Additional results, evaluation and further information are available in the supplementary material.
\begin{table*}[]
    \captionof{table}{
    Comparison of \textbf{\gls{oursearch}} against state-of-the-art search baselines across modern \gls{llm}s and two compression rates.
    }
    \centering
    \resizebox{0.97\textwidth}{!}{%
        \input{data/tables/search_approach_comparison_cr}
    }
    \label{tab:search_results_wide}
\end{table*}

\textbf{Models, Baselines and Datasets.} We compare \gls{oursearch} to a variation of searches. Specifically, we pick one representative search from each group mentioned: greedy, \gls{mrcs}~\cite{azizi2024memoryViT}; similar error, \gls{asvd}~\cite{yuan2024asvd} and \gls{svdllmv2}~\cite{2025-svd-llmv2}); gradient based, \gls{ars}~\cite{gao2024adaptive}; and the principled \gls{atp}~\cite{atp_huang2025determining} search. %
Additionally, we compare our method against a diverse set of recent \gls{svd}-based approaches, namely \glslink{fwsvd}{\acrshort{fwsvd}}\glsunset{fwsvd}~\cite{hsu2022fwsvd}, \gls{asvd}~\cite{yuan2024asvd}, \gls{svdllm}~\cite{wang2024svd-llm}, \gls{svdllmv2}~\cite{2025-svd-llmv2}, \gls{dobisvd}~\cite{qinsi2025dobisvd} and \gls{gfwsvd}~\cite{chekalina2025generalizedfisherweightedsvdscalable}. We evaluate \gls{mistral}~\cite{jiang2023mistral7b}, \gls{llama3}~\cite{grattafiori2024llama3}, and \gls{qwen3}~\cite{yang2025qwen3} models on \gls{wiki}~\cite{merity2017pointer_wikitext2}, PTB~\cite{ptb}, C4~\cite{c4} and six zero-shot benchmarks (PIQA~\cite{bisk2020piqa}, OpenBookQA~\cite{mihaylov2018openbookqa}, HellaSwag~\cite{zellers2019hellaswag}, ARC-Easy/ARC-Challenge~\cite{clark2018arc}, and Winogrande~\cite{sakaguchi2021winogrande}), using the LM-Eval library~\cite{eval-harness}. %

\textbf{Implementation Details.} To ensure fair comparison, all methods are tested under identical settings (incl. seed). For calibration of \gls{svd} algorithms, we use 256 random sequences of length 2048 from \gls{wiki}. %
For search methods involving data, we use 32 samples from \gls{wiki} following \gls{asvd}. For the \gls{ilp} search we use the Gurobi solver \mbox{\cite{gurobi}}, with a per layer search range of $c_i=[0.3,1.0]$. Compression ratios refer to the remaining parameters of linear layers (excluding head).\ All experiments are run on a single NVIDIA H200 GPU.

\textbf{Additional Results.} For detailed tables, data ablations, more architectures and results on non-standard \gls{svd} (i.e. \glslink{basissharing}{BasisSharing}\glsunset{basissharing}, \glslink{bitstack}{\acrshort{bitstack}}\glsunset{bitstack}) we refer to the Supplement.

\begin{table*}[t]
\centering
\caption{
Comparison of \textbf{\gls{ours}} with activation- and Fisher-based \gls{svd} across modern \gls{llm}s under uniform compression and \textbf{\gls{oursearch}}.}
\resizebox{0.965\textwidth}{!}{%
\input{data/tables/llm_main_table_cr}
}
\label{tab:llm_main_table_short}
\end{table*}

\subsection{Rank Allocation Results}
We evaluate the proposed  \textbf{\textsc{\gls{oursearch}}} rank allocation against representative searches. %
All methods employ the \emph{same} decomposition pipeline, calibration data and candidate ranks, differing only in rank selection (refer to Supplement~\ref{sup:experiment_setup_long}). We report perplexity on \acrshort{wiki}, \acrshort{c4} and \acrshort{ptb} (lower is better), and average zero-shot accuracy (higher is better) for two compression ratios (0.8, 0.6) across three architectures (\cref{tab:search_results_wide}). Per-task accuracies are provided in Supplement~\ref{supp:sec:detailed_main_tables}.

\emph{Uniform search}: Across all six settings, \gls{oursearch} achieves the highest accuracy and substantially reduces perplexity relative to uniform rank allocation. The accuracy gain over the \textit{second-best} method ranges from \textbf{2.14} to \textbf{4.67} \glslink{pp}{percentage points (p.p.)}\glsunset{pp} while the
(mean) perplexity averaged over \gls{wiki}, \acrshort{c4} and \acrshort{ptb} is reduced, relative to uniform allocation, by \textbf{26.06} to \textbf{75.40}\%.
\emph{Baseline searches}: Among the baselines, \gls{asvd} and \gls{svdllmv2} frequently fail to outperform uniform allocation, consistent with the limitations of noisy, sensitivity proxies (cf. Supplement~\ref{supp:sec:sensitivity_metric_discussion}) and constraints that enforce overly similar compression across heterogeneous layer types (cf. Supplement~\ref{sup:search_detailed_comparison_atp_asvd}), respectively. Similarly, \gls{mrcs} struggles in most settings.
\gls{ars} is the strongest baseline in accuracy, ranking second in four of the six settings. However, its \gls{wiki} perplexity is worse than uniform allocation in four of the six settings, and it requires gradients, more data, and a stabilized schedule to reliably meet target ratios.
Despite its simplicity, \gls{atp}, which is the only baseline explicitly modeling accumulating errors, generally matches or narrowly outperforms uniform allocation and remains competitive with other searches, especially on \gls{llama3}.\ However, because it assumes error accumulation as a linear function, it cannot capture layer- and type-specific importance deviations. Overall, the results strongly validate the efficacy of \gls{oursearch}.%

\subsection{\gls{svd}-Compression Results}

We evaluate \textbf{\gls{ours}} against state-of-the-art methods across three \gls{llm} families: \gls{mistral}, \gls{llama3}, and \gls{qwen3}. \Cref{tab:llm_main_table_short} details the results for all approaches under uniform compression rates (0.9, 0.7) and calibration sample count, with an additional line for our search-enhanced variant.
\gls{ours} consistently yields the lowest perplexity, with gains most pronounced on \gls{llama3}. At a 0.7 compression ratio, \gls{ours} improves perplexity on \gls{wiki} by \textbf{49}\% over the best baseline (\gls{dobisvd}), extending to \textbf{66}\% with \gls{oursearch}.
Zero-shot accuracy largely follows this trend: under mild compression (0.9), \gls{ours} avoids the sharp accuracy drop seen in several baselines and stays in the mid-\textbf{50}s, while \gls{oursearch} recovers near-lossless performance (\textbf{61.8}\% vs. \textbf{63.3}\% original).
Strictly comparing to prior Fisher-based methods, \gls{ours} outperforms the baselines by up to \textbf{10.3} \gls{pp}\ \emph{without} the addition of \gls{oursearch}. \gls{gfwsvd} degrades significantly under identical conditions, reaffirming its conditioning challenges.
The only exception is \gls{mistral} at 0.9, where \gls{asvd} achieves a slightly higher zero-shot accuracy (+0.35 \gls{pp}) despite worse perplexity. This implies \gls{mistral} relies on features well-aligned with activation magnitude, favoring simpler proxies in uniform settings. Overall, \gls{ours} provides the most effective decomposition, especially when compared with other Fisher-based methods, outperforming state-of-the-art baselines in the vast majority of evaluated settings.

\begin{figure}[t!]
    \centering
    \resizebox{0.95\linewidth}{!}{%
    \input{data/plots/generalization_plot_qwen_family_cr}
    }
    \caption{\gls{wiki} perplexity (log scale) and average accuracy of \gls{ours} and \gls{svdllm} on \gls{qwen3} models with uniform compression and \gls{oursearch} across multiple compression ratios (0.5-0.9) and a range of sizes 4B-32B. %
    }
    \label{fig:generalization_experiments}
    \vspace{-0.2in}
\end{figure}
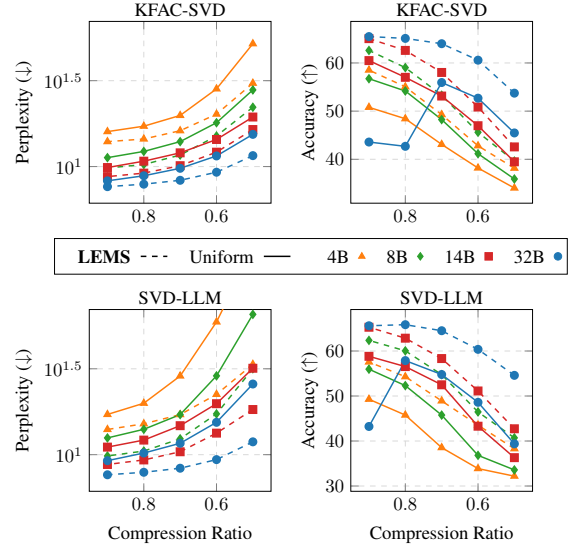

\subsection{Ablations}
This section investigates the source of \gls{ours} and \gls{oursearch} performance gains through empirical validation.
\subsubsection{Scalability of \gls{oursearch}}\label{sec:generalization_ablation}
To assess the scalability of \gls{oursearch}, we evaluate the \gls{qwen3} family (4B-32B) across five compression ratios (0.5-0.9). To demonstrate that \gls{oursearch} is agnostic to a particular decomposition objective, we report results for both \gls{ours} and the activation-based \gls{svdllm}. \cref{fig:generalization_experiments} contrasts uniform allocation (solid) to \gls{oursearch} (dashed), with numerical results provided in Supplement~\ref{supp:sec:detailed_main_tables}.

Across all budgets and model sizes, \gls{oursearch} consistently improves both perplexity and downstream utility.\
Averaged over ratios and models, it reduces perplexity by \textbf{17\%} for \gls{ours} and \textbf{34\%} for \gls{svdllm}, while increasing zero-shot accuracy by \textbf{+7.2} and \textbf{+9.1} \gls{pp}, respectively. Gains are largest under aggressive compression:\ at 0.5, \gls{oursearch} cuts perplexity by \textbf{$\sim$26\%} (\gls{ours}) / \textbf{$\sim$57\%} (\gls{svdllm}) on average and recovers \textbf{+4.8} / \textbf{+8.7} \gls{pp}\ accuracy.\ Crucially, it also resolves the sharp accuracy dips observed in \gls{qwen3}-32B under uniform allocation for both decompositions.
The benefits extend beyond \gls{qwen3}: on the substantially larger \gls{llama3}.1-70B, \gls{oursearch} reduces \gls{wiki} perplexity from 7.9 to \textbf{4.5} and increases accuracy from 62.6 to \textbf{66.6} (see Supplement~\ref{supp:sec:llama_3_70b_results}).\
Comparing decompositions methods, \gls{ours} achieves approximately \textbf{26\%} lower perplexity than \gls{svdllm} under uniform allocation, demonstrating stronger alignment with the downstream loss.\ When combined with \gls{oursearch}, perplexities become tightly matched between the two methods, with \gls{svdllm} retaining a marginal edge in zero-shot accuracy ($\sim$0.3 \gls{pp}).\
Notably, the larger gains observed for \gls{svdllm} with \gls{oursearch} suggest that suboptimal rank allocation is often a primary source of degradation under uniform compression.
Differently, through the calibrated layer-wise error model and global \gls{ilp} optimization, \gls{oursearch}  overcomes the degradation of uniform compression yielding consistent gains.

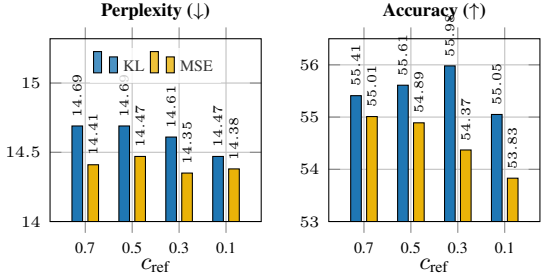
\begin{figure}[tp!]
    \centering
    \input{data/plots/scale_ablation_bars_main_content_cr}
    \caption{\gls{wiki} perplexity and zero-shot accuracy comparison of \gls{kl} (\textcolor{icmlblue}{blue bars}) and \glslink{mse}{\acrshort{mse}}\glsunset{mse} (\textcolor{DarkYellow}{yellow bars}) as rescale importance metrics with different $c_\text{ref}$ values on \gls{qwen3}-4B.}
    \label{fig:search_ablation_kl_vs_mse_and_cref}
\end{figure}

\subsubsection{Individual Contributions to LEMS}\label{sec:individual_contribution_lems}
To validate components of our rank allocation strategy, we evaluate \gls{oursearch}'s error modeling components on \gls{llama3}-8B and \gls{qwen3}-8B at 0.6 parameter retention (Table~\ref{tab:incremental_methods}).\
Introducing \gls{ilp} search via spectral energy, $S_i$, cannot accurately reflect the relative importance of layers, which results in a perplexity increase of 93\% for \gls{llama3} and 81\% for \gls{qwen3} over uniform allocation, adding 1 minute to the runtime.
Incorporating \gls{kl}-based calibration ($l_i$) yields significant improvements, reducing perplexity by 19\% for \gls{llama3} and 13\% for \gls{qwen3} below uniform allocation. Although this adds 17 to 22 minutes, the scores are fully cacheable and reusable across compression ratios, making this a \textit{one-time investment}.
\input{data/tables/search_parameter_comparison_cr}
Finally, cross-layer bias correction, $\beta$, incurs negligible overhead (2 minutes) yet yields further perplexity reductions of 42\% for \gls{llama3} (27.56 to 16.08) and 3\% for \gls{qwen3} (15.58 to 15.07). This behavior highlights the method's adaptivity: the corrections scale with error accumulation, acting aggressively for sensitive models while remaining nearly neutral for resilient ones (see Supplement~\ref{sup:search_bias_ablations}). Overall, \gls{oursearch} requires 20-25 minutes for the initial run and only \textbf{3 minutes} for subsequent runs, on the evaluated 8B-parameter architectures. An extended analysis about the performance contributions of global optimization and data-driven bias calibration is provided in Supplement \cref{supp:sec:global_optimization_vs_bias_calibration}.
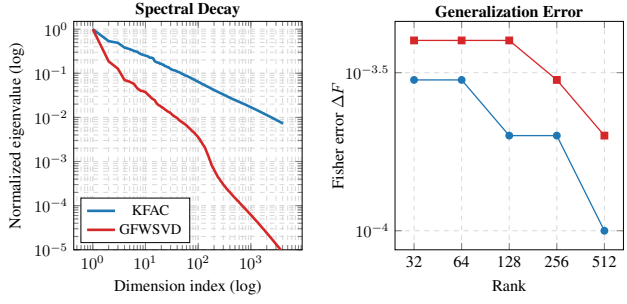
\begin{figure}[tp!]
    \centering
    \resizebox{1\linewidth}{!}{%
    \begin{tikzpicture}
    \definecolor{colKFAC}{HTML}{1F77B4}
    \definecolor{colCoupled}{HTML}{D62728}
    \definecolor{colGFWSVD}{HTML}{2CA02C}
    \definecolor{colShampoo}{HTML}{D62728}
    \begin{groupplot}[
        group style={
            group size=2 by 1,
            horizontal sep=2cm,
        },
        width=6.5cm, height=6.5cm,
        grid=both, grid style={dashed, gray!30},
        legend pos=south west,
        legend style={font=\footnotesize},
    ]

    \nextgroupplot[
        xmode=log, ymode=log,
        title={\textbf{Spectral Decay}},
        title style={yshift=-1.5ex},
        xlabel={Dimension index (log)},
        ylabel={Normalized eigenvalue (log)},
        ymin=1e-5, ymax=1.5, %
    ]

    \addplot[color=colKFAC, ultra thick] coordinates { (1, 1.00000e+00) (2, 5.34976e-01) (3, 4.88744e-01) (4, 3.81913e-01) (5, 3.53599e-01) (6, 3.22211e-01) (7, 3.06180e-01) (8, 2.74561e-01) (9, 2.61466e-01) (10, 2.51061e-01) (11, 2.43633e-01) (12, 2.26467e-01) (13, 2.25508e-01) (14, 2.19245e-01) (15, 1.86658e-01) (16, 1.79461e-01) (17, 1.74640e-01) (18, 1.72819e-01) (19, 1.63874e-01) (20, 1.63799e-01) (21, 1.57479e-01) (22, 1.56157e-01) (23, 1.46883e-01) (24, 1.46109e-01) (26, 1.37981e-01) (27, 1.37402e-01) (28, 1.34312e-01) (29, 1.29212e-01) (30, 1.25417e-01) (32, 1.21154e-01) (33, 1.19074e-01) (34, 1.17291e-01) (36, 1.15951e-01) (37, 1.15021e-01) (39, 1.10840e-01) (41, 1.08265e-01) (43, 1.06072e-01) (44, 1.05163e-01) (46, 1.02578e-01) (48, 9.86547e-02) (50, 9.70844e-02) (53, 9.34285e-02) (55, 9.07774e-02) (57, 8.92343e-02) (60, 8.78384e-02) (62, 8.64905e-02) (65, 8.37324e-02) (68, 8.13289e-02) (71, 8.00784e-02) (74, 7.70887e-02) (77, 7.54737e-02) (80, 7.38165e-02) (83, 7.19491e-02) (87, 7.06431e-02) (91, 6.71543e-02) (95, 6.66974e-02) (99, 6.46411e-02) (103, 6.25043e-02) (107, 6.19158e-02) (112, 6.00813e-02) (117, 5.76849e-02) (122, 5.62597e-02) (127, 5.51749e-02) (132, 5.45411e-02) (138, 5.26375e-02) (144, 5.09527e-02) (150, 4.98142e-02) (157, 4.87899e-02) (163, 4.76931e-02) (170, 4.60861e-02) (178, 4.47356e-02) (185, 4.38373e-02) (193, 4.27597e-02) (202, 4.12779e-02) (210, 4.06420e-02) (219, 3.97490e-02) (228, 3.87832e-02) (238, 3.77251e-02) (248, 3.68876e-02) (259, 3.59285e-02) (270, 3.51647e-02) (282, 3.42059e-02) (294, 3.34424e-02) (306, 3.26324e-02) (319, 3.20336e-02) (333, 3.12653e-02) (347, 3.03306e-02) (362, 2.95844e-02) (378, 2.87523e-02) (394, 2.82495e-02) (411, 2.75312e-02) (428, 2.68491e-02) (446, 2.61741e-02) (466, 2.56108e-02) (485, 2.51017e-02) (506, 2.45988e-02) (528, 2.40890e-02) (550, 2.35619e-02) (574, 2.29969e-02) (598, 2.25468e-02) (624, 2.20510e-02) (651, 2.15179e-02) (678, 2.10539e-02) (707, 2.06055e-02) (738, 2.01394e-02) (769, 1.97431e-02) (802, 1.92777e-02) (836, 1.88066e-02) (872, 1.83990e-02) (909, 1.79625e-02) (948, 1.75257e-02) (988, 1.71603e-02) (1031, 1.67461e-02) (1075, 1.62941e-02) (1121, 1.59447e-02) (1168, 1.55759e-02) (1218, 1.52489e-02) (1270, 1.48250e-02) (1325, 1.44998e-02) (1381, 1.41544e-02) (1440, 1.38165e-02) (1502, 1.34394e-02) (1566, 1.31315e-02) (1633, 1.27941e-02) (1702, 1.24722e-02) (1775, 1.21743e-02) (1851, 1.18623e-02) (1930, 1.15698e-02) (2012, 1.12791e-02) (2098, 1.10169e-02) (2188, 1.07249e-02) (2281, 1.04636e-02) (2378, 1.01986e-02) (2480, 9.94054e-03) (2586, 9.67836e-03) (2696, 9.43960e-03) (2811, 9.19299e-03) (2931, 8.96833e-03) (3056, 8.73448e-03) (3187, 8.50514e-03) (3323, 8.27906e-03) (3465, 8.07270e-03) (3613, 7.86325e-03) (3767, 7.66496e-03) (3928, 7.45688e-03) (4096, 7.26609e-03)  };
    \addlegendentry{KFAC}

    \addplot[color=colCoupled, ultra thick] coordinates { (1, 1.00000e+00) (2, 1.86481e-01) (3, 1.26843e-01) (4, 7.04466e-02) (5, 6.42799e-02) (6, 5.77550e-02) (7, 4.56320e-02) (8, 4.04310e-02) (9, 3.93495e-02) (10, 3.73304e-02) (11, 3.33248e-02) (12, 3.03614e-02) (13, 2.74196e-02) (14, 2.58216e-02) (15, 2.45770e-02) (16, 2.15591e-02) (17, 1.97695e-02) (18, 1.87417e-02) (19, 1.82740e-02) (20, 1.73173e-02) (21, 1.67768e-02) (22, 1.61366e-02) (23, 1.53847e-02) (24, 1.48786e-02) (26, 1.38537e-02) (27, 1.34484e-02) (28, 1.31702e-02) (29, 1.29158e-02) (30, 1.24591e-02) (32, 1.15967e-02) (33, 1.11938e-02) (34, 1.09456e-02) (36, 1.04671e-02) (37, 1.02712e-02) (39, 9.80803e-03) (41, 9.45191e-03) (43, 8.76780e-03) (44, 8.67547e-03) (46, 8.20532e-03) (48, 7.98025e-03) (50, 7.71904e-03) (53, 7.33793e-03) (55, 7.14625e-03) (57, 6.95429e-03) (60, 6.60433e-03) (62, 6.32926e-03) (65, 5.98912e-03) (68, 5.70742e-03) (71, 5.43655e-03) (74, 5.19253e-03) (77, 4.87810e-03) (80, 4.71946e-03) (83, 4.58663e-03) (87, 4.29794e-03) (91, 4.05101e-03) (95, 3.82959e-03) (99, 3.64317e-03) (103, 3.34495e-03) (107, 3.22447e-03) (112, 2.86103e-03) (117, 2.71737e-03) (122, 2.45674e-03) (127, 2.28963e-03) (132, 2.16257e-03) (138, 1.92896e-03) (144, 1.68128e-03) (150, 1.49659e-03) (157, 1.31566e-03) (163, 1.14824e-03) (170, 1.02145e-03) (178, 8.59120e-04) (185, 7.77920e-04) (193, 7.19071e-04) (202, 6.34966e-04) (210, 5.84368e-04) (219, 5.32486e-04) (228, 4.90257e-04) (238, 4.40472e-04) (248, 4.20140e-04) (259, 3.85941e-04) (270, 3.56545e-04) (282, 3.37014e-04) (294, 3.14028e-04) (306, 2.90709e-04) (319, 2.76657e-04) (333, 2.58834e-04) (347, 2.40672e-04) (362, 2.29521e-04) (378, 2.14124e-04) (394, 2.06817e-04) (411, 1.93402e-04) (428, 1.84197e-04) (446, 1.72775e-04) (466, 1.63319e-04) (485, 1.55167e-04) (506, 1.47425e-04) (528, 1.40727e-04) (550, 1.31622e-04) (574, 1.26881e-04) (598, 1.19112e-04) (624, 1.12711e-04) (651, 1.07174e-04) (678, 1.01554e-04) (707, 9.60004e-05) (738, 9.11412e-05) (769, 8.69428e-05) (802, 8.24572e-05) (836, 7.78297e-05) (872, 7.38773e-05) (909, 6.97525e-05) (948, 6.62543e-05) (988, 6.27252e-05) (1031, 5.95030e-05) (1075, 5.62317e-05) (1121, 5.31620e-05) (1168, 5.02509e-05) (1218, 4.76909e-05) (1270, 4.49798e-05) (1325, 4.28013e-05) (1381, 4.03190e-05) (1440, 3.79248e-05) (1502, 3.58220e-05) (1566, 3.39253e-05) (1633, 3.18896e-05) (1702, 3.02240e-05) (1775, 2.85820e-05) (1851, 2.68634e-05) (1930, 2.53528e-05) (2012, 2.39011e-05) (2098, 2.24734e-05) (2188, 2.13286e-05) (2281, 2.00135e-05) (2378, 1.89335e-05) (2480, 1.78023e-05) (2586, 1.67590e-05) (2696, 1.57683e-05) (2811, 1.48552e-05) (2931, 1.39751e-05) (3056, 1.31283e-05) (3187, 1.23634e-05) (3323, 1.16701e-05) (3465, 1.09718e-05) (3613, 1.02930e-05) (3767, 9.70733e-06) (3928, 9.12319e-06) (4096, 8.58638e-06)  };
    \addlegendentry{GFWSVD}

    \nextgroupplot[
        xmode=log, ymode=log,
        title={\textbf{Generalization Error}},
        title style={yshift=-1.5ex},
        xlabel={Rank},
        ylabel={Fisher error $\Delta F$},
        xtick={32,64,128,256,512},
        xticklabels={32,64,128,256,512},
    ]
    \addplot[color=colKFAC, mark=*, thick, solid, mark options={solid, draw=none}] coordinates { (32, 0.0003) (64, 0.0003) (128, 0.0002) (256, 0.0002) (512, 0.0001)  };
    \addplot[color=colCoupled, mark=square*, thick, solid, mark options={solid, draw=none}] coordinates { (32, 0.0004) (64, 0.0004) (128, 0.0004) (256, 0.0003) (512, 0.0002)  };
    \end{groupplot}
\end{tikzpicture}
}
    \caption{Spectral decay of Lanczos Fisher approximation (\gls{gfwsvd}) compared to our \acrshort{kfac} approximation (left) and resulting fisher error on unseen gradients (right), showing lower spectral decay and better generalization error for \acrshort{kfac}.}
    \label{fig:factor_quality_estimate}
\end{figure}

\subsubsection{\gls{oursearch} Error Calibration}
We investigate the impact of the calibration point $c_\text{ref}$ used to estimate sensitivity during the search.\  \cref{fig:search_ablation_kl_vs_mse_and_cref} compares \gls{kl} divergence vs.\ \glslink{mse}{\acrshort{mse}}\glsunset{mse} across varying $c_\text{ref}$ on \gls{qwen3}-4B, indicating
that \gls{kl} divergence consistently outperforms \gls{mse} in zero shot accuracy; improving it up to $1.6$ \gls{pp}\ For perplexity, \gls{mse} performs slightly better than \gls{kl}, but with small margins. When evaluating different $c_\text{ref}$ values for the calibration, we observe that calibrating in the high-compression regime ($c_\text{ref}$=$0.1$) yields slightly lower perplexity (14.5) than calibrating in the low-compression one (14.7).
We hypothesize that severe constraints isolate the parameters most critical for preserving the data distribution. Hence, although $c_{\text{ref}}$=$0.1$ incurs a marginal zero-shot reduction relative to $c_{\text{ref}}$=$0.3$ (-0.9 \gls{pp}), we prioritize its better generative fidelity (perplexity) and adopt it as our default configuration, see Supplement~\ref{supp:sec:scale_investigations} for more details. %
\subsubsection{\gls{ours} Estimation Quality}\label{sec:fisher_ablation}

After demonstrating consistent gains in task performance of the \glslink{kfac}{\acrshort{kfac}}\glsunset{kfac} estimation of $\widehat{\mathbf{A}}$ and $\widehat{\mathbf{G}}$ over \gls{gfwsvd}'s estimation in \cref{tab:llm_main_table_short}, we further analyze the impact of the \textbf{sample-induced rank bottleneck}. \cref{fig:factor_quality_estimate} (left) displays the normalized eigenvalue decay of the induced Kronecker metric, $\widehat{\mathbf{A}}\otimes\widehat{\mathbf{G}}$, for both approaches, demonstrating that \gls{gfwsvd} suffers from severe \textbf{spectral collapse}, where the eigenspectrum decays rapidly. %
As a result, truncation quickly discards the %
information curvature structure, thereby limiting generalization. This is quantified by measuring the Fisher degradation, $\Delta F$, on held-out data across truncation ranks, indicating how well the estimator captures true loss curvature on unseen sequences (see Supplement~\ref{supp:sec:kfac_long_ablations} for details). \cref{fig:factor_quality_estimate} (right) confirms that our token-wise estimator consistently achieves lower $\Delta F$ across all truncation ranks. This supports the idea that preventing spectral collapse is essential for robust generalization, a finding further supported by effective rank analysis in Supplement~\ref{supp:sec:kfac_long_ablations}.

%% file: data/tables/search_approach_comparison_cr.tex
\begin{tabular}{cc| ccc c| ccc c| ccc c}
\toprule
\midrule
&\multirow{2}{*}{\begin{tabular}{@{}c@{}} \textbf{Search} \\ \textbf{Method} \end{tabular}}& \multicolumn{4}{c}{\gls{mistral}-7B} & \multicolumn{4}{c}{\gls{llama3}-8B} & \multicolumn{4}{c}{\gls{qwen3}-8B} \\
\cmidrule(lr){3-6} \cmidrule(lr){7-10} \cmidrule(lr){11-14}
&  & \glslink{wiki}{Wiki}\glsunset{wiki} $\downarrow$ & \glslink{c4}{C4}\glsunset{c4} $\downarrow$ & \glslink{ptb}{PTB}\glsunset{ptb} $\downarrow$ & Acc $\uparrow$ & \glslink{wiki}{Wiki}\glsunset{wiki} $\downarrow$ & \gls{c4} $\downarrow$ & \gls{ptb} $\downarrow$ & Acc $\uparrow$ & \glslink{wiki}{Wiki}\glsunset{wiki} $\downarrow$ & \gls{c4} $\downarrow$ & \gls{ptb} $\downarrow$ & Acc $\uparrow$ \\
\midrule
& Baseline & 5.25 & 8.10 & 27.72 & 63.95 & 6.14 & 9.47 & 9.90 & 63.34 & 9.71 & 15.52 & 15.43 & 62.03 \\
\midrule
\multirow{7}{*}{0.8} &  \textbf{uniform} & 7.08 & 18.58 & 57.22 & 52.67 & 10.61 & 41.94 & 60.91 & 50.84 & 12.25 & 30.63 & 41.70 & 54.18 \\
 &  \gls{asvd} & \underline{6.98} & 19.66 & 64.84 & 51.41 & 12.02 & 59.27 & 65.86 & 47.28 & 13.28 & 36.16 & 46.52 & 50.74 \\
 &  \gls{svdllmv2} & 7.06 & 18.58 & 56.77 & 52.77 & 10.61 & 41.54 & 60.20 & 51.05 & 12.28 & 30.51 & 41.46 & 54.39 \\
&  \gls{mrcs} & 7.06 & 18.32 & 58.12 & 52.96 & 10.54 & 41.29 & 59.27 & 51.01 & 14.24 & 34.23 & 47.34 & 53.16 \\
  &  \gls{ars} & 7.32 & \underline{11.76} & \underline{44.65} & \underline{53.68} & 11.33 & \underline{19.62} & \textbf{18.43} & 49.58 & \underline{11.05} & \underline{20.09} & \underline{19.89} & \underline{56.08} \\
 &  \gls{atp} & 7.06 & 18.15 & 58.35 & 52.75 & \underline{10.28} & 35.11 & 60.20 & \underline{52.10} & 12.25 & 30.63 & 41.70 & 54.18 \\
 &  \midrowcolor{m}\textbf{\gls{oursearch}} & \textbf{5.92} & \textbf{11.11} & \textbf{34.91} & \textbf{57.61} & \textbf{8.04} & \textbf{16.91} & \underline{20.56} & \textbf{56.77} & \textbf{10.28} & \textbf{17.93} & \textbf{18.22} & \textbf{59.03} \\
\midrule
\multirow{7}{*}{0.6} &  \textbf{uniform} & 14.47 & 67.16 & 181.84 & 39.22 & 33.97 & 325.43 & 536.55 & 36.24 & \underline{18.00} & 77.30 & 151.34 & 41.16 \\
 &  \gls{asvd} & 16.33 & 91.43 & 268.74 & 36.35 & 47.99 & 446.55 & 870.90 & 34.73 & 23.25 & 97.33 & 184.70 & 39.88 \\
 &  \gls{svdllmv2} & 14.50 & 67.16 & 183.27 & 39.37 & 33.25 & 317.89 & 532.37 & 36.32 & 18.04 & 76.40 & 150.16 & 41.23 \\
 & \gls{mrcs} & \underline{14.41} & 70.93 & 222.79 & 39.58 & 36.58 & 315.42 & 520.04 & 35.76 & 27.13 & 109.86 & 224.54 & 40.26 \\
  &  \gls{ars} & 18.47 & \textbf{33.05} & \underline{176.24} & 40.66 & 23.62 & \textbf{44.39} & \textbf{64.08} & \underline{42.05} & 19.47 & \textbf{53.65} & \textbf{57.89} & \underline{42.60} \\
 &  \gls{atp} & 15.73 & 59.73 & 188.35 & \underline{41.13} & \underline{21.09} & 129.95 & 300.98 & 41.96 & \underline{18.00} & 77.30 & 151.34 & 41.16 \\
& \midrowcolor{m}\textbf{\gls{oursearch}} & \textbf{10.32} & \underline{35.18} & \textbf{112.47} & \textbf{43.84} & \textbf{16.08} & \underline{63.83} & \underline{140.51} & \textbf{44.19} & \textbf{15.07} & \underline{54.39} & \underline{112.91} & \textbf{45.60} \\
\midrule
\bottomrule
\end{tabular}

%% file: data/tables/llm_main_table_cr.tex
\begin{tabular}{cc| ccc c| ccc c| ccc c}
\toprule
\midrule
&\multirow{2}{*}{\begin{tabular}{@{}c@{}} \textbf{SVD} \\ \textbf{Method} \end{tabular}}& \multicolumn{4}{c}{Mistral-7B} & \multicolumn{4}{c}{\gls{llama3}-8B} & \multicolumn{4}{c}{\gls{qwen3}-8B} \\
\cmidrule(lr){3-6} \cmidrule(lr){7-10} \cmidrule(lr){11-14}
& & \glslink{wiki}{Wiki}\glsunset{wiki} $\downarrow$ & \glslink{c4}{C4}\glsunset{c4} $\downarrow$ & \glslink{ptb}{PTB}\glsunset{ptb} $\downarrow$ & Acc$\uparrow$ & \glslink{wiki}{Wiki}\glsunset{wiki} $\downarrow$ & \gls{c4} $\downarrow$ & \gls{ptb} $\downarrow$ & Acc $\uparrow$ & \glslink{wiki}{Wiki}\glsunset{wiki} $\downarrow$ & \gls{c4} $\downarrow$ & \gls{ptb} $\downarrow$ & Acc $\uparrow$ \\
\midrule
& Baseline & 5.25 & 8.10 & 27.72 & 63.96 & 6.14 & 9.47 & 9.90 & 63.33 & 9.71 & 15.52 & 15.43 & 62.04 \\
\midrule
\multirow{8}{*}{0.9} &  \gls{fwsvd} & 9.47 & 13.75 & 58.35 & 56.93 & 42.11 & 59.96 & 69.83 & 48.88 & 16.95 & 24.75 & 31.54 & 53.82 \\
 &  \gls{asvd} & 9.14 & 13.41 & 49.13 & \underline{57.30} & 65.09 & 77.60 & 207.67 & 47.89 & 20.36 & 29.98 & 36.37 & 51.19 \\
 &  \gls{svdllm} & 6.46 & 15.07 & 66.12 & 56.81 & 10.14 & 39.63 & 49.91 & 52.27 & 12.52 & 27.40 & 36.66 & 55.96 \\
 &  \gls{svdllmv2} & 6.46 & 15.07 & 66.12 & 56.78 & 10.18 & 39.71 & 50.40 & 52.17 & 12.52 & 27.45 & 36.80 & 55.80 \\
 &  \gls{dobisvd} & 7.11 & 15.19 & 63.58 & 55.17 & 11.22 & 34.77 & 42.36 & 52.98 & 13.46 & 28.88 & 35.81 & 55.48 \\
 &  \gls{gfwsvd} & 31.66 & 65.09 & 305.72 & 40.73 & 2569.75 & 3827.63 & 8491.96 & 33.40 & 171.49 & 350.51 & 765.57 & 46.66 \\
 &  \midrowcolor{m}\textbf{\gls{ours}} & \underline{6.17} & \underline{12.79} & \underline{44.91} & 56.95 & \underline{8.49} & \underline{22.98} & \underline{26.15} & \underline{55.84} & \underline{11.27} & \underline{23.44} & \underline{25.69} & \underline{56.73} \\
& \midrowcolor{m}\textbf{$+$\gls{oursearch}} & \textbf{5.35} & \textbf{8.44} & \textbf{29.28} & \textbf{62.71} & \textbf{6.51} & \textbf{10.90} & \textbf{10.94} & \textbf{61.83} & \textbf{9.81} & \textbf{15.98} & \textbf{15.70} & \textbf{62.60} \\
\midrule
\multirow{8}{*}{0.7} &  \gls{fwsvd} & 34.84 & 50.30 & 215.10 & 44.24 & 716.39 & 837.54 & 1939.75 & 33.40 & 41.37 & 68.75 & 90.72 & 42.68 \\
 &  \gls{asvd} & 28.16 & 42.94 & 163.64 & 46.09 & 10989.41 & 9399.74 & 36032.89 & 33.03 & 70.66 & 96.95 & 131.49 & 43.10 \\
 &  \gls{svdllm} & 10.96 & 42.60 & 169.49 & 44.39 & 34.64 & 376.04 & 600.91 & 37.69 & 17.11 & 63.09 & 136.19 & 45.76 \\
 &  \gls{svdllmv2} & 10.96 & 42.60 & 169.49 & 44.35 & 34.98 & 378.99 & 615.16 & 37.70 & 17.15 & 63.33 & 136.72 & 45.62 \\
 &  \gls{dobisvd} & 12.37 & 39.71 & 170.82 & 42.94 & 31.54 & 500.12 & 223.67 & 38.48 & 18.54 & 63.83 & 94.34 & 46.24 \\
 &  \gls{gfwsvd} & 2549.75 & 3404.36 & 5355.81 & 32.29 & 41798.89 & 53670.83 & 72222.00 & 31.77 & 21684.98 & 27627.38 & 50814.52 & 34.06 \\
 &  \midrowcolor{m}\textbf{\gls{ours}} & \underline{9.18} & \underline{32.73} & \underline{98.86} & \underline{46.26} & \underline{16.23} & \underline{107.32} & \underline{180.42} & \underline{43.69} & \underline{13.97} & \underline{45.09} & \underline{72.33} & \underline{48.24} \\
& \midrowcolor{m}\textbf{$+$\gls{oursearch}} & \textbf{7.28} & \textbf{17.42} & \textbf{54.28} & \textbf{51.68} & \textbf{10.71} & \textbf{33.70} & \textbf{56.11} & \textbf{49.88} & \textbf{11.67} & \textbf{28.05} & \textbf{37.16} & \textbf{53.30} \\
\midrule
\bottomrule
\vspace{-0.55cm}
\end{tabular}

%% file: data/plots/generalization_plot_qwen_family_cr.tex
\definecolor{c4B}{HTML}{FF7F0E}
\definecolor{c8B}{HTML}{2CA02C}
\definecolor{c14B}{HTML}{D62728}
\definecolor{c32B}{HTML}{1F77B4}
\begin{tikzpicture}
    \begin{groupplot}[
        group style={
            group size=2 by 2,
            vertical sep=2cm,   %
            horizontal sep=1.5cm,
        },
        width=5cm,
        height=5cm,
        xlabel={Compression Ratio},
        grid=major,
        grid style={dashed, gray!30},
        xmin=0.45, xmax=0.95,  %
        x dir=reverse,       %
        ylabel style={yshift=-1.3em},
        every mark/.append style=solid,
        ]
    \nextgroupplot[ymode=log, ymin=7, ymax=250, ylabel style={yshift=+1.3em}, ylabel={Perplexity ($\downarrow$)}, title={\gls{ours}},
    xlabel={},
    ymin=0, title style={yshift=-1.5ex}, ymax=70]
        \addplot[color=c4B, mark=triangle*, mark options={solid}, thick] coordinates { (0.50, 51.89) (0.60, 28.33) (0.70, 19.85) (0.80, 17.18) (0.90, 15.98)  };
        \addplot[color=c4B, mark=triangle*, mark options={solid}, dashed, thick] coordinates { (0.50, 30.57) (0.60, 20.20) (0.70, 16.17) (0.80, 14.47) (0.90, 13.97)  };
        \addplot[color=c8B, mark=diamond*, mark options={solid}, thick] coordinates { (0.50, 27.94) (0.60, 18.00) (0.70, 13.97) (0.80, 12.25) (0.90, 11.27)  };
        \addplot[color=c8B, mark=diamond*, mark options={solid}, dashed, thick] coordinates { (0.50, 22.15) (0.60, 15.07) (0.70, 11.67) (0.80, 10.28) (0.90, 9.81)  };
        \addplot[color=c14B, mark=square*, mark options={solid}, thick] coordinates { (0.50, 19.43) (0.60, 14.38) (0.70, 12.02) (0.80, 10.75) (0.90, 9.87)  };
        \addplot[color=c14B, mark=square*, mark options={solid}, dashed, thick] coordinates { (0.50, 16.43) (0.60, 12.13) (0.70, 10.12) (0.80, 9.16) (0.90, 8.74)  };
        \addplot[color=c32B, mark=*, mark options={solid}, thick] coordinates { (0.50, 15.40) (0.60, 11.53) (0.70, 9.77) (0.80, 8.84) (0.90, 8.26)  };
        \addplot[color=c32B, mark=*, mark options={solid}, dashed, thick] coordinates { (0.50, 11.58) (0.60, 9.27) (0.70, 8.32) (0.80, 7.90) (0.90, 7.64)  };
    \nextgroupplot[ylabel={Accuracy ($\uparrow$)}, title={\gls{ours}},
    xlabel={}, title style={yshift=-1.5ex}]
        \addplot[color=c4B, mark=triangle*, mark options={solid}, thick] coordinates { (0.50, 34.03) (0.60, 38.19) (0.70, 43.09) (0.80, 48.42) (0.90, 50.79)  };
        \addplot[color=c4B, mark=triangle*, mark options={solid}, dashed, thick] coordinates { (0.50, 38.15) (0.60, 42.80) (0.70, 49.27) (0.80, 55.05) (0.90, 58.52)  };
        \addplot[color=c8B, mark=diamond*, mark options={solid}, thick] coordinates { (0.50, 35.91) (0.60, 41.16) (0.70, 48.24) (0.80, 54.18) (0.90, 56.73)  };
        \addplot[color=c8B, mark=diamond*, mark options={solid}, dashed, thick] coordinates { (0.50, 39.63) (0.60, 45.60) (0.70, 53.30) (0.80, 59.03) (0.90, 62.60)  };
        \addplot[color=c14B, mark=square*, mark options={solid}, thick] coordinates { (0.50, 39.48) (0.60, 46.97) (0.70, 53.13) (0.80, 57.02) (0.90, 60.49)  };
        \addplot[color=c14B, mark=square*, mark options={solid}, dashed, thick] coordinates { (0.50, 42.57) (0.60, 50.85) (0.70, 58.03) (0.80, 62.61) (0.90, 65.05)  };
        \addplot[color=c32B, mark=*, mark options={solid}, thick] coordinates { (0.50, 45.47) (0.60, 52.69) (0.70, 55.94) (0.80, 42.69) (0.90, 43.58)  };
        \addplot[color=c32B, mark=*, mark options={solid}, dashed, thick] coordinates { (0.50, 53.73) (0.60, 60.58) (0.70, 64.00) (0.80, 65.10) (0.90, 65.48)  };
    \nextgroupplot[ymode=log, ymin=7, ymax=250, ylabel style={yshift=+1.3em}, ylabel={Perplexity ($\downarrow$)}, title={SVD-LLM}, ymin=0, title style={yshift=-1.5ex}, ymax=70]
        \addplot[color=c4B, mark=triangle*, mark options={solid}, thick] coordinates { (0.50, 159.22) (0.60, 59.27) (0.70, 28.72) (0.80, 19.93) (0.90, 17.15)  };
        \addplot[color=c4B, mark=triangle*, mark options={solid}, dashed, thick] coordinates { (0.50, 33.57) (0.60, 22.45) (0.70, 17.11) (0.80, 15.13) (0.90, 14.02)  };
        \addplot[color=c8B, mark=diamond*, mark options={solid}, thick] coordinates { (0.50, 65.60) (0.60, 28.77) (0.70, 17.11) (0.80, 14.05) (0.90, 12.52)  };
        \addplot[color=c8B, mark=diamond*, mark options={solid}, dashed, thick] coordinates { (0.50, 31.66) (0.60, 17.28) (0.70, 12.37) (0.80, 10.50) (0.90, 9.83)  };
        \addplot[color=c14B, mark=square*, mark options={solid}, thick] coordinates { (0.50, 31.85) (0.60, 19.81) (0.70, 14.75) (0.80, 12.16) (0.90, 11.07)  };
        \addplot[color=c14B, mark=square*, mark options={solid}, dashed, thick] coordinates { (0.50, 18.32) (0.60, 13.35) (0.70, 10.40) (0.80, 9.32) (0.90, 8.77)  };
        \addplot[color=c32B, mark=*, mark options={solid}, thick] coordinates { (0.50, 25.79) (0.60, 15.43) (0.70, 11.65) (0.80, 10.22) (0.90, 9.23)  };
        \addplot[color=c32B, mark=*, mark options={solid}, dashed, thick] coordinates { (0.50, 11.88) (0.60, 9.36) (0.70, 8.34) (0.80, 7.90) (0.90, 7.64)  };
    \nextgroupplot[ylabel={Accuracy ($\uparrow$)}, title={SVD-LLM}, title style={yshift=-1.5ex}]
        \addplot[color=c4B, mark=triangle*, mark options={solid}, thick] coordinates { (0.50, 32.19) (0.60, 33.87) (0.70, 38.53) (0.80, 45.76) (0.90, 49.28)  };
        \addplot[color=c4B, mark=triangle*, mark options={solid}, dashed, thick] coordinates { (0.50, 38.28) (0.60, 43.53) (0.70, 48.91) (0.80, 54.29) (0.90, 57.56)  };
        \addplot[color=c8B, mark=diamond*, mark options={solid}, thick] coordinates { (0.50, 33.58) (0.60, 36.81) (0.70, 45.76) (0.80, 52.31) (0.90, 55.96)  };
        \addplot[color=c8B, mark=diamond*, mark options={solid}, dashed, thick] coordinates { (0.50, 40.70) (0.60, 46.46) (0.70, 54.75) (0.80, 60.04) (0.90, 62.36)  };
        \addplot[color=c14B, mark=square*, mark options={solid}, thick] coordinates { (0.50, 36.30) (0.60, 43.29) (0.70, 52.49) (0.80, 56.54) (0.90, 58.82)  };
        \addplot[color=c14B, mark=square*, mark options={solid}, dashed, thick] coordinates { (0.50, 42.72) (0.60, 51.12) (0.70, 58.32) (0.80, 62.85) (0.90, 65.28)  };
        \addplot[color=c32B, mark=*, mark options={solid}, thick] coordinates { (0.50, 39.37) (0.60, 48.60) (0.70, 54.78) (0.80, 57.87) (0.90, 43.23)  };
        \addplot[color=c32B, mark=*, mark options={solid}, dashed, thick] coordinates { (0.50, 54.59) (0.60, 60.37) (0.70, 64.51) (0.80, 65.85) (0.90, 65.61)  };
    \end{groupplot}

    \node[anchor=center] at ($(group c1r1.south)!0.5!(group c2r2.north)$) {
        \begin{tikzpicture}
            \matrix[draw, inner sep=3pt, column sep=2pt, ampersand replacement=\&] {
                \node[label=right:\textbf{LEMS}] {};    \& \draw[thick, dashed] (0,0) -- (0.5,0); \&
                \node[label=right:Uniform] {}; \& \draw[thick] (0,0) -- (0.5,0); \&

                \node[minimum width=0.2cm] {}; \&

                \node[label=right:4B] {};   \& \draw[c4B, mark=triangle*, only marks] plot coordinates {(0,0)}; \&
                \node[label=right:8B] {};   \& \draw[c8B, mark=diamond*, only marks] plot coordinates {(0,0)}; \&
                \node[label=right:14B] {};  \& \draw[c14B, mark=square*, only marks] plot coordinates {(0,0)}; \&
                \node[label=right:32B] {};  \& \draw[c32B, mark=*, only marks] plot coordinates {(0,0)}; \\
            };
        \end{tikzpicture}
    };
\end{tikzpicture}

%% file: data/plots/scale_ablation_bars_main_content_cr.tex
\pgfplotsset{
        compactplot/.style={
            ybar,
            width=0.53\linewidth,
            height=4.0cm,
            enlarge x limits=0.25,
            symbolic x coords={0.7, 0.5, 0.3, 0.1},
            xtick=data,
            xlabel={$c_{\text{ref}}$},
            xlabel style={font=\small, yshift=1ex},
            ylabel style={font=\scriptsize, align=center, yshift=-2ex},
            tick label style={font=\tiny},
            nodes near coords,
            nodes near coords style={font=\fontsize{3}{4}\selectfont, rotate=90, anchor=west},
            bar width=4pt,
            grid=major,
            title style={font=\bfseries\scriptsize, yshift=-0.5ex},
        }
    }
    \definecolor{green}{HTML}{2CA02C}
    \hspace{-0.3cm}
    \begin{tikzpicture}
        \begin{axis}[
            compactplot,
            title={Perplexity ($\downarrow$)},
            ymin=14.0, ymax=15.32, %
            legend columns=-1, %
            legend style={
                at={(0.5,0.95)}, %
                anchor=north,
                font=\tiny,
                draw=none,
                fill=white, %
                fill opacity=0.8,
                inner sep=1pt
            }
        ]
            \addplot[fill=icmlblue] coordinates {(0.7, 14.69) (0.5, 14.69) (0.3, 14.61) (0.1, 14.47)};
            \addplot[fill=DarkYellow] coordinates {(0.7, 14.41) (0.5, 14.47) (0.3, 14.35) (0.1, 14.38)};

            \legend{KL, MSE}
        \end{axis}
    \end{tikzpicture}%
    \hspace{+0.3cm} %
    \begin{tikzpicture}
        \begin{axis}[
            compactplot,
            title={Accuracy ($\uparrow$)},
            ymin=53.0, ymax=56.5,
            ytick={53, 54, 55, 56},
        ]
            \addplot[fill=icmlblue] coordinates {(0.7, 55.41) (0.5, 55.61) (0.3, 55.98) (0.1, 55.05)};
            \addplot[fill=DarkYellow] coordinates {(0.7, 55.01) (0.5, 54.89) (0.3, 54.37) (0.1, 53.83)};
        \end{axis}
    \end{tikzpicture}

%% file: data/tables/search_parameter_comparison_cr.tex
\begin{table}[b]
    \centering
    \caption{\gls{wiki} performance and added execution time of individual parts of our method on \gls{llama3}-8B and \gls{qwen3}-8B.}
    \footnotesize
    \resizebox{0.98\linewidth}{!}{
    \begin{tabular}{lcccccc}
    \toprule
    \midrule
      \multirow{2}{*}{\textbf{\gls{oursearch}} components} & \multicolumn{2}{c}{\gls{llama3}-8B} & \multicolumn{2}{c}{\gls{qwen3}-8B}  \\
                           & \glslink{wiki}{Wiki}\glsunset{wiki} $\downarrow$ & time $\downarrow$ &  \glslink{wiki}{Wiki}\glsunset{wiki} $\downarrow$ & time $\downarrow$\\
    \midrule
    \gls{svd} (uniform)          & 33.97            & \textbf{start}& 18.00           & \textbf{start}\\
    + \gls{ilp} search (energy)  & 65.70             & +1 min        & 32.57           & +1 min\\
    + \gls{kl} calibration       & 27.56             & +17 min       & 15.58           & +22 min\\
    + cross layer bias     & \textbf{16.08}   & +2 min        & \textbf{15.07}  & +2 min\\
    \midrule
    search time (w/wo cache)& \multicolumn{2}{c}{\textbf{3 min} / 20 min}  &    \multicolumn{2}{c}{\textbf{3 min} / 25 min}\\

    \midrule
    \bottomrule
    \end{tabular}
    }
    \label{tab:incremental_methods}

\end{table}

%% file: content/06_conclusion.tex
In this work, to enable effective rank allocation for \glspl{llm}, we propose \textbf{\gls{oursearch}}, a rank allocator driven by a calibrated layer-wise error model of spectral shape, scale, and bias. \gls{oursearch} demonstrates consistent gains independent of model size and across activation- and Fisher-based \gls{svd}. Additionally, we addressed the rank bottleneck of prior Fisher-based \gls{svd} by introducing \textbf{\gls{ours}}, which estimates the Fisher matrix with a token-wise approximation, improving decomposition quality.
While improving Fisher-\gls{svd}, our proposed \gls{ours} approach still relies on gradient computation, limiting computational efficiency. %
Moreover, \gls{oursearch} relies on a one-time layer-wise calibration step, which scales linearly with model depth.
Future work might focus on formalizing analytical bounds on error propagation to further improve modeling accuracy and extending the approach to \glslink{vlm}{vision-language models (VLMs)}\glsunset{vlm} and \glslink{moe}{mixture-of-experts (MoE) models}\glsunset{moe}.

%% file: content/07_supplementary.tex
\section{Extended Experimental Setup}\label{sup:experiment_setup_long}

\paragraph{Models Evaluated.}
We evaluate the proposed \glslink{oursearch}{layer-wise error model search (LEMS)}\glsunset{oursearch} and \gls{ours} compression method across a diverse set of modern open-weights \glslink{llm}{large language models (LLMs)}\glsunset{llm}, covering different parameter scales and architectural families. All models were sourced from HuggingFace, utilizing optimized implementations from Unsloth where available. For the \textbf{\glslink{qwen3}{Qwen3}\glsunset{qwen3} Family}, we utilize the official checkpoints for the entire scaling suite, specifically \texttt{Qwen/Qwen3-32B}, \texttt{-14B}, \texttt{-8B}, \texttt{-4B}, and \texttt{-1.7B}~\cite{yang2025qwen3}. For the \textbf{\glslink{llama3}{Llama 3}\glsunset{llama3} Family}, we evaluate \gls{llama3} , the very large \gls{llama3}.1 (70B) and the lightweight \gls{llama3}.2 variants (1B, 3B)~\cite{grattafiori2024llama3}, alongside the established \glslink{llama2}{Llama 2}\glsunset{llama2}-7B~\cite{touvron2023llama2}, using the unrestricted Unsloth-optimized versions (e.g., \texttt{unsloth/llama-3-8b}). Finally, for \textbf{\glslink{mistral}{Mistral}\glsunset{mistral}}, we evaluate the \gls{mistral}-7B-v0.1 model~\cite{jiang2023mistral7b} via \texttt{unsloth/mistral-7b}.

\paragraph{Search Baselines and Implementation Details.}
To ensure a fair comparison, all search methods utilize the same decomposition approach and candidate rank sets where applicable. We implemented or adapted the baselines as follows:
\begin{itemize}
    \item \textbf{\gls{asvd}} \cite{yuan2024asvd}: We utilize the official public codebase with default settings as reference, where the search evaluates sensitivity on the calibration set to select ranks from a discrete set of multipliers.
    \item \textbf{\gls{mrcs}} \cite{azizi2024memoryViT}: As no official code was available, we reimplemented the algorithm following the published methodology. Following their setup, we set the sensitivity parameter $\gamma=80$ and limit the search to 500 iterations. To maintain a valid search space comparable to other methods, we lower-bound the compression ratio per layer at $c_i=0.3$.
    \item  \textbf{\gls{svdllmv2}} \cite{2025-svd-llmv2}: We re-implemented the search algorithm based on the paper description, as the code remains unavailable. %
    \item \textbf{\gls{ars}} \cite{gao2024adaptive}: We used the public \gls{ars} implementation (\url{https://github.com/sidhantls/adaptive-rank-selection-svd}) with three crucial modifications to ensure fair benchmarking. First, we initialize the model with our pre-computed \glslink{svd}{singular value decomposition (SVD)}\glsunset{svd} factors rather than using their decomposition implementation to isolate the benefit of the search algorithm. Second, we aligned their parameter counting and compression ratio logic with our framework to ensure all methods target the exact same parameter budget. Third, we adjusted the training schedule, because the default values resulted in severe overcompression. With the adjustment, once the ratio is within 0.2\% of the target retention, the compression loss is disabled and the alignment loss is scaled by 0.1. For hyperparameters, we used the default settings for \gls{llama3}-8B across all models ($\gamma=1$ for \gls{mistral}-7B at 0.6). In our tests, \gls{ars} typically converges within $\sim200$ iterations using about 7$\times$ the data of \gls{asvd} and \gls{oursearch}.
    \item \textbf{\gls{atp}} \cite{atp_huang2025determining}: We execute the grid search procedure as specified in the original paper. We perform 10 grid-search steps over the valid linearly  space and report the best performing configuration.
\end{itemize}
\textbf{\gls{oursearch} (Proposed):} For our method, we use the Optuna framework~\cite{Akiba_Optuna_A_next_generation_2019} to fit the bias parameters. To maintain computational parity with \gls{atp}, we limit the optimization to 10 trials. The search space for the bias magnitude is bounded to $\kappa \in [0.0, 3.0]$ and the decay shape to $\gamma \in (0.0, 7.0)$. For the scale calibration, we measure the \gls{kl} sensitivity at a reference compression of $c_{\text{ref}}=0.1$, as the signal-to-noise ratio of the relative importance is larger under strong compression. Finally, we constrain the layer-wise compression range to match \gls{asvd}. Additional details on our methods design choices and the algorithm are provided in the supplementary~\cref{supp:sec:in_depth_search_ablations} and \ref{supp:sec:algorithms}, respectively.
\section{Additional Ablations}

\subsection{Extended Explanations and Ablations for \gls{oursearch}}\label{supp:sec:in_depth_search_ablations}
This section provides empirical validation for the design choices of our \gls{oursearch} algorithm. We first investigate the necessity of the bias term by quantifying error propagation in Supplement~\ref{sup:search_bias_ablations}. We then analyze the scale component in Supplement~\ref{supp:sec:scale_investigations}, comparing different importance metrics and calibration points to identify the most robust signal for global rank allocation. Next, we provide a detailed comparison between \gls{oursearch}, \gls{atp}, and \gls{asvd} in \cref{sup:search_detailed_comparison_atp_asvd}. Finally, in \cref{supp:sec:global_optimization_vs_bias_calibration} we discuss how the individual components of \gls{oursearch} contribute to its performance and how the combination of the components allows surpassing prior works.

\subsubsection{Bias Function in Error Modeling}\label{sup:search_bias_ablations}
\paragraph{Theoretical Justification.}
The necessity of a depth-dependent bias term is grounded in the theoretical analysis of error propagation. Recent work by \citet{atp_huang2025determining} formalizes the concept of \textit{Reconstruction Error Explosion} (Theorem 3.2), proving that in a simplified linear network, increasing the error in layer $i$ strictly increases the \textbf{lower bound} of the error in layer $i+1$:
\begin{equation}
\mathcal{L}(\boldsymbol{W}_{i+1}, \boldsymbol{X}_{i+1}) > \sigma_{\min}^2(\widetilde{\boldsymbol{W}}_{i+1}) \cdot \mathcal{L}(\boldsymbol{W}_{i}, \boldsymbol{X}_{i}).
\end{equation}
This inequality implies a cascade effect where early-layer errors are amplified by the singular values of all subsequent layers.
However, this theoretical bound assumes a strictly linear feed-forward structure. Modern \glspl{llm} introduce complex non-linearities (e.g., SwiGLU), normalizations (RMSNorm), and crucial residual connections that dampen or distort this propagation in ways that are analytically highly difficult to model precisely.
Consequently, while theory confirms that an error penalty \emph{must} exist and should be higher for initial  layers, deriving its exact analytical form is unfeasible for Transformers. We therefore adopt a hybrid approach: we accept the theoretical mandate for a depth-dependent penalty but determine its specific \emph{shape} and \emph{magnitude} empirically. Our bias term $\beta(d_i)$ is thus a parameterized correction that adapts the theoretical \textit{error explosion} principle to the empirical reality of residual-heavy \gls{llm}  architectures.

\begin{figure}[ht]
    \begin{minipage}[t]{0.49\textwidth}
        \input{data/plots/supp_search/structural_error_underestimation_llama32_3B_05}
    \end{minipage}
    \begin{minipage}[t]{0.49\textwidth}
        \input{data/plots/supp_search/structural_error_underestimation_Qwen3_4B_05}
    \end{minipage}
    \caption{Systematic error underestimation. The \textcolor{DarkYellow}{yellow bars} (sum of individual errors) consistently fall below the \textcolor{icmlblue}{blue bars} (measured total error), confirming that cross-layer interactions amplify degradation. This gap motivates the bias correction $\beta(d_i)$.}\label{fig:search_measured_bias}
\end{figure}
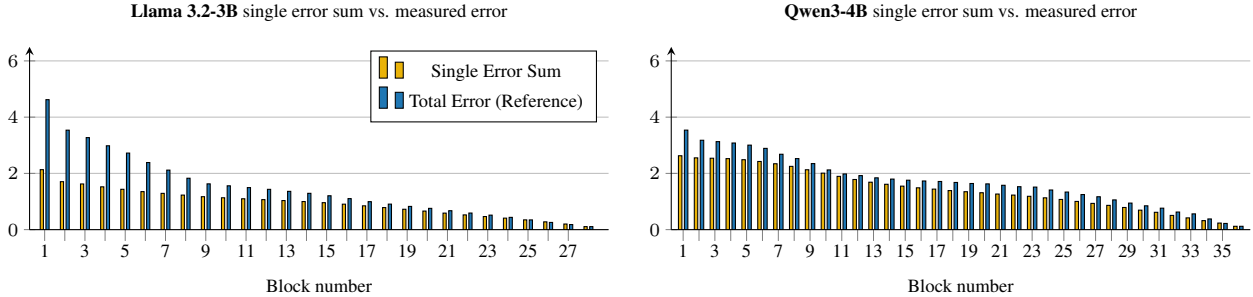

\paragraph{Quantifying Error Underestimation.}
To verify that this theoretical error accumulation manifests in practice, despite the mitigating effects of residual connections and norm layers, we perform a sequential compression experiment.
We first measure the error induced by compressing each block $d_i \in \{1, ...,B\}$ individually. We then compress the model cumulatively from the last block to the first, measuring the actual, total degradation at each step.
\cref{fig:search_measured_bias} compares the naive sum of individual errors against the measured total error. The results confirm a systematic underestimation: as more blocks are compressed, the true error diverges from the additive proxy. This divergence is particularly pronounced in the \gls{llama3}.2-3B model, where early perturbations amplify significantly as they propagate. In \gls{qwen3}-4B, while the additive approximation holds better, the underestimation remains present. %
This confirms that simply minimizing local errors is insufficient. Hence, we find that the optimization objective needs to be corrected with a depth-dependent bias $\beta(d_i)$ to approximate the true upper bound of the loss.

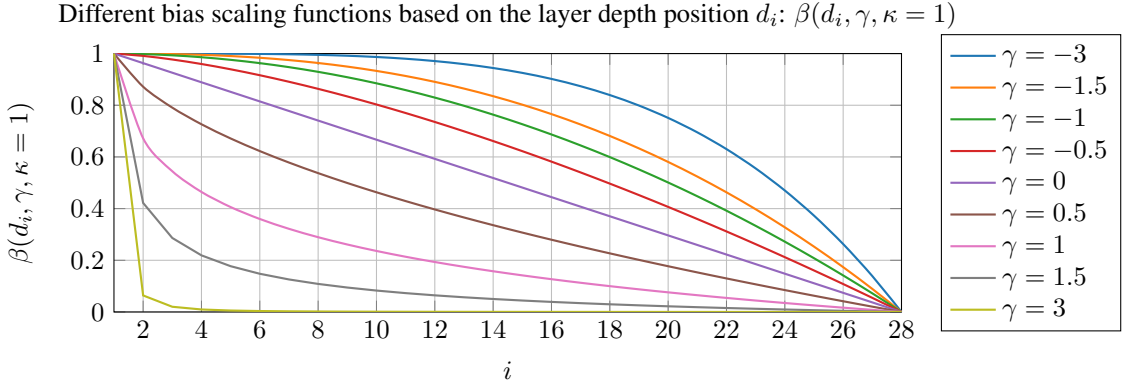
\begin{figure}[t]
    \input{data/plots/supp_search/harmonic_scale}
    \vspace{-0.5em}
    \caption{Candidate bias functions $\beta(d_i, \gamma, \kappa=1)$. We test various decay profiles, from linear ($\gamma=0$) to strictly exponential ($\gamma=\pm3$), to model the error propagation penalty.}
    \label{fig:search_bias_functions}
\end{figure}

\begin{figure}[t]
  \begin{minipage}[t]{0.45\textwidth}
    \input{data/plots/supp_search/grid_search_Llama32_3B_05}
  \end{minipage}
  \begin{minipage}[t]{0.54\textwidth}
    \input{data/plots/supp_search/grid_search_Qwen3_4B_05}
  \end{minipage}\vspace{-1.5em}
  \caption{Grid Search for bias parameters $\kappa$ and $\gamma$. Scatter plots show perplexity (lower is better) for varying magnitude $\kappa$ and decay $\gamma$ for \gls{llama3}.2-3B and \gls{qwen3}-4B. The improvements observed across both models support the benefit of assigning higher importance to earlier layers.}\label{fig:search_parameter_grid}
\end{figure}
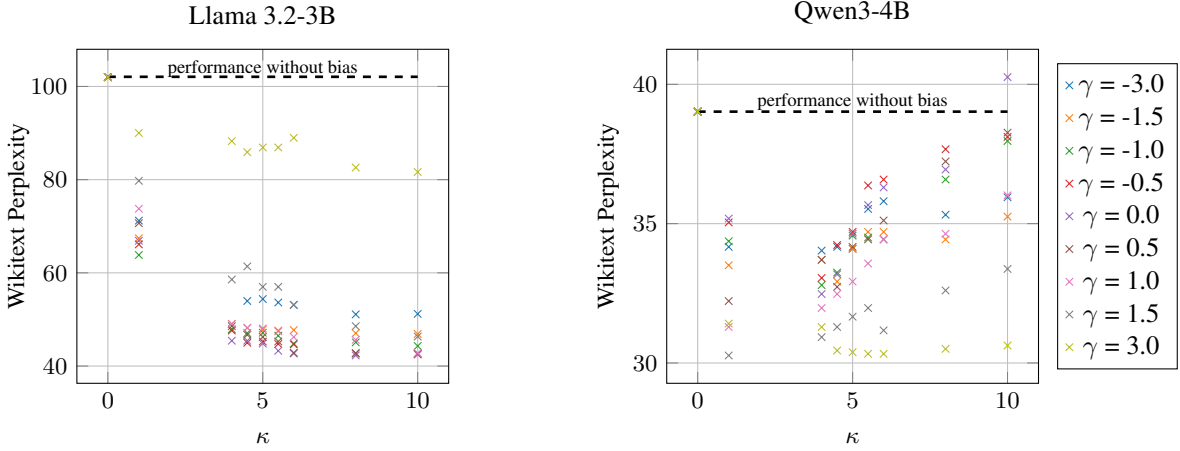

\paragraph{Modeling the Bias Function.}
We model the bias $\beta(d_i)$ as a depth-dependent penalty that decreases for later blocks, reflecting the observation that errors introduced early in the network have more opportunity to propagate and amplify~\cite{atp_huang2025determining}. We parameterize $\beta(d_i)$ using a harmonic series controlled by a magnitude scalar $\kappa$ and a decay shape $\gamma$:
\begin{equation}\label{eq:detailed_bias_function}
\beta(d_i, \kappa, \gamma) = \kappa \cdot \frac{\sum^{B}_{k=d_i} k^{-\gamma}}{\sum^{B}_{k=1} k^{-\gamma}}
\end{equation}
\cref{fig:search_bias_functions} illustrates some shapes tested, ranging from steep decay ($\gamma=3$, heavily penalizing the early blocks) to linear decay ($\gamma=0$). The scalar $\kappa$ ranges from $0$ (no bias) to $10$.
\cref{fig:search_parameter_grid} presents the ablation results. The impact of the bias correction is significant: for \gls{llama3}.2-3B, the best configuration ($\kappa=8, \gamma=0$) achieves a perplexity of \textbf{42.27}, whereas a poor choice yields \textbf{90.01}. Crucially, \emph{any} bias correction consistently outperforms the uncorrected baseline. We observe similar trends for \gls{qwen3}-4B, where the optimal setting ($\kappa=6, \gamma=3$) yields a perplexity of 30.32 compared to $>40$ for the uncorrected search. These results highlight that accounting for error accumulation is not merely a theoretical refinement but a prerequisite for stability in high-compression regimes.


\paragraph{Determining Bias Parameters in Practice.}
Since the exact magnitude of error propagation depends on architecture depth and residual structure, the optimal bias parameters, $\kappa$ and $\gamma$, cannot be fixed universally. To determine them efficiently, we implement a lightweight hyperparameter search using the \gls{tpe} algorithm provided by Optuna \cite{Akiba_Optuna_A_next_generation_2019}.
The process is formulated as a bilevel optimization: in the outer loop, the \gls{tpe} sampler suggests candidate values for $\kappa$ and $\gamma$. In the inner loop, we (1) update the cost function $L_i(c)$ with these parameters, (2) solve the \glslink{ilp}{integer linear program (ILP)}\glsunset{ilp} to obtain the optimal rank configuration, (3) temporarily compress the model to these ranks, and (4) measure the \gls{kl} divergence on a small calibration set.
Because the \gls{ilp} solver is extremely fast ($\approx1$s) and we utilize precomputed outputs for the uncompressed model, this entire loop is computationally inexpensive. We run 10 trials, which complete in minutes on a single GPU, yielding a fitted error propagation model that is calibrated to the specific target model and compression budget.

\paragraph{Discussion - Optimization formulation.}

We note that our interaction correction term, $\beta(d_i, \kappa, \gamma)$, admits two complementary interpretations: (i) as a calibrated reweighting of the per-module surrogate costs, and (ii) as an implicit mechanism for enforcing a depth-wise structural prior supported by theory and prior empirical findings.

A classical Lagrangian formulation would operationalize this explicitly by introducing a constraint on depth-weighted compression (or degradation) in early blocks and associating a multiplier $\lambda$ that quantifies the cost of violating this prior, formalizing a surrogate Lagrangian relaxation~\cite{bragin_surrogate-lagrangian_2015}. In practice, we instead fit $\kappa$ and $\gamma$ via a lightweight bilevel loop that directly optimizes the ground-truth proxy metric (\gls{kl} on a small calibration set). This approach is more robust in the discrete \gls{ilp} setting: small dual-ascent updates of $\kappa$ can induce abrupt jumps in the rank allocation, and hand-specifying a constraint threshold is model- and budget-dependent. Our \gls{tpe}-based calibration avoids these issues while operationalizing the same prior, selecting bias parameters that best align the \gls{ilp} solution with observed KL behavior.

Conceptually, the Lagrangian perspective remains useful: $\kappa$ can be interpreted as an implicit “shadow price” on early-block compression, and extending the method to additional depth-wise structural constraints (e.g., monotonicity or smoothness across blocks) naturally corresponds to introducing additional multipliers. Future work could explore surrogate- or level-based Lagrangian relaxations in the \gls{ilp} to tighten the relaxation and further improve rank allocation search efficiency and performance~\cite{bragin_surrogate-lagrangian_2015, Bessa_Dabert_Bourgeat_Rousseau_Cappart_2025}.

\subsubsection{Correct Cross-Layer Scaling (Importance)}\label{supp:sec:scale_investigations}
\paragraph{The Need for Calibrated Importance.}
To allocate ranks optimally, the search algorithm must weigh the cost of compressing an \glslink{mlp}{multi-layer perceptron (MLP)}\glsunset{mlp} projection layer in the first block of a model against an Attention projection layer near the model output. This requires a global importance metric. If the proxy objective wrongly assumes one layer-type is easier to compress than another, the \gls{ilp} solver will aggressively truncate the sensitive layer, degrading performance.

\paragraph{Challenges in Importance Estimation.}
Accurate sensitivity analysis is computationally expensive. \gls{llama3}-8B has 32 blocks with 7 linear layers each, which in turn have hundreds or even thousands of valid rank options, making the evaluation of every rank configuration intractable.
\begin{figure}[bp!]
    \centering
    \resizebox{1\linewidth}{!}{%
    \input{data/plots/supp_search/scale_metric_ablation_llama3_3b}
    }
    \vspace{-0.3cm} %

    \resizebox{1\linewidth}{!}{%
        \input{data/plots/supp_search/scale_metric_ablation_qwen3_4b}
    }
    \caption{Comparison between different sensitivity assessment metrics across 9 compression ratios for all linear layers of \gls{llama3}.2-3B (top) and \gls{qwen3}-4B (bottom) in block 15.}
    \label{fig:ablation_scale_metric}
\end{figure}
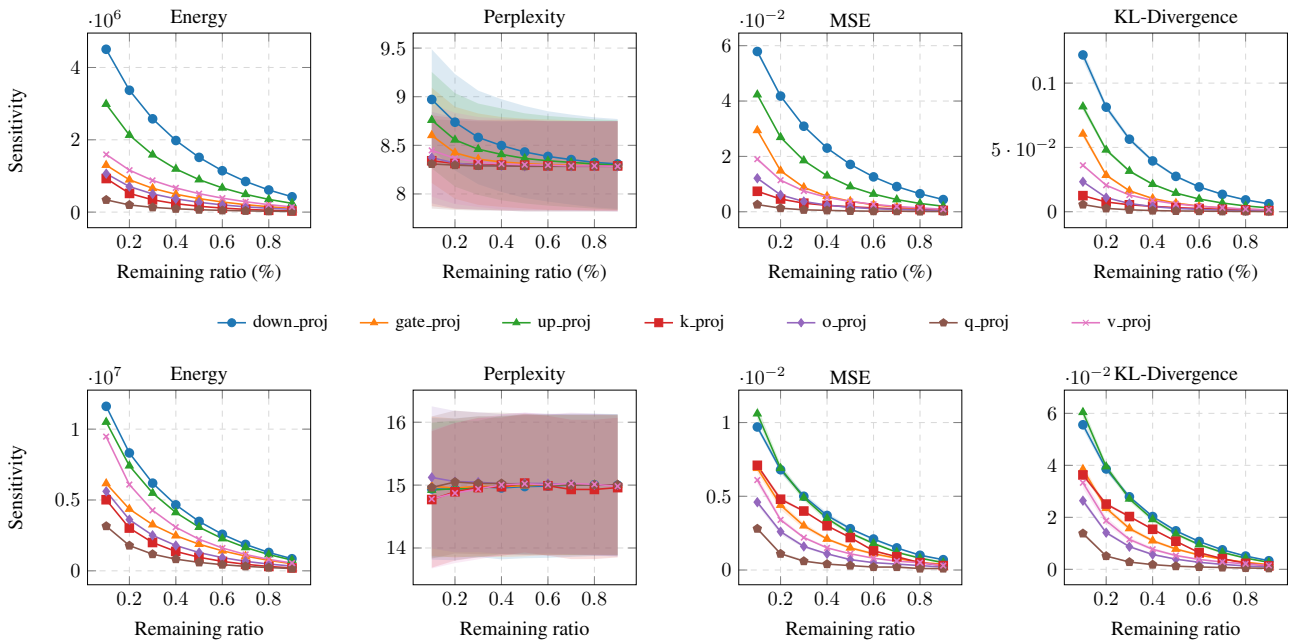
Existing solutions fall into two camps:
\begin{enumerate}
\item \textbf{Calibration-based} (e.g., \gls{asvd}~\cite{yuan2024asvd}): Measures empirical loss on a validation set. While accurate, it is costly (hours per model) and sparse (checking only a few discrete ranks), often missing the optimal operating point.
\item \textbf{Energy-based} (e.g., \gls{svdllmv2} and \gls{mrcs}~\cite{2025-svd-llmv2, azizi2024memoryViT}): Uses the Frobenius norm of discarded weights (or activations). This is computationally inexpensive and dense (available for every rank), but uncalibrated. A 10\% energy loss in layer $i$ might damage the output far more than a 10\% energy loss in layer $j$.
\end{enumerate}

Fisher-weighted objectives theoretically provide a more downstream (loss-aware) energy estimate. However, standard estimation only captures \emph{local} curvature, inherently assuming fixed inputs. Therefore, they fail to model how the \emph{distribution shift} caused by compressing layer $i$ affects the compression of layer $i+1$~\cite{atp_huang2025determining}. Moreover, the reliability of the estimation is limited by the approximation quality of the Fisher information~\cite{EmpiricalFisherLimitations} and may be subject to different scaling across layers due to different gradient magnitudes.
\paragraph{Selecting the Best Proxy Metric.}\label{supp:sec:sensitivity_metric_discussion}

We aim to combine the dense resolution of the energy metric (shape) with the global accuracy of a calibration metric (scale, $\alpha_i$). We first identify the most reliable metric for this calibration. \cref{fig:ablation_scale_metric} compares the sensitivity curves of perplexity, \glslink{mse}{mean squared error (MSE)}\glsunset{mse}, and \gls{kl}-divergence across five compression ratios and four seeds on \gls{llama3}.2-3B and \gls{llama3}-4B. The line represents the mean while the shaded area represents standard deviation. As can be seen, \textbf{perplexity} is excessively noisy. Confidence intervals overlap significantly, making it difficult to reliably rank layer importance. Differently, \textbf{\gls{mse} and \gls{kl} divergence} show stable, distinct trends that correctly differentiate layer sensitivities (e.g., correctly ranking \texttt{gate\_proj} as more sensitive than \texttt{o\_proj}).
We select \gls{kl}-divergence as it directly measures the distributional shift at the output, offering slightly better task connection than \gls{mse}, yielding in better perplexity and zero-shot accuracy results for our search tests (\cref{tab:search_ablation_cref}).

\paragraph{Optimal Calibration Point.}
Finally, we determine the compression ratio $c_\text{ref}$ at which to measure this scalar importance $\alpha_i$. \cref{tab:search_ablation_cref} shows that calibrating at stonger compression ratios $c_\text{ref}=0.1$ and $c_\text{ref}=0.3$ yields strong performance. Ultimately, we opt for using $c_\text{ref}=0.1$ (high compression), as its performance is close to 0.3 and we expect the degradation signal to be easier detectable for larger models.

\begin{table}[]
    \centering
    \caption{Evaluation of different $c_\text{ref}$ values for \gls{mse} and \gls{kl} on both \gls{llama3}.2-3B and \gls{qwen3}-4B.}
    \begin{tabular}{c|cc|cc|cc|cc}
    \toprule
    \midrule
       & \multicolumn{4}{c}{\gls{llama3}.2-3B} & \multicolumn{4}{c}{\gls{qwen3}-4B} \\
     $c_\text{ref}$ & \multicolumn{2}{c}{\glslink{wiki}{Wiki}\glsunset{wiki} $\downarrow$} & \multicolumn{2}{c}{Acc $\uparrow$} & \multicolumn{2}{c}{\glslink{wiki}{Wiki}\glsunset{wiki} $\downarrow$} & \multicolumn{2}{c}{Acc $\uparrow$} \\
    \cmidrule(lr){2-5} \cmidrule(lr){6-9}
       & \gls{kl} & \gls{mse} & \gls{kl} & \gls{mse} & \gls{kl} & \gls{mse} & \gls{kl} & \gls{mse} \\
    \midrule
      0.7 & 10.00 & 9.94            & 50.42 & 50.13             & 14.69 & 14.41 & 55.41 & 55.01 \\
      0.5 & 9.77 & 9.85            & 50.47 & 49.92             & 14.69 & 14.47 & \underline{55.61} & 54.89 \\
      0.3 & \textbf{9.52} & 9.73 & \textbf{51.04} & 49.93 & 14.61 & \textbf{14.35} & \textbf{55.98} & 54.37 \\
      0.1 & \underline{9.54} & 9.64     & 50.89 & \underline{51.03} & 14.47 & \underline{14.38} & 55.05 & 53.83 \\
    \midrule
    \bottomrule
    \end{tabular}
    \label{tab:search_ablation_cref}
\end{table}

\subsubsection{Detailed Comparison between \gls{oursearch}, \gls{atp} and \gls{asvd}} \label{sup:search_detailed_comparison_atp_asvd}
This section provides a detailed analysis of our pruning search strategy against two leading competitors, \gls{atp} and \gls{asvd}. %
To isolate the differences in their allocation strategies, we configure all methods to target 80\% parameter retention on the \gls{llama3}.2-3B model. For \gls{asvd}, we leverage our more accurate error model (without bias) to ensure the comparison focuses strictly on the search algorithm's efficacy.

\begin{figure}[tp!]
  \begin{minipage}[t]{0.65\textwidth}
    \input{data/plots/supp_search/relative_comp_layer_inportance_Llama32_3B_05}
  \end{minipage}
  \hfill
  \begin{minipage}[t]{0.3\textwidth}
    \input{data/plots/supp_search/relative_comp_layer_type_importance_Llama32_3B_05}
  \end{minipage}
  \begin{minipage}[t]{0.65\textwidth}
    \input{data/plots/supp_search/relative_comp_layer_inportance_Qwen3_4B_05}
  \end{minipage}
  \hfill
  \begin{minipage}[t]{0.3\textwidth}
    \input{data/plots/supp_search/relative_comp_layer_type_importance_Qwen3_4B_05}
  \end{minipage}
  \caption{Average relative importance of block positions and layer types in \gls{qwen3}-4B and \gls{llama3}.2-3B using \gls{kl} divergence as importance metric.}\label{fig:search_importance_comparisons}
\end{figure}
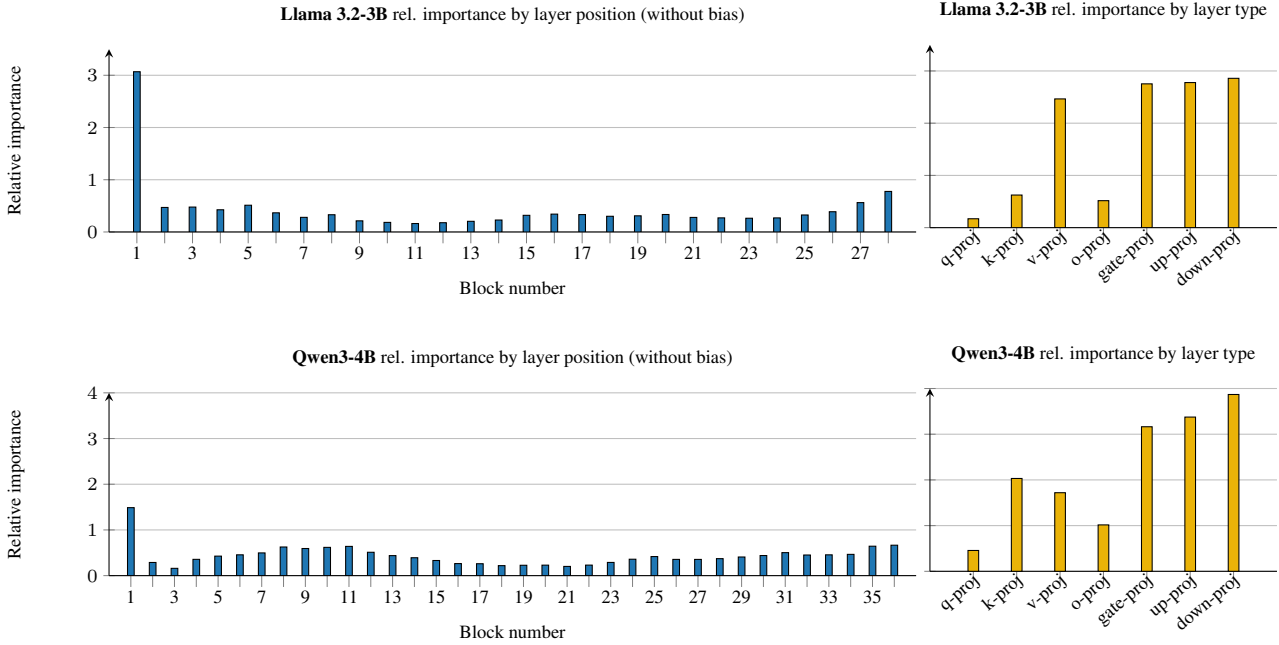

An effective compression strategy must account for the heterogeneous importance of different model components. As illustrated in ~\cref{fig:search_importance_comparisons}, the relative importance of layers in both \gls{llama3}.2-3B and \gls{qwen3}-4B is highly non-uniform. This variation exists across block positions/depth (\cref{fig:search_importance_comparisons}, left plots) and is especially pronounced across different layer types (\cref{fig:search_importance_comparisons}, right plots). For instance, the feed-forward layers (up-proj, down-proj) are substantially more critical than the attention projection layers (q-proj, k-proj).
The primary limitation of existing methods lies in how they address this heterogeneity.

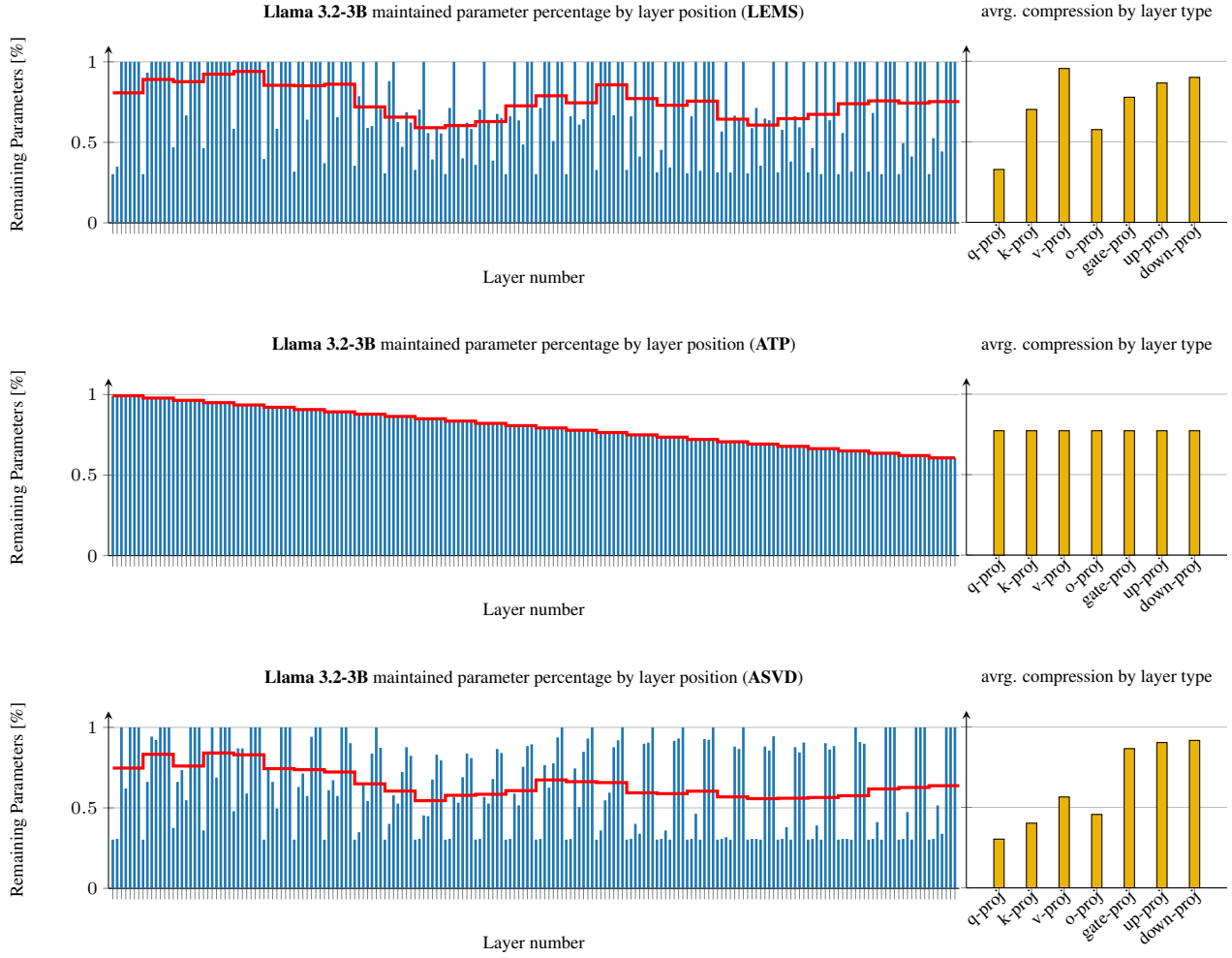
\begin{figure}[ht]
  \begin{minipage}[t]{0.7\textwidth}
    \input{data/plots/supp_search/compression_comparison_Llama32_3B_08_OURS}
  \end{minipage}
  \hfill
  \begin{minipage}[t]{0.25\textwidth}
    \input{data/plots/supp_search/compression_comparison_layer_type_Llama32_3B_08_OURS}
  \end{minipage}
  \begin{minipage}[t]{0.7\textwidth}
    \input{data/plots/supp_search/compression_comparison_Llama32_3B_08_ATP}
  \end{minipage}
  \hfill
  \begin{minipage}[t]{0.25\textwidth}
    \input{data/plots/supp_search/compression_comparison_layer_type_Llama32_3B_08_ATP}
  \end{minipage}
  \begin{minipage}[t]{0.7\textwidth}
    \input{data/plots/supp_search/compression_comparison_Llama32_3B_08_ASVD}
  \end{minipage}
  \hfill
  \begin{minipage}[t]{0.25\textwidth}
    \input{data/plots/supp_search/compression_comparison_layer_type_Llama32_3B_08_ASVD}
  \end{minipage}
  \caption{Average relative importance of block positions and block layer types in \gls{qwen3}-4B and \gls{llama3}.2-3B using \gls{kl}-divergence as importance metric.}\label{fig:search_comparisons_atp_asvd}
\end{figure}

\textbf{\gls{atp}} simply employs a monotonically increasing compression ratio across blocks. As seen in ~\cref{fig:search_comparisons_atp_asvd} (middle), this approach disregards the nuanced importance landscape, resulting in uniform compression across all block layer types. This indiscriminate compression leads to suboptimal performance (13.48 perplexity) due to over compression of critical layers.

\textbf{\gls{asvd}} represents an improvement by considering layer-specific importance, achieving a better perplexity of 12.06. Its compression profile is more varied than \gls{atp}'s (Figure~\ref{fig:search_comparisons_atp_asvd}, bottom). However, \gls{asvd}'s greedy, layer-by-layer optimization aiming towards most equal error fails to capture the global picture. While it correctly preserves more parameters in important layer types, its allocation is not as finely tuned as our \gls{oursearch}, leading to less efficient parameter-performance trade-offs.

In contrast, \textbf{our} global search strategy excels by finding an optimal configuration for the given error model. The compression configuration discovered by \gls{oursearch} (\cref{fig:search_comparisons_atp_asvd}, top) shows a strong correlation with the importance data from ~\cref{fig:search_importance_comparisons}. Moreover, its ability to perform a global trade-off based on both layer importance and absolute parameter count gives it a major advantage. The up-proj and down-proj layers, for example, are significantly larger than the q-proj layer. Consequently, a minor increase in the compression of these large layers frees up a substantial parameter budget. Our search strategically reallocates this budget to preserve smaller, but functionally critical, layers at a much higher fidelity. This global optimization, which is beyond the scope of \gls{asvd}'s greedy approach, allows our method to find a superior architectural trade-off and achieve better performance.

\subsubsection{How do Components Contribute to \gls{oursearch} performance?}\label{supp:sec:global_optimization_vs_bias_calibration}

In the main material \cref{sec:individual_contribution_lems} we discuss incremental contributions to \gls{oursearch}'s performance. However, the main section does not cover how big the advantage of our global \gls{ilp} solver is over using the same score with other searches (how much benefit comes from the optimization) and how much the data-based calibration of bias aids this performance. To disentangle these effects, we perform a cross-ablation on \gls{mistral}-7B, \gls{llama3}-8B, and \gls{qwen3}-8B at 0.8 parameter retention. For this ablation, we combine each of the three score-based search procedures (\gls{mrcs}, \gls{asvd}, and our ILP search) with four proxy scores: the dense spectral-energy proxy used by \gls{mrcs}, the sparse \gls{kl} proxy used by \gls{asvd}, our dense hybrid proxy, and our full hybrid proxy with adaptive bias. The results are presented in \cref{tab:proxy_search_cross_ablation}, except \gls{mrcs} on the sparse proxy as it is not applicable to the sparse \gls{kl} proxy requiring dense per-rank sensitivity curves.

The results show that neither data-based calibration nor global \gls{ilp} optimization alone explains the full gain our search has over others. First, global optimization is only useful when the optimized proxy is reliable: applying our ILP search to uncalibrated energy performs poorly on \gls{llama3}-8B, despite the solution being globally optimal under that proxy. This indicates that a global optimizer can be misled when the proxy does not reflect the true cross-layer importance. Second, proxy calibration alone is also insufficient. Retrofitting the \gls{mrcs} and \gls{asvd} searches with our hybrid or hybrid+bias scores improves results for some models, but does not recover the performance of our combined method. For example, the full hybrid+bias proxy with \gls{mrcs} achieves 10.71 perplexity, while the same proxy with our ILP search reaches a much better perplexity of 8.04. Consequently, the largest improvement appears only when the calibrated proxy and the global allocation through \gls{ilp} are used together. The dense, rank-dependent estimate of layer impact is provided by our hybrid+bias proxy, while the ILP can globally trade off rank choices across layers under the parameter budget. This interaction yields the best perplexity and accuracy in the cross-ablation, improving from 9.79 perplexity with \gls{asvd} search and 10.71 with \gls{mrcs} search to 8.04 with our ILP search. Thus, the main gain should be attributed to the alignment between proxy calibration and global optimization, rather than to either component in isolation, demonstrating that the combination presented in our work is key to achieving the highest possible performance.

\begin{table}[ht]
\caption{Cross-ablation of proxy scores and search algorithms on \gls{mistral}-7B, \gls{llama3}-8B and \gls{qwen3}-8B at 0.8 parameter retention. We report \gls{wiki} perplexity (lower is better) and average zero-shot accuracy in parentheses (higher is better). Best result for each proxy score is shown in bold, second best underlined. \gls{mrcs} cannot use the sparse \gls{kl} proxy because it requires dense per-rank sensitivity scores.}
\label{tab:proxy_search_cross_ablation}
\centering
\begin{tabular}{llcccc}
\toprule
\midrule
\textbf{Model} & \textbf{Proxy score} & \textbf{Type} & \gls{mrcs} search & \gls{asvd} search & Ours / ILP search \\
\midrule
\multirow{4}{*}{\textbf{\gls{mistral}-7B}} & Energy (MRCS) & Dense & \underline{7.06 (0.53)} & \textbf{6.48 (0.56)} & 9.45 (0.49) \\
 & KL (ASVD) & Sparse & n.a. & \underline{6.70 (0.53)} & \textbf{6.43 (0.56)} \\
 & Hybrid (ours) & Dense & 7.11 (0.51) & \underline{6.63 (0.53)} & \textbf{5.96 (0.57)} \\
 & Hybrid + bias (ours) & Dense & 7.03 (0.52) & \underline{6.62 (0.54)} & \textbf{5.92 (0.58)} \\
\midrule
\multirow{4}{*}{\textbf{\gls{llama3}-8B}} & Energy (MRCS) & Dense & \underline{10.54 (0.51)} & \textbf{8.91 (0.54)} & 10.77 (0.51) \\
 & KL (ASVD) & Sparse & n.a. & \textbf{11.65 (0.47)} & \underline{9.14 (0.53)} \\
 & Hybrid (ours) & Dense & 12.25 (0.47) & \underline{11.05 (0.48)} & \textbf{8.29 (0.54)} \\
 & Hybrid + bias (ours) & Dense & 10.71 (0.51) & \underline{9.79 (0.53)} & \textbf{8.04 (0.57)} \\
\midrule
\multirow{4}{*}{\textbf{\gls{qwen3}-8B}} & Energy (MRCS) & Dense & \underline{14.24 (0.53)} & 12.11 (0.58) & \textbf{12.47 (0.54)} \\
 & KL (ASVD) & Sparse & n.a. & \underline{11.65 (0.57)} & \textbf{10.98 (0.59)} \\
 & Hybrid (ours) & Dense & 12.21 (0.54) & \underline{11.62 (0.56)} & \textbf{10.24 (0.59)} \\
 & Hybrid + bias (ours) & Dense & 12.28 (0.54) & \underline{11.72 (0.56)} & \textbf{10.28 (0.59)} \\
\midrule
\bottomrule
\end{tabular}
\end{table}

\newpage
\subsection{Additional Explanations and Ablations on \gls{ours}}\label{supp:sec:kfac_long_ablations}

\input{content/supplementary_sections/fisher_section}

\subsection{\gls{llama2}-7B Results}\label{supp:sec:llama_2_7b_results}
While we evaluate \gls{ours} on a range of different relevant architectures in the main content, in this section, we want to provide additional results for \gls{llama2}-7B ~\cite{touvron2023llama2} as it is a very popular example in the \gls{svd} compression domain. The results are presented in \cref{tab:llama2_7b_results} and reiterate the strong results for both \gls{ours} and \gls{oursearch}.

\input{content/supp_llama7b}

\input{content/supp_concurrent}

\subsection{\gls{llama3}.1-70B Results}\label{supp:sec:llama_3_70b_results}
To demonstrate the scalability of \gls{oursearch} beyond the \gls{qwen3}-32B architecture, we extend our evaluation to \gls{llama3}.1-70B. As shown in \cref{supp:tab:llama_70B}, our method substantially outperforms the uniform baseline, reducing \glslink{wiki}{WikiText2}\glsunset{wiki} perplexity by 43\% (7.94 to 4.53). This advantage extends to zero-shot tasks, where \gls{oursearch} improves or matches the uniform baseline on every benchmark, ranging from identical performance on Winogrande to a substantial 9.0 \glslink{pp}{percentage points (p.p.)}\glsunset{pp} gain on HellaSwag, averaging a 4.0 \gls{pp} increase in accuracy overall. While \gls{asvd} achieves the second-best results, it incurs nearly 3.5$\times$ the computational cost of \gls{oursearch} and trails our accuracy by 2.5 \gls{pp}, highlighting the efficiency and effectiveness of our approach. Conversely, \gls{atp}'s simple linear compression strategy (designed for large models) and \gls{svdllmv2} fail to identify rank allocations with clear superiority over the uniform baseline. Similarly, \gls{mrcs} struggles, falling below the uniform baseline in average accuracy while achieving the third best \gls{wiki} perplexity. Notably, we adjusted its hyperparameters as it required more steps to compress the larger model. Overall, the consistent performance gains seen with our method, combined with the full cacheability of sensitivity metrics make our search a compelling option for large model compression.

\begin{table}[t]%
    \centering
    \captionof{table}{Additional results of all methods on \textbf{\gls{llama2}-7B} across multiple uniform compression rates and \textbf{\gls{svd} baselines}. \textbf{Bold}/\underline{Underline} denote best/second-best compressed results.}
    \resizebox{1\textwidth}{!}{%
        \input{data/tables/llama2_results}

    }
    \label{tab:llama2_7b_results}
\end{table}

\begin{table}[tp!]
    \centering %
    \caption{Extended format zero-shot performance and perplexity of \gls{llama3}.1-70B with a \textbf{0.8 parameter retention}.}
    \label{supp:tab:llama_70B}

    \resizebox{0.95\textwidth}{!}{%
        \begin{tabular}{c|ccccccccccc}
            \toprule
            \midrule
            \multicolumn{1}{c|}{Method} & \glslink{wiki}{Wiki}\glsunset{wiki} $\downarrow$ & \glslink{c4}{C4}\glsunset{c4} $\downarrow$ & \glslink{ptb}{PTB}\glsunset{ptb} $\downarrow$ & Openb. $\uparrow$ & ARC\_e $\uparrow$ & ARC\_c $\uparrow$ & WinoG. $\uparrow$ & HellaS. $\uparrow$ & PIQA $\uparrow$ & \textbf{Avg. $\uparrow$} & \textbf{time [h] $\downarrow$}\\ \midrule
            \multicolumn{1}{c|}{\color[HTML]{9B9B9B}Original} & {\color[HTML]{9B9B9B}2.81} & {\color[HTML]{9B9B9B}7.65} & {\color[HTML]{9B9B9B}7.22} & {\color[HTML]{9B9B9B}0.37} & {\color[HTML]{9B9B9B}0.60} & {\color[HTML]{9B9B9B}0.85} & {\color[HTML]{9B9B9B}0.82} & {\color[HTML]{9B9B9B}0.67} & {\color[HTML]{9B9B9B}0.83} & {\color[HTML]{9B9B9B}68.95} & -\\
            \midrule
            \multicolumn{1}{c|}{uniform} & 7.94 & 21.42 & 21.68 & 0.35 & 0.81 & 0.52 & \textbf{0.78} & 0.53 & 0.78 & 62.61 & - \\
            \multicolumn{1}{c|}{\gls{asvd}} & \underline{7.33} & 26.25 & \textbf{18.72} & \underline{0.36} & \underline{0.83} & \underline{0.54} & 0.75 & \underline{0.57} & \underline{0.79} & \underline{64.11} & 22.5 \\
            \multicolumn{1}{c|}{\gls{svdllmv2}} & 7.82 & \underline{21.13} & 21.46 & 0.34 & 0.81 & 0.52 & \underline{0.77} & 0.53 & 0.78 & 62.61 & $<$0.1\\
            \multicolumn{1}{c|}{\gls{mrcs}} & 7.52 & 22.02 & 22.23 & 0.35 & 0.81 & 0.51 & 0.76 & 0.53 & 0.78 & 62.38 & $<$0.1\\
            \multicolumn{1}{c|}{\gls{atp}} & 7.94 & 21.42 & 21.68 & 0.35 & 0.81 & 0.52 & \textbf{0.78} & 0.53 & 0.78 & 62.61 & 0.2\\
            \multicolumn{1}{c|}{\midrowcolor{m} \textbf{\gls{oursearch}}} & \textbf{4.53} & \textbf{18.50} & \underline{18.98} & \textbf{0.37} & \textbf{0.85} & \textbf{0.58} & \textbf{0.78} & \textbf{0.62} & \textbf{0.81} & \textbf{66.60} & 6.6 \\
            \midrule
            \bottomrule
        \end{tabular}
    }
\end{table}

\subsection{Comparison with Alternative Decomposition Techniques}
Besides standard layer-wise SVD-based compression we have discussed and compared with in the main material, two more recent approaches exist in related work: \glslink{bitstack}{BitStack}\glsunset{bitstack}~\cite{wang2025bitstack} and \glslink{basissharing}{BasisSharing}\glsunset{basissharing}~\cite{wang2025basis}. Since both of them deviate from the other works in how the decomposition is done, in this section we explain them in more detail and compare our approach to them.
\subsubsection{\acrshort{bitstack}}\label{supp:sec:bitstack}

\paragraph{\glslink{bitstack}{BitStack}\glsunset{bitstack} Differences to Standard \gls{svd}.}
\gls{bitstack}~\cite{wang2025bitstack} uses a fundamentally different approach to standard low-rank decomposition of weights. Specifically, they iteratively separate the sign $\mathcal{S}$ out of the weight matrix $W$ resulting in $\mathcal{S}\cdot |W|$, decompose it to \gls{svd}$(|W|)$, truncate the result to rank k=16 and reconstruct it to $|W|_k$. Subsequently, they compute the difference of the reconstructed and the original weight and separate sign and decompose the difference again. This way, with an increasing number of iterative steps, their weight approximation becomes better and better. As shown in \citet{wang2025bitstack} , this method is highly effective, preserving perplexity and zero shot accuracy well, even at low memory footprints. However, there are two main drawbacks: first, the approach is reconstructing the full weight before the forward pass, which makes it computationally expensive and results in slowdowns compared to the baseline, even with highly optimized kernels. In their paper they report it to be about 10\% slower than the fp16 baseline at approximately 25\% remaining size; second, separating out the binary mask makes combining it with quantization ineffective, specifically, in every iterative step a binary tensor with the size of the original weight is extracted. Hence, when using only 4 iterative steps, which is very few already, the binary masks alone take up the same amount of space as the original weight when quantized to 4 bit (excluding scaling factors). While the decomposed weight also contributes to the memory footprint and is quantizable, the space occupied by the non-quantizable binary mask is significantly larger. Hence, different to other \gls{svd} approaches, applying quantization to \gls{bitstack} has negligible memory advantages. Looking at a \gls{llama3}-8b example, the q\_proj layer has a shape of $4096\times4096$. With just 4 iterative blocks \gls{bitstack} uses $4096\times4096\times4 \times 1\,\text{bit} = 67,108,864$ for the sign mask, while using 16 ranks for each of the 4 blocks for the decomposed weight results in $(4096\times4\times16 + 4096\times4\times16) \times 16\,\text{bit} = 8,388,608\,\text{bit}$. In total this makes $75,497,472 \,\text{bit}$, which is 28.125\% of the original weight. Now, when considering 4 bit quantization with 128 group size, the total bit count for the decomposed part comes down from $8,388,608$ to $2,162,688 \,\text{bit}$. Considering this quantization, the total comes down to 25.81\% of the original weight size. Meanwhile, when considering the same weight compressed with other \gls{svd} methods and combined with quantization, \textbf{all} layer components are quantizable. Hence, when performing mild \gls{svd} compression that removes only 10\% of parameters and perform 4 bit quantization on top of it, we get $(4096\times1844 + 1844\times4096) \times 16\,\text{bit} = 241,696,768 \,\text{bit}$ for the decomposition, and $62,312,448 \,\text{bit}$ when quantizing to 128 groups $4\,\text{bit}$, reducing it to 23.21\% of the original size. Hence, the total reduction achieved when combining it with quantization is similar while likely achieving runtime acceleration. Therefore, overall, combining traditional \gls{svd} with quantization is more effective as it allows for inference acceleration and high weight reduction, which \gls{bitstack}'s sign separation approach cannot offer.

\paragraph{\gls{bitstack} Performance Evaluation.}
In this section, we compare \gls{ours} with \gls{bitstack}~\cite{wang2025bitstack}, which we treat separately due to its fundamentally different decomposition. Unlike the other \gls{svd}-based methods evaluated so far, which preserve separate low-rank layers at inference time (\cref{eq:factors_prelim}), \gls{bitstack} requires reconstructing full weights prior to computation with limited quantization impact as described above. %
In \cref{tab:bitstack_comparison}, we compare the memory–accuracy trade-off on \gls{llama3}-8B, as provided in the \gls{bitstack} repository. To reflect realistic deployment, we report performance under AWQ quantization for both methods. For \gls{bitstack}, however, we only report the \textit{theoretical} footprint under AWQ \textit{without actually compressing it}, since its low-rank structure is incompatible with standard quantization kernels. Even under this favorable assumption for the baseline, our decomposition and search yield a better trade-off, improving compression while maintaining higher accuracy.

\begin{table}[]
    \centering
    \captionof{table}{Perplexity on \gls{wiki} ($\downarrow$) and zero shot accuracy of \gls{llama3}-8B. For \gls{bitstack} we report the quantized memory, but float accuracy due to compatibility issues, favoring its results.}
    \resizebox{0.4\textwidth}{!}{%
    \begin{tabular}{c | c c}
    \toprule
    \midrule
    Method                 & Weight Memory & Accuracy\\ \midrule
    \gls{bitstack}                       & 5.19 GB            & 59.93 \\
    \gls{ours}                          & \textbf{5.07 GB}   & \textbf{60.83} \\
    \midrule
    \bottomrule
    \end{tabular}
    }
    \label{tab:bitstack_comparison}

\end{table}

\subsubsection{BasisSharing}\label{supp:sec:basissharing}
\paragraph{BasisSharing Differences to Standard SVD.}

Standard \gls{svd} compresses every layer individually into two separate sublayers (see \cref{eq:factors_prelim}), with both of the factors being specific to that layer. Differently, \glslink{basissharing}{BasisSharing}\glsunset{basissharing} \cite{wang2025basis} groups adjacent layers of the same type, and shares one low-rank component between the layers of that group. Considering a group of layers $\mathcal{G}=\{i_1,\ldots,i_m\}$, members are first concatenated, then jointly compressed and finally split again into a shared and individual component
\begin{equation}
    [\widehat{\mathbf{W}}_{i_1},\ldots,\widehat{\mathbf{W}}_{i_m}]
    =
    \mathbf{C}_{\mathcal{G}}
    [\mathbf{B}_{i_1},\ldots,\mathbf{B}_{i_m}],
    \qquad
    \widehat{\mathbf{W}}_i
    =
    \mathbf{C}_{\mathcal{G}}\mathbf{B}_i .
\end{equation}
Since $\mathbf{C}_{\mathcal{G}}$ is only stored once per group and just $\mathbf{B}_i$ is layer specific, the number of stored parameters compared to independent \gls{svd} is reduced. Specifically, for $m$ layers of the same type with shared rank $r_{\mathcal{G}}$, independent \gls{svd} stores
$m r_{\mathcal{G}}(d_{\mathrm{out}}+d_{\mathrm{in}})$ parameters, whereas
\gls{basissharing} only stores $r_{\mathcal{G}}d_{\mathrm{out}} + m r_{\mathcal{G}}d_{\mathrm{in}}$. However, while the sharing results in a lower storage footprint, the shared basis are still \textit{applied} independently during the model inference. Hence, the computational load of the approach is higher than the one of traditional \gls{svd} at the same number of parameters.

\paragraph{BasisSharing Evaluation.} In this paragraph, we compare \gls{ours} to \gls{basissharing} and demonstrate applicability of \gls{oursearch} to \gls{basissharing}.
For the comparison, we fully reimplement \gls{basissharing} in our repository. We further extend our \gls{oursearch} formulation to work with the concept of layer sharing. To do that, our \gls{ilp} formulation (\cref{eq:ilp_obj}) must be amended to ensure, that all layers inside a sharing group $\mathcal{G}$ select the same rank candidate. For that we additionally impose the constraint
\begin{equation}
    x_{ij} = x_{i'j},
    \qquad
    \forall i,i'\in\mathcal{G},
    \quad
    \forall j,
    \quad
    \forall \mathcal{G},
    \label{eq:bs_rank_tying}
\end{equation}
which forces the same intermediate rank on all layers within a group.

Starting with the assessment of the reimplementation, as shown in \cref{tab:basis_sharing_reference_and_lems}, when using the same 32-bit float model, dataloader and seed, our performance matches the reference (Tab. 2 in \cite{wang2025basis}), validating our implementation. However, for the rest of the comparisons we opt to use the pipeline as is used throughout this paper (i.e. fp16, same data and seed) to ensure comparability. With these settings we notice a degradation which is consistent \gls{llama2}'s high sensitivity to calibration data we have noticed in prior experiments. When applying our updated search formulation, we observe a major improvement across perplexity and zero-shot results. In particular \gls{wiki} perplexity reduces from 8.12 to just 6.25. For zero-shot accuracy the increase is even more impressive with an improvement of \textbf{9.3 \gls{pp}} over the uniform baseline and strong improvements of 6.0 \gls{pp} over the much more computationally expensive fp32 reference. Interestingly, \gls{ptb} is the only benchmark where it suffers substantial degradation. However, with the improvements across the board in all other benchmarks, we regard this as an outlier.

\begin{table}[]
    \centering
    \caption{\gls{llama2}-7B compressed with \gls{basissharing} to 0.8 parameter count with the data and precision used in the original work, the data and precision of our work and with \gls{oursearch} applied to it.}
    \label{tab:basis_sharing_reference_and_lems}
    \resizebox{0.95\textwidth}{!}{%
        \input{data/tables/ablation_basis_sharing_lems}
    }
\end{table}

\begin{table}[]
    \centering
    \caption{Comparing \gls{ours} with BasisSharing at matched compute (TFLOPs) and memory (GParams) budgets with respect to the linear layers that are considered in the compression of \gls{llama2}-7B \textit{without any search} applied.}
    \label{tab:basis_sharing_comparison}
    \resizebox{0.95\textwidth}{!}{%
    \input{data/tables/ablation_basis_sharing_matched}
    }
\end{table}

In the head-to-head comparison without search, due to the differences between standard \gls{svd} and \gls{basissharing}, we compare \gls{ours} against \gls{basissharing} under both matched memory and matched compute budgets using the same data and precision.
The results of the comparison are presented in \cref{tab:basis_sharing_comparison}. As can be seen, for both matching memory and matching compute our method comes out ahead in perplexity and zero-shot comparisons, positioning it as a capable alternative to \gls{basissharing} in both scenarios.

\subsection{Calibration Data Ablations}
Since our method uses data-based calibration, we investigate the impact the data has on the final result. For that we investigate the impact of the sample count, the sample seed and the dataset.

\paragraph{Impact of Sample Count.}
We start with investigating the sample count. In \cref{fig:ablation_data_impact} (left), the performance improves when more samples become available, as one would expect. Going from 64 to 512 (8$\times$) samples, perplexity improves from 20.54 to 17.54, with diminishing returns: doubling from 64 to 128 samples gains 1.98, doubling from 256 (our default) to 512 only 0.46.

\begin{figure}[t]
    \centering
    \begin{minipage}[t]{0.48\textwidth}
        \centering
        \input{data/plots/ablation_seed_and_samples}
        \caption{Perplexity of \gls{qwen3}-8B when compressed using different sample counts and different sampling seeds when removing 40\% of parameters with \gls{ours} at uniform ranks (no search), isolating the data dependence of the decomposition.}
        \label{fig:ablation_data_impact}
    \end{minipage}
    \hfill %
    \begin{minipage}[t]{0.48\textwidth}
        \centering
        \input{data/plots/ablation_throughput}
        \caption{Throughput (token/s) of the baseline \gls{qwen3}-8B and \gls{oursearch} compressed variants at different compression ratios.}
        \label{fig:throughput}
    \end{minipage}
\end{figure}

\paragraph{Impact of Random Seed.}
We continue with the evaluation of the random seed (\cref{fig:ablation_data_impact}, right).
The seed has a smaller impact than the sample count, with the largest difference being 1.05 perplexity (between seed 0 and seed 42).

\paragraph{Impact of Calibration Dataset.}
In all experiments, we used the \gls{wiki} dataset for calibration. To understand the impact of the calibration set, we repeated the experiments for the 7B and 8B models using the fineweb 16M dataset introduced by~\cite{penedo2024_fineweb}. This dataset was chosen for its diversity and absence from most common comparisons, allowing for an unbiased evaluation. The results are given in Table~\ref{tab:fineweb_ablation}. When comparing the results to the \gls{wiki} calibration, it is clear that the calibration set has a significant impact on the performance of all methods. While the \gls{wiki} performance, as expected, drops significantly, \gls{ptb}, \gls{c4} and the classification task performance improve. Specifically, for our method at 0.7 compression, on \gls{llama3}-8B the accuracy improves from 43.69 to 45.19 without search and from 49.88 to 51.18 with search. Similarly, the performance improves from 46.26 to 46.87 without and from 51.68 to 52.61 with search for \gls{mistral}-7B and from 48.24 to 50.86 without and from 53.30 to 54.89 with search for \gls{qwen3}-8B, experiencing the biggest improvement. Other methods see comparable improvements e.g. \glslink{svdllm}{SVD-LLM}\glsunset{svdllm} which improves by 1.07 \gls{pp} and 1.93 \gls{pp} for 0.7 and 0.9 on \gls{qwen3}-8B, respectively. These improvements highlight the higher quality of the fineweb dataset compared to its \gls{wiki} counterpart. Despite the improvements for the baseline methods, our method still outperforms baselines, demonstrating effectiveness independent of the dataset used.

\input{data/tables/fineweb_ablation_cr}

\subsection{Inference Acceleration}
To quantify the practical efficiency gains of \gls{svd} when used with \gls{oursearch}, we evaluated the wall-clock inference speedup of the compressed \gls{qwen3}-8B model against the uncompressed baseline. All measurements were conducted on a single NVIDIA A100 GPU. Adopting the evaluation protocol from prior work~\cite{2025-svd-llmv2}, we fixed the batch size to 4, with a prefill sequence length of 1024 tokens and a decoding length of 256 tokens.
As illustrated in Figure~\ref{fig:throughput}, \gls{oursearch} demonstrates a consistent improvement in generation speed as the model is compressed further. The method achieves a $1.14\times$ speedup at a 20\% compression ratio, scaling up to $1.61\times$ at 60\% compression. At the most aggressive setting of 80\% compression, the model reaches a substantial $2.01\times$ speedup over the dense baseline.

\subsection{Qualitative Results}
We present qualitative generations from \gls{svd}-compressed \gls{qwen3}-8B models \emph{without} any post-compression finetuning. We compare \gls{ours} and \gls{oursearch} against \gls{svdllm}, a state-of-the-art \gls{svd}-based compression baseline. \cref{tab:qualitative_results_no_ft} reports representative model outputs for the same prompt across compression ratios, illustrating the effects of compression on response coherence, formatting, and content fidelity.
\input{data/tables/qualitative_asking_questions}

\clearpage
\section{Algorithms and Implementation Details}\label{supp:sec:algorithms}
\paragraph{End-to-End Compression Pipeline.}
Algorithm~\ref{alg:high_level} outlines the high-level workflow of our framework. Crucially, the pipeline is modular, decoupling the local decomposition strategy from the global rank allocation.
\begin{enumerate}
\item \textbf{Phase I: Decomposition.} We iterate through layers to compute decompositions, including the singular value spectra. While this work proposes the robust \gls{ours} estimator, this step can be substituted with any \gls{svd} variant (e.g., standard \gls{svd}, activation-\gls{svd}) provided it yields a valid singular value spectrum for the search.
\item \textbf{Phase II: Global Rank Allocation (\gls{oursearch}).} We solve for the optimal integer ranks $\{r_\ell^\star\}$ that maximize performance under the parameter budget $P_{\text{target}}$. This search utilizes the precomputed spectra but is agnostic to how they were generated.
\item \textbf{Phase III: Reconstruction.} Layers are replaced with their low-rank factorizations. For data dependent methods, this may optionally involve recalibrating statistics used in the decomposition, but is not used for our results.
\end{enumerate}
\input{data/algorithms/overview_algorithm}

\newpage
\paragraph{\gls{kfac} Fisher-Weighted \gls{svd} Decomposition.}
\label{sec:supp:kfac_svd}
The core of our \gls{svd} method is the robust estimation of the whitening matrices, detailed in Algorithm~\ref{alg:kfac_svd}.
Unlike prior Fisher-based \gls{svd} \cite{chekalina2025generalizedfisherweightedsvdscalable} that suffered from rank deficiency ($k \ll d_{\text{in}}d_{\text{out}}$), we use a decoupled token-wise estimation (Lines 4-8).
To ensure numerical stability, we shrink both estimates towards a scaled identity (Line 10), $\mathbf{M} \gets (1-\alpha)\mathbf{M} + \alpha\,\overline{\operatorname{diag}}(\mathbf{M})\,\mathbf{I}$ with the mean diagonal entry $\overline{\operatorname{diag}}(\mathbf{M})$, using $\alpha_A = 0.1$ for the input and $\alpha_G = 0.3$ for the gradient covariance. With the damping applied, the factors are well conditioned so that the subsequent Cholesky decomposition works properly.

Implementation Note on Whitening: While the main text formulates the whitening transform using the symmetric square root $\mathbf{G}^{1/2}$ for notational clarity, our implementation uses Cholesky factors for numerical efficiency. Specifically, we compute $\mathbf{G} = \mathbf{L}_G \mathbf{L}_G^\top$ and form the whitened weights as $\widetilde{\mathbf{W}} = \mathbf{L}_G^\top \mathbf{W} \mathbf{L}_A$ (Line 12), so that $\|\mathbf{L}_G^\top \Delta\mathbf{W}\, \mathbf{L}_A\|_F^2 = \operatorname{tr}(\Delta\mathbf{W}^\top \mathbf{G}\, \Delta\mathbf{W}\, \mathbf{A})$, and un-whiten the rank-$r$ factors with $\mathbf{L}_G^{-\top}$ and $\mathbf{L}_A^{-1}$ (Line 14). This transformation is isometric to the symmetric square root formulation with respect to the Frobenius norm objective, yielding identical singular values and optimal low-rank indices.
\input{data/algorithms/kfac_algorithm}
\newpage
\paragraph{Global Rank Allocation.}
\label{sec:supp:rank_alloc}

Algorithm~\ref{alg:rank_allocation} details the \gls{oursearch} procedure. This process converts the raw Fisher spectral energy into a calibrated additive proxy as discussed in \cref{sec:erorr_modelling}.
The procedure begins by anchoring the \textbf{scale} of the error: for each layer $\ell$, we measure the empirical \gls{kl}-divergence at a reference compression (e.g., $c_{\text{ref}}=10\%$) to determine the scalar importance $\alpha_\ell$ (Line 6).
Next, we address the error propagation correction using our \textbf{bias}-function. We parameterize the bias $\beta(d_\ell)$ using a harmonic decay function controlled by the block based depth $d_\ell$ of layer $\ell$, magnitude $\kappa$ and decay rate $\gamma$ (Eq.~\ref{eq:detailed_bias_function}). This formulation assigns higher penalties to earlier blocks to counteract the \textit{error explosion} effect~\cite{atp_huang2025determining}.
Crucially, because the optimal propagation penalty depends on the specific architecture (depth, residual connections), we cannot use fixed values. Instead, we optimize $\kappa$ and $\gamma$ via a bilevel optimization loop (Lines 8-25). We employ a \gls{tpe} sampler provided by \cite{Akiba_Optuna_A_next_generation_2019} to propose candidate $(\kappa, \gamma)$ pairs. For each pair, we:
1. Construct Costs: Calculate the layer-wise costs $L_\ell(r)$ using the current bias parameters (Lines 11-16).
2. Solve \gls{ilp}: Obtain the optimal rank configuration for this specific cost landscape (Line 18).
3. Validate: Temporarily apply these ranks and measure the true validation loss (Line 19, 20).
This \textit{loop-in-the-loop} ensures that the error model is dynamically fitted to maximize the generalization of the final rank assignment.
\input{data/algorithms/search_algorithm}

\newpage
\section{Detailed Result Tables}\label{supp:sec:detailed_main_tables}

\paragraph{Detailed Search Table.}
\cref{tab:search_results_wide} in the main content only reports the average across the zero-shot datasets. Here we present the full results, with all individual accuracy values.
The detailed results for 0.8 ratio are presented in \cref{supp:tab:extended_search_results_norecalib_08}. The results for 0.6 are presented in \cref{supp:tab:extended_search_results_norecalib_06}.

\paragraph{Detailed SVD Table.}
In the main content \cref{tab:llm_main_table_short} only reports the average across the zero-shot datasets. In \cref{supp:tab:llm_main_extended_09} and \cref{supp:tab:llm_main_extended_07} we present the detailed results with all individual accuracy values for 0.9 and 0.7 compression ratio, respectively.
\setboolean{shorttable}{false}

\paragraph{Numerical Results for \cref{fig:generalization_experiments}.}
In \cref{tab:ablation_generalization_table_results}, we provide the extended version of \cref{fig:generalization_experiments}, which includes the detailed data for all models. Please note that because of stability issues for \gls{svdllm} on the 32B model, we opted for \gls{svdllmv2} instead, which performs nearly identical in our experiments but is much more expensive due to its \gls{svd}-based whitening.
\input{data/tables/supp_extended_main_results_norecalib_cr}
\input{data/tables/supp_extended_search_results_norecalib_cr}

\begin{table}[h!]
    \centering
    \caption{Detailed performance evaluation of \textbf{\gls{oursearch}} for both \gls{ours} and \gls{svdllm} across five \gls{qwen3} model sizes (1.7B-32B). The table contrasts the baseline uniform rank allocation with our proposed \gls{oursearch} across compression ratios $c \in [0.5,0.9]$. \textbf{Bold} indicates the best result between the uniform and search-based allocation for each method, showing the massive improvements brought about by \gls{oursearch}.}
    \input{data/tables/ablation_generalize_results_table_norecalib_cr}
    \label{tab:ablation_generalization_table_results}
\end{table}

%% file: data/plots/supp_search/structural_error_underestimation_llama32_3B_05.tex
\begin{tikzpicture}
\begin{axis}[
    ybar, ymin=0, %
    xlabel=\scriptsize{Block number},
    xlabel shift        = {-0.1 cm},
    xtick=data,
    xticklabel style = { font=\scriptsize, yshift=+0.08cm
    },
    xtick = {2, 6, 10, 14, 18, 22, 26, 30, 34, 38, 42, 46, 50, 54, 58, 62, 66, 70, 74, 78, 82, 86, 90, 94, 98, 102, 106},
    xticklabels         = { 1, , 3, , 5, , 7, , 9, , 11, , 13, , 15, , 17, , 19, , 21, , 23, , 25, , 27, , 29, , 31, , 33, , 35, ,},
    axis x line* = bottom,
    axis y line = left,
    height = 4.0cm,
    width = 1.1\textwidth,
    title = {\scriptsize\textbf{\gls{llama3}.2-3B} single error sum vs. measured error },
    ymin = 0.0, ymax = 6.50,
    xmin=-1.0, xmax=114.0,
    bar width=1.2px,
    yticklabel style = { font=\scriptsize, },
    yticklabel shift = {-0.05cm},
    ylabel shift    = {-2.9 cm},
    ymajorgrids,
    ybar=0.033cm,
    ]
\addplot[fill=DarkYellow] coordinates {
(2, 2.1331396102905273)
(6, 1.7052183151245117)
(10, 1.6241178512573242)
(14, 1.5226831436157227)
(18, 1.434849739074707)
(22, 1.3517084121704102)
(26, 1.2901582717895508)
(30, 1.2287626266479492)
(34, 1.1714458465576172)
(38, 1.1332406997680664)
(42, 1.097975730895996)
(46, 1.0667967796325684)
(50, 1.0340080261230469)
(54, 0.9991850852966309)
(58, 0.9589986801147461)
(62, 0.9055156707763672)
(66, 0.8468179702758789)
(70, 0.7842311859130859)
(74, 0.7252974510192871)
(78, 0.6608471870422363)
(82, 0.588158130645752)
(86, 0.5245985984802246)
(90, 0.4657416343688965)
(94, 0.4090757369995117)
(98, 0.3477010726928711)
(102, 0.27744197845458984)
(106, 0.20006752014160156)
(110, 0.10692787170410156)
};
\addlegendentry{\scriptsize Single Error Sum};

\addplot[fill=icmlblue] coordinates {
(2, 4.621826171875)
(6, 3.5374755859375)
(10, 3.27239990234375)
(14, 2.982177734375)
(18, 2.72265625)
(22, 2.3861083984375)
(26, 2.116943359375)
(30, 1.828338623046875)
(34, 1.628204345703125)
(38, 1.55902099609375)
(42, 1.494598388671875)
(46, 1.43292236328125)
(50, 1.3616180419921875)
(54, 1.29150390625)
(58, 1.2050018310546875)
(62, 1.1029815673828125)
(66, 0.9947509765625)
(70, 0.9076080322265625)
(74, 0.8261260986328125)
(78, 0.7567901611328125)
(82, 0.6730880737304688)
(86, 0.5905685424804688)
(90, 0.5161895751953125)
(94, 0.4365234375)
(98, 0.34619140625)
(102, 0.256378173828125)
(106, 0.18029403686523438)
(110, 0.10692787170410156)
};
\addlegendentry{\scriptsize Total Error (Reference)};

xtick = {2, 6, 10, 14, 18, 22, 26, 30, 34, 38, 42, 46, 50, 54, 58, 62, 66, 70, 74, 78, 82, 86, 90, 94, 98, 102, 106, 110},

\end{axis}
\end{tikzpicture}

%% file: data/plots/supp_search/structural_error_underestimation_Qwen3_4B_05.tex
\begin{tikzpicture}
\begin{axis}[
    ybar, ymin=0, %
    xlabel=\scriptsize{Block number},
    xlabel shift        = {-0.1 cm},
    xtick=data,
    xticklabel style = { font=\scriptsize, yshift=+0.08cm
    },
    xtick = {2, 6, 10, 14, 18, 22, 26, 30, 34, 38, 42, 46, 50, 54, 58, 62, 66, 70, 74, 78, 82, 86, 90, 94, 98, 102, 106, 110, 114, 118, 122, 126, 130, 134, 138},
    xticklabels         = { 1, , 3, , 5, , 7, , 9, , 11, , 13, , 15, , 17, , 19, , 21, , 23, , 25, , 27, , 29, , 31, , 33, , 35, ,},
    axis x line* = bottom,
    axis y line = left,
    height = 4.0cm,
    width = 1.1\textwidth,
    title = {\scriptsize\textbf{\gls{qwen3}-4B} single error sum vs. measured error},
    ymin = 0.0, ymax = 6.50,
    xmin=-1.0, xmax=145.0,
    bar width=1.3px,
    yticklabel style = { font=\scriptsize, },
    yticklabel shift = {-0.05cm},
    ylabel shift    = {-2.9 cm},
    ymajorgrids,
    ybar=0.033cm,
    ]
\addplot[fill=DarkYellow] coordinates {
(2, 2.6274633407592773)
(6, 2.55258846282959)
(10, 2.5391746759414673)
(14, 2.5248360633850098)
(18, 2.486814022064209)
(22, 2.4238243103027344)
(26, 2.3435115814208984)
(30, 2.2499589920043945)
(34, 2.129487991333008)
(38, 2.008197784423828)
(42, 1.8947343826293945)
(46, 1.783005714416504)
(50, 1.6870298385620117)
(54, 1.6133575439453125)
(58, 1.546823501586914)
(62, 1.4874420166015625)
(66, 1.4409737586975098)
(70, 1.3884434700012207)
(74, 1.3484406471252441)
(78, 1.312490463256836)
(82, 1.2662715911865234)
(86, 1.2288422584533691)
(90, 1.1855077743530273)
(94, 1.1313629150390625)
(98, 1.0735702514648438)
(102, 1.0032463073730469)
(106, 0.9334774017333984)
(110, 0.8633127212524414)
(114, 0.7879619598388672)
(118, 0.6926889419555664)
(122, 0.6164283752441406)
(126, 0.5056076049804688)
(130, 0.41826725006103516)
(134, 0.3203115463256836)
(138, 0.2338695526123047)
(142, 0.11920738220214844)
};

\addplot[fill=icmlblue] coordinates {
(2, 3.538818359375)
(6, 3.17828369140625)
(10, 3.12939453125)
(14, 3.08050537109375)
(18, 3.0054931640625)
(22, 2.8897705078125)
(26, 2.6807861328125)
(30, 2.52587890625)
(34, 2.3486328125)
(38, 2.123748779296875)
(42, 1.979949951171875)
(46, 1.9212646484375)
(50, 1.8424072265625)
(54, 1.798004150390625)
(58, 1.757598876953125)
(62, 1.73016357421875)
(66, 1.7125244140625)
(70, 1.677825927734375)
(74, 1.640594482421875)
(78, 1.6272430419921875)
(82, 1.577423095703125)
(86, 1.5269012451171875)
(90, 1.5132904052734375)
(94, 1.4097900390625)
(98, 1.3366241455078125)
(102, 1.2472991943359375)
(106, 1.1697998046875)
(110, 1.0577239990234375)
(114, 0.9444427490234375)
(118, 0.8473052978515625)
(122, 0.763214111328125)
(126, 0.6258926391601562)
(130, 0.560516357421875)
(134, 0.380462646484375)
(138, 0.22052383422851562)
(142, 0.11920738220214844)
};

xtick = {2, 6, 10, 14, 18, 22, 26, 30, 34, 38, 42, 46, 50, 54, 58, 62, 66, 70, 74, 78, 82, 86, 90, 94, 98, 102, 106, 110},

\end{axis}
\end{tikzpicture}

%% file: data/plots/supp_search/harmonic_scale.tex
\begin{tikzpicture}
\begin{axis}[
    title={Different bias scaling functions based on the layer depth position $d_i$: $\beta(d_i, \gamma, \kappa=1)$},
    xlabel={$i$},
    ylabel={$\beta(d_i, \gamma, \kappa=1)$},
    xmin=1, xmax=28,
    ymin=0, ymax=1,
    grid=both,
    legend style={at={(1.05,0.5)},anchor=west},
    legend cell align={left},
    width=12cm, height=5cm
]

\addplot[smooth, thick, color={rgb,255:red,31; green,119; blue,180}] coordinates {
(1, 1.0)
(2, 0.9999959187837976)
(3, 0.9999546531533063)
(4, 0.999800020406081)
(5, 0.9994123048668503)
(6, 0.9986282578875172)
(7, 0.9972410978471584)
(8, 0.9950005101520253)
(9, 0.9916126472355429)
(10, 0.9867401285583104)
(11, 0.9800020406081013)
(12, 0.9709739368998628)
(13, 0.9591878379757167)
(14, 0.9441322314049587)
(15, 0.9252520717840583)
(16, 0.9019487807366595)
(17, 0.8735802469135803)
(18, 0.8394608259928126)
(19, 0.7988613406795225)
(20, 0.7510090807060504)
(21, 0.6950878028319106)
(22, 0.6302377308437914)
(23, 0.5555555555555556)
(24, 0.47009443480823954)
(25, 0.37286399347005406)
(26, 0.262830323436384)
(27, 0.13891598362978835)
(28, 0.0)
};
\addlegendentry{$\gamma=-3$};

\addplot[smooth, thick, color={rgb,255:red,255; green,127; blue,14}] coordinates {
(1, 1.0)
(2, 0.9995634909560124)
(3, 0.9980588939790666)
(4, 0.9951084222414659)
(5, 0.9904168525172972)
(6, 0.9837328093485255)
(7, 0.9748334773296625)
(8, 0.9635165930699406)
(9, 0.9495956116992335)
(10, 0.932896519441281)
(11, 0.9132555988337188)
(12, 0.8905177894424934)
(13, 0.8645354428107128)
(14, 0.8351673503025205)
(15, 0.8022779667162764)
(16, 0.7657367785117218)
(17, 0.7254177815090141)
(18, 0.6811990431960914)
(19, 0.6329623316057664)
(20, 0.5805927973888706)
(21, 0.5239786989792318)
(22, 0.46301116308850954)
(23, 0.39758397447963556)
(24, 0.32759339023876727)
(25, 0.25293797472484425)
(26, 0.1735184521097028)
(27, 0.08923757399028029)
(28, 0.0)
};
\addlegendentry{$\gamma=-1.5$};

\addplot[smooth, thick, color={rgb,255:red,44; green,160; blue,44}] coordinates {
(1, 1.0)
(2, 0.997979797979798)
(3, 0.9932659932659933)
(4, 0.9858585858585859)
(5, 0.9757575757575757)
(6, 0.9629629629629629)
(7, 0.9474747474747475)
(8, 0.9292929292929293)
(9, 0.9084175084175085)
(10, 0.8848484848484849)
(11, 0.8585858585858586)
(12, 0.8296296296296296)
(13, 0.797979797979798)
(14, 0.7636363636363637)
(15, 0.7265993265993266)
(16, 0.6868686868686869)
(17, 0.6444444444444445)
(18, 0.5993265993265994)
(19, 0.5515151515151515)
(20, 0.501010101010101)
(21, 0.4478114478114478)
(22, 0.39191919191919194)
(23, 0.3333333333333333)
(24, 0.27205387205387205)
(25, 0.2080808080808081)
(26, 0.1414141414141414)
(27, 0.07205387205387205)
(28, 0.0)
};
\addlegendentry{$\gamma=-1$};

\addplot[smooth, thick, color={rgb,255:red,214; green,39; blue,40}] coordinates {
(1, 1.0)
(2, 0.99099230882088)
(3, 0.9770676247484719)
(4, 0.9595853035470433)
(5, 0.939160553401611)
(6, 0.9161684308371819)
(7, 0.8908688093029746)
(8, 0.8634555813725742)
(9, 0.8340806313904007)
(10, 0.8028671296346044)
(11, 0.7699175859828433)
(12, 0.7353190547538294)
(13, 0.6991466692743613)
(14, 0.6614661382690863)
(15, 0.6223355661816009)
(16, 0.5818068161344833)
(17, 0.5399265534936664)
(18, 0.49673706029667436)
(19, 0.45227688145217576)
(20, 0.4065813449233354)
(21, 0.35968298583891367)
(22, 0.3116118962119156)
(23, 0.26239601624922565)
(24, 0.2120613792281545)
(25, 0.16063231904443603)
(26, 0.10813164744471353)
(27, 0.054580806410273006)
(28, 0.0)
};
\addlegendentry{$\gamma=-0.5$};

\addplot[smooth, thick, color={rgb,255:red,148; green,103; blue,189}] coordinates {
(1, 1.0)
(2, 0.9629629629629629)
(3, 0.9259259259259259)
(4, 0.8888888888888888)
(5, 0.8518518518518519)
(6, 0.8148148148148148)
(7, 0.7777777777777778)
(8, 0.7407407407407407)
(9, 0.7037037037037037)
(10, 0.6666666666666666)
(11, 0.6296296296296297)
(12, 0.5925925925925926)
(13, 0.5555555555555556)
(14, 0.5185185185185185)
(15, 0.48148148148148145)
(16, 0.4444444444444444)
(17, 0.4074074074074074)
(18, 0.37037037037037035)
(19, 0.3333333333333333)
(20, 0.2962962962962963)
(21, 0.25925925925925924)
(22, 0.2222222222222222)
(23, 0.18518518518518517)
(24, 0.14814814814814814)
(25, 0.1111111111111111)
(26, 0.07407407407407407)
(27, 0.037037037037037035)
(28, 0.0)
};
\addlegendentry{$\gamma=0$};

\addplot[smooth, thick, color={rgb,255:red,140; green,86; blue,75}] coordinates {
(1, 1.0)
(2, 0.8716897160607034)
(3, 0.7907134581808558)
(4, 0.7264148690726979)
(5, 0.6714322131428417)
(6, 0.6226096262888362)
(7, 0.5782497798197525)
(8, 0.5373154712895016)
(9, 0.49911801079100204)
(10, 0.46317249200650146)
(11, 0.42912222033999897)
(12, 0.3966957520496037)
(13, 0.3656808376399297)
(14, 0.33590778868209586)
(15, 0.30723841039706784)
(16, 0.27955838023554547)
(17, 0.25277184539697073)
(18, 0.22679749739094504)
(19, 0.20156565818952063)
(20, 0.17701607651796378)
(21, 0.15309623356790586)
(22, 0.12976002120131544)
(23, 0.10696669719523662)
(24, 0.08468004970206663)
(25, 0.06286772189146879)
(26, 0.04150066076873502)
(27, 0.020552663354963106)
(28, 0.0)
};
\addlegendentry{$\gamma=0.5$};

\addplot[smooth, thick, color={rgb,255:red,227; green,119; blue,194}] coordinates {
(1, 1.0)
(2, 0.6721619889462495)
(3, 0.5446694290920133)
(4, 0.46453124861220757)
(5, 0.40598874663832346)
(6, 0.3598485821196473)
(7, 0.3217663889164339)
(8, 0.2893428493616674)
(9, 0.26111235396537225)
(10, 0.23611381281748284)
(11, 0.21368279100854207)
(12, 0.19334074991862685)
(13, 0.17473158745543085)
(14, 0.15758313764646545)
(15, 0.14168270501158337)
(16, 0.12686090987581988)
(17, 0.1129806715045925)
(18, 0.0999294850871229)
(19, 0.0876138772644688)
(20, 0.07595534202073657)
(21, 0.06488730660481934)
(22, 0.05435282928485753)
(23, 0.04430282683183559)
(24, 0.034694691886298944)
(25, 0.02549120162326516)
(26, 0.016659647039776373)
(27, 0.008171131419229283)
(28, 0.0)
};
\addlegendentry{$\gamma=1$};

\addplot[thick, color={rgb,255:red,127; green,127; blue,127}] coordinates {
(1, 1.0)
(2, 0.4219200541111137)
(3, 0.28634244633530687)
(4, 0.21908362882066237)
(5, 0.17714871205622967)
(6, 0.14782525748927242)
(7, 0.1258448285184096)
(8, 0.10858011234267806)
(9, 0.0945554426837782)
(10, 0.08286985993320171)
(11, 0.0729381026838622)
(12, 0.064361302170273)
(13, 0.056857015663478665)
(14, 0.05021889174583558)
(15, 0.044292195197535286)
(16, 0.038958312787879756)
(17, 0.03412458863926907)
(18, 0.02971745000514782)
(19, 0.025677634549227744)
(20, 0.021956799812023094)
(21, 0.018515065442363803)
(22, 0.015319199358898377)
(23, 0.012341257516947922)
(24, 0.009557549043507233)
(25, 0.006947838592038834)
(26, 0.004494724222093593)
(27, 0.0021831469109018418)
(28, 0.0)
};
\addlegendentry{$\gamma=1.5$};

\addplot[thick, color={rgb,255:red,188; green,189; blue,34}] coordinates {
(1, 1.0)
(2, 0.06397313571428796)
(3, 0.020157063363464994)
(4, 0.009648909924076425)
(5, 0.005598137688874959)
(6, 0.0036247916155934517)
(7, 0.0025181844597644255)
(8, 0.001836259666356962)
(9, 0.001386603242547899)
(10, 0.0010745864764774725)
(11, 0.0008492795289726307)
(12, 0.0006812988583298678)
(13, 0.0005527282949533594)
(14, 0.0004521401299326952)
(15, 0.00037196697513078644)
(16, 0.0003070365569904105)
(17, 0.0002537165039997417)
(18, 0.00020939527658493697)
(19, 0.00017215631253846033)
(20, 0.00014056736259546962)
(21, 0.00011354072674222755)
(22, 9.023837683408465e-05)
(23, 7.000621634893099e-05)
(24, 5.2327682446762494e-05)
(25, 3.679044977835548e-05)
(26, 2.3062172458255407e-05)
(27, 1.0872565390375482e-05)
(28, 0.0)
};
\addlegendentry{$\gamma=3$};

\end{axis}
\end{tikzpicture}

%% file: data/plots/supp_search/grid_search_Llama32_3B_05.tex
\begin{tikzpicture}
                \begin{axis}[
                    width=6.5cm, height=6cm,
                    xlabel={$\kappa$},
                    ylabel={Wikitext Perplexity},
                    title={\gls{llama3}.2-3B},
                    legend style={at={(1.05,0.5)}, anchor=west},
                    grid=both,
                    ticklabel style={font=\small},
                    label style={font=\small},
                    ylabel style={yshift=-1.3em},
                ]
                \addplot[only marks, mark=x, color={rgb,255:red,31; green,119; blue,180}] table[row sep=\\] {
hgamma wikitext_ppl \\
5.0 54.38529201854991 \\
0.0 102.0027730826997 \\
4.5 53.96206231832687 \\
1.0 71.20941195067294 \\
8.0 51.09025374715125 \\
5.5 53.6468028730505 \\
10.0 51.19013690939979 \\
6.0 53.12545808585661 \\
};

\addplot[only marks, mark=x, color={rgb,255:red,255; green,127; blue,14}] table[row sep=\\] {
hgamma wikitext_ppl \\
5.0 47.80773755820942 \\
0.0 102.0027730826997 \\
1.0 67.41971617548265 \\
10.0 46.88305220915622 \\
8.0 47.06654728734683 \\
6.0 47.714454197477 \\
4.5 48.18269829109882 \\
4.0 49.03714759152334 \\
5.5 47.62135285300038 \\
};

\addplot[only marks, mark=x, color={rgb,255:red,44; green,160; blue,44}] table[row sep=\\] {
hgamma wikitext_ppl \\
5.0 47.15856396830611 \\
4.5 47.06654728734683 \\
6.0 44.82358155812516 \\
0.0 102.0027730826997 \\
10.0 44.30137101535166 \\
5.5 46.51820540231634 \\
4.0 47.90120329104558 \\
1.0 63.83170433047908 \\
8.0 45.086990684122 \\
};

\addplot[only marks, mark=x, color={rgb,255:red,214; green,39; blue,40}] table[row sep=\\] {
hgamma wikitext_ppl \\
1.0 66.11570082213345 \\
10.0 42.52108200006278 \\
4.0 47.62135285300038 \\
5.5 44.73612093895834 \\
0.0 102.0027730826997 \\
4.5 44.99901609609206 \\
8.0 42.77096031537661 \\
5.0 45.26345617625877 \\
6.0 44.64883097460398 \\
};

\addplot[only marks, mark=x, color={rgb,255:red,148; green,103; blue,189}] table[row sep=\\] {
hgamma wikitext_ppl \\
8.0 42.27266353442275 \\
6.0 42.68750480955783 \\
4.5 45.44061233480379 \\
5.5 43.27513086110299 \\
4.0 45.44061233480379 \\
10.0 42.52108200006278 \\
1.0 66.89505178172818 \\
5.0 44.82358155812516 \\
0.0 102.0027730826997 \\
};

\addplot[only marks, mark=x, color={rgb,255:red,140; green,86; blue,75}] table[row sep=\\] {
hgamma wikitext_ppl \\
5.5 45.26345617625877 \\
4.0 48.56059987746486 \\
4.5 46.70027251047953 \\
8.0 42.68750480955783 \\
5.0 46.246434773094 \\
0.0 102.0027730826997 \\
6.0 42.85457897951262 \\
10.0 42.68750480955783 \\
1.0 70.65525591042105 \\
};

\addplot[only marks, mark=x, color={rgb,255:red,227; green,119; blue,194}] table[row sep=\\] {
hgamma wikitext_ppl \\
4.5 48.18269829109882 \\
1.0 73.757394160674 \\
5.5 47.43569479289336 \\
0.0 102.0027730826997 \\
10.0 42.77096031537661 \\
8.0 45.61846186294896 \\
5.0 48.08868329990042 \\
6.0 46.246434773094 \\
4.0 49.03714759152334 \\
};

\addplot[only marks, mark=x, color={rgb,255:red,127; green,127; blue,127}] table[row sep=\\] {
hgamma wikitext_ppl \\
10.0 46.33684810653952 \\
4.0 58.57526450222689 \\
5.5 56.9952966774873 \\
8.0 48.56059987746486 \\
0.0 102.0027730826997 \\
6.0 53.12545808585661 \\
1.0 79.75075829332496 \\
4.5 61.38635013907805 \\
5.0 56.9952966774873 \\
};

\addplot[only marks, mark=x, color={rgb,255:red,188; green,189; blue,34}] table[row sep=\\] {
hgamma wikitext_ppl \\
4.0 88.27604237373367 \\
1.0 90.01713130052181 \\
5.5 86.90744918987926 \\
0.0 102.0027730826997 \\
4.5 85.89494673838158 \\
10.0 81.641993024771 \\
5.0 86.90744918987926 \\
8.0 82.60436300370014 \\
6.0 88.96839995507139 \\
};

\draw[dashed, line width=1pt] (0,598) -- (100,598);
\node[] at (50,615) {\scriptsize{performance without bias}};
\end{axis}
\end{tikzpicture}
\vspace{+3mm}

%% file: data/plots/supp_search/grid_search_Qwen3_4B_05.tex
\begin{tikzpicture}
                \begin{axis}[
                    width=6.5cm, height=6cm,
                    xlabel={$\kappa$},
                    ylabel={Wikitext Perplexity},
                    title={\gls{qwen3}-4B},
                    legend style={at={(1.05,0.5)}, anchor=west},
                    grid=both,
                    ticklabel style={font=\small},
                    label style={font=\small},
                    ylabel style={yshift=-1.3em},
                ]
                \addplot[only marks, mark=x, color={rgb,255:red,31; green,119; blue,180}] table[row sep=\\] {
hgamma wikitext_ppl \\
0.0 39.01953819342525 \\
6.0 35.80634097753966 \\
4.0 34.03344610288473 \\
8.0 35.32013242426123 \\
4.5 34.16664924473404 \\
5.5 35.52769382137767 \\
10.0 35.94648303373689 \\
1.0 34.16664924473404 \\
5.0 34.09998263323346 \\
};
\addlegendentry{$\gamma$ = -3.0}

\addplot[only marks, mark=x, color={rgb,255:red,255; green,127; blue,14}] table[row sep=\\] {
hgamma wikitext_ppl \\
0.0 39.01953819342525 \\
10.0 35.25121511458855 \\
5.0 34.09998263323346 \\
1.0 33.50580642730562 \\
6.0 34.70469568042362 \\
4.0 33.70270580214453 \\
4.5 32.92198346312246 \\
5.5 34.70469568042362 \\
8.0 34.4346215959791 \\
};
\addlegendentry{$\gamma$ = -1.5}

\addplot[only marks, mark=x, color={rgb,255:red,44; green,160; blue,44}] table[row sep=\\] {
hgamma wikitext_ppl \\
10.0 37.96705272185789 \\
4.0 32.79363281328884 \\
5.5 34.50194243789271 \\
1.0 34.3674321117641 \\
6.0 34.4346215959791 \\
4.5 33.24506217304801 \\
0.0 39.01953819342525 \\
8.0 36.58394105020859 \\
5.0 34.56939489431364 \\
};
\addlegendentry{$\gamma$ = -1.0}

\addplot[only marks, mark=x, color={rgb,255:red,214; green,39; blue,40}] table[row sep=\\] {
hgamma wikitext_ppl \\
5.0 34.70469568042362 \\
8.0 37.6715907734998 \\
10.0 38.11565156471445 \\
1.0 35.04526897307925 \\
5.5 36.37020880080946 \\
0.0 39.01953819342525 \\
4.0 33.05083646319603 \\
4.5 34.23344619169951 \\
6.0 36.58394105020859 \\
};
\addlegendentry{$\gamma$ = -0.5}

\addplot[only marks, mark=x, color={rgb,255:red,148; green,103; blue,189}] table[row sep=\\] {
hgamma wikitext_ppl \\
1.0 35.18243227767251 \\
0.0 39.01953819342525 \\
8.0 36.94295625024152 \\
6.0 36.29924256227225 \\
5.0 34.63697922255266 \\
4.0 32.47494113727571 \\
10.0 40.25815129472595 \\
5.5 35.66674528344124 \\
4.5 33.1801937796528 \\
};
\addlegendentry{$\gamma$ = 0.0}

\addplot[only marks, mark=x, color={rgb,255:red,140; green,86; blue,75}] table[row sep=\\] {
hgamma wikitext_ppl \\
5.0 34.16664924473404 \\
4.5 32.72964525738904 \\
4.0 33.70270580214453 \\
5.5 34.4346215959791 \\
8.0 37.23270344720572 \\
6.0 35.11378365112736 \\
0.0 39.01953819342525 \\
1.0 32.22221914035047 \\
10.0 38.26483200699776 \\
};
\addlegendentry{$\gamma$ = 0.5}

\addplot[only marks, mark=x, color={rgb,255:red,227; green,119; blue,194}] table[row sep=\\] {
hgamma wikitext_ppl \\
5.5 33.57131140436464 \\
4.5 32.47494113727571 \\
6.0 34.4346215959791 \\
8.0 34.63697922255266 \\
1.0 31.29190303178612 \\
4.0 31.9714638416083 \\
5.0 32.92198346312246 \\
0.0 39.01953819342525 \\
10.0 36.01675961554626 \\
};
\addlegendentry{$\gamma$ = 1.0}

\addplot[only marks, mark=x, color={rgb,255:red,127; green,127; blue,127}] table[row sep=\\] {
hgamma wikitext_ppl \\
10.0 33.3751796674398 \\
4.0 30.92734131898647 \\
6.0 31.16990746328828 \\
0.0 39.01953819342525 \\
5.5 31.9714638416083 \\
1.0 30.26997357328364 \\
4.5 31.29190303178612 \\
8.0 32.60204446315635 \\
5.0 31.6607620826941 \\
};
\addlegendentry{$\gamma$ = 1.5}

\addplot[only marks, mark=x, color={rgb,255:red,188; green,189; blue,34}] table[row sep=\\] {
hgamma wikitext_ppl \\
5.5 30.32915238841817 \\
0.0 39.01953819342525 \\
1.0 31.41437607743883 \\
4.0 31.29190303178612 \\
8.0 30.50738391845551 \\
6.0 30.32915238841817 \\
5.0 30.38844690012418 \\
10.0 30.62678644311483 \\
4.5 30.44785733459232 \\
};
\addlegendentry{$\gamma$ = 3.0}
\draw[dashed, line width=1pt] (0,875) -- (100,875);
\node[] at (50,909) {\scriptsize{performance without bias}};
\end{axis}
\end{tikzpicture}

%% file: data/plots/supp_search/scale_metric_ablation_llama3_3b.tex
\begin{tikzpicture}
    \definecolor{col0}{HTML}{1f77b4}
    \definecolor{col1}{HTML}{ff7f0e}
    \definecolor{col2}{HTML}{2ca02c}
    \definecolor{col3}{HTML}{d62728}
    \definecolor{col4}{HTML}{9467bd}
    \definecolor{col5}{HTML}{8c564b}
    \definecolor{col6}{HTML}{e377c2}
    \definecolor{col7}{HTML}{7f7f7f}
    \definecolor{col8}{HTML}{bcbd22}
    \definecolor{col9}{HTML}{17becf}
        \begin{groupplot}[
        group style={
            group size=4 by 1,
            horizontal sep=1.8cm,
            vertical sep=2cm,
        },
        width=5.5cm, height=5cm,
        xlabel={Remaining ratio (\%)},
        grid=major, grid style={dashed, gray!30},
        legend style={draw=none},
    ]
\nextgroupplot[title={Energy}, title style={yshift=-1.5ex}, ylabel={Sensitivity}]
        \addplot[color=col0, mark=*, thick, mark options={solid}] coordinates { (0.09982094897045658, 4503616.5000) (0.20008952551477172, 3371018.7500) (0.2999104744852283, 2580218.0000) (0.40017905102954343, 1980106.6250) (0.5, 1514024.6250) (0.5998209489704566, 1144215.3750) (0.7000895255147717, 847392.6875) (0.7999104744852283, 611660.6875) (0.9001790510295434, 423828.0000)  };
        \addplot[color=col1, mark=triangle*, thick, mark options={solid}] coordinates { (0.09982094897045658, 1299298.6250) (0.20008952551477172, 893914.6875) (0.2999104744852283, 658189.3125) (0.40017905102954343, 494344.0000) (0.5, 373054.6562) (0.5998209489704566, 279676.0000) (0.7000895255147717, 206124.0000) (0.7999104744852283, 148341.3281) (0.9001790510295434, 102681.3359)  };
        \addplot[color=col2, mark=triangle*, thick, mark options={solid}] coordinates { (0.09982094897045658, 2986825.2500) (0.20008952551477172, 2129376.0000) (0.2999104744852283, 1585612.0000) (0.40017905102954343, 1192852.6250) (0.5, 897990.6875) (0.5998209489704566, 670204.6875) (0.7000895255147717, 491517.3438) (0.7999104744852283, 352014.0000) (0.9001790510295434, 242484.0000)  };
        \addplot[color=col3, mark=square*, thick, mark options={solid}] coordinates { (0.10026041666666667, 928856.0000) (0.20052083333333334, 516736.0000) (0.2994791666666667, 340104.0000) (0.3997395833333333, 232736.0000) (0.5, 161304.0000) (0.6002604166666666, 111432.0000) (0.7005208333333334, 75816.0000) (0.7994791666666666, 50280.0000) (0.8997395833333334, 31624.0000)  };
        \addplot[color=col4, mark=diamond*, thick, mark options={solid}] coordinates { (0.10026041666666667, 1063402.8750) (0.19986979166666666, 706434.3125) (0.3001302083333333, 502726.5000) (0.3997395833333333, 367865.0000) (0.5, 270785.3438) (0.6002604166666666, 199179.3281) (0.6998697916666666, 146048.3281) (0.8001302083333334, 105869.3359) (0.8997395833333334, 75792.6641)  };
        \addplot[color=col5, mark=pentagon*, thick, mark options={solid}] coordinates { (0.10026041666666667, 338233.3438) (0.19986979166666666, 198188.0000) (0.3001302083333333, 131424.0000) (0.3997395833333333, 92040.0000) (0.5, 65816.0000) (0.6002604166666666, 47446.6680) (0.6998697916666666, 34237.3320) (0.8001302083333334, 24461.3340) (0.8997395833333334, 17270.6660)  };
        \addplot[color=col6, mark=x, thick, mark options={solid}] coordinates { (0.10026041666666667, 1593693.7500) (0.20052083333333334, 1158600.0000) (0.2994791666666667, 876962.0000) (0.3997395833333333, 669018.0000) (0.5, 510067.7500) (0.6002604166666666, 385469.0000) (0.7005208333333334, 286210.0000) (0.7994791666666666, 207100.7500) (0.8997395833333334, 142922.5000)  };
\nextgroupplot[title={Perplexity}, title style={yshift=-1.5ex}, ylabel={ }]
        \addplot[name path=U10, draw=none, forget plot] coordinates { (0.1, 9.4876) (0.2, 9.2332) (0.3, 9.0624) (0.4, 8.9744) (0.5, 8.9045) (0.6, 8.8506) (0.7, 8.8161) (0.8, 8.7879) (0.9, 8.7707)  };
        \addplot[name path=L10, draw=none, forget plot] coordinates { (0.1, 8.4556) (0.2, 8.2423) (0.3, 8.0996) (0.4, 8.0209) (0.5, 7.9585) (0.6, 7.9214) (0.7, 7.8905) (0.8, 7.8619) (0.9, 7.8466)  };
        \addplot[col0, opacity=0.15, forget plot] fill between[of=U10 and L10];
        \addplot[color=col0, mark=*, thick, mark options={solid}] coordinates { (0.1, 8.9716) (0.2, 8.7377) (0.3, 8.5810) (0.4, 8.4976) (0.5, 8.4315) (0.6, 8.3860) (0.7, 8.3533) (0.8, 8.3249) (0.9, 8.3087)  };
        \addplot[name path=U11, draw=none, forget plot] coordinates { (0.1, 9.0921) (0.2, 8.8969) (0.3, 8.8276) (0.4, 8.7912) (0.5, 8.7741) (0.6, 8.7665) (0.7, 8.7570) (0.8, 8.7474) (0.9, 8.7474)  };
        \addplot[name path=L11, draw=none, forget plot] coordinates { (0.1, 8.1127) (0.2, 7.9498) (0.3, 7.8880) (0.4, 7.8666) (0.5, 7.8513) (0.6, 7.8351) (0.7, 7.8359) (0.8, 7.8291) (0.9, 7.8291)  };
        \addplot[col1, opacity=0.15, forget plot] fill between[of=U11 and L11];
        \addplot[color=col1, mark=triangle*, thick, mark options={solid}] coordinates { (0.1, 8.6024) (0.2, 8.4233) (0.3, 8.3578) (0.4, 8.3289) (0.5, 8.3127) (0.6, 8.3008) (0.7, 8.2965) (0.8, 8.2883) (0.9, 8.2883)  };
        \addplot[name path=U12, draw=none, forget plot] coordinates { (0.1, 9.2548) (0.2, 9.0370) (0.3, 8.9297) (0.4, 8.8795) (0.5, 8.8292) (0.6, 8.8023) (0.7, 8.7851) (0.8, 8.7679) (0.9, 8.7570)  };
        \addplot[name path=L12, draw=none, forget plot] coordinates { (0.1, 8.2633) (0.2, 8.0750) (0.3, 7.9905) (0.4, 7.9343) (0.5, 7.9020) (0.6, 7.8798) (0.7, 7.8644) (0.8, 7.8491) (0.9, 7.8359)  };
        \addplot[col2, opacity=0.15, forget plot] fill between[of=U12 and L12];
        \addplot[color=col2, mark=triangle*, thick, mark options={solid}] coordinates { (0.1, 8.7591) (0.2, 8.5560) (0.3, 8.4601) (0.4, 8.4069) (0.5, 8.3656) (0.6, 8.3410) (0.7, 8.3248) (0.8, 8.3085) (0.9, 8.2965)  };
        \addplot[name path=U13, draw=none, forget plot] coordinates { (0.1, 8.8085) (0.2, 8.7803) (0.3, 8.7613) (0.4, 8.7631) (0.5, 8.7631) (0.6, 8.7474) (0.7, 8.7474) (0.8, 8.7474) (0.9, 8.7474)  };
        \addplot[name path=L13, draw=none, forget plot] coordinates { (0.1, 7.8742) (0.2, 7.8457) (0.3, 7.8397) (0.4, 7.8304) (0.5, 7.8304) (0.6, 7.8291) (0.7, 7.8291) (0.8, 7.8291) (0.9, 7.8291)  };
        \addplot[col3, opacity=0.15, forget plot] fill between[of=U13 and L13];
        \addplot[color=col3, mark=square*, thick, mark options={solid}] coordinates { (0.1, 8.3413) (0.2, 8.3130) (0.3, 8.3005) (0.4, 8.2968) (0.5, 8.2968) (0.6, 8.2883) (0.7, 8.2883) (0.8, 8.2883) (0.9, 8.2883)  };
        \addplot[name path=U14, draw=none, forget plot] coordinates { (0.1, 8.8449) (0.2, 8.7836) (0.3, 8.7570) (0.4, 8.7474) (0.5, 8.7494) (0.6, 8.7494) (0.7, 8.7494) (0.8, 8.7494) (0.9, 8.7494)  };
        \addplot[name path=L14, draw=none, forget plot] coordinates { (0.1, 7.9034) (0.2, 7.8504) (0.3, 7.8359) (0.4, 7.8291) (0.5, 7.8198) (0.6, 7.8198) (0.7, 7.8198) (0.8, 7.8198) (0.9, 7.8198)  };
        \addplot[col4, opacity=0.15, forget plot] fill between[of=U14 and L14];
        \addplot[color=col4, mark=diamond*, thick, mark options={solid}] coordinates { (0.1, 8.3741) (0.2, 8.3170) (0.3, 8.2965) (0.4, 8.2883) (0.5, 8.2846) (0.6, 8.2846) (0.7, 8.2846) (0.8, 8.2846) (0.9, 8.2846)  };
        \addplot[name path=U15, draw=none, forget plot] coordinates { (0.1, 8.7679) (0.2, 8.7570) (0.3, 8.7474) (0.4, 8.7474) (0.5, 8.7494) (0.6, 8.7494) (0.7, 8.7494) (0.8, 8.7494) (0.9, 8.7494)  };
        \addplot[name path=L15, draw=none, forget plot] coordinates { (0.1, 7.8491) (0.2, 7.8359) (0.3, 7.8291) (0.4, 7.8291) (0.5, 7.8198) (0.6, 7.8198) (0.7, 7.8198) (0.8, 7.8198) (0.9, 7.8198)  };
        \addplot[col5, opacity=0.15, forget plot] fill between[of=U15 and L15];
        \addplot[color=col5, mark=pentagon*, thick, mark options={solid}] coordinates { (0.1, 8.3085) (0.2, 8.2965) (0.3, 8.2883) (0.4, 8.2883) (0.5, 8.2846) (0.6, 8.2846) (0.7, 8.2846) (0.8, 8.2846) (0.9, 8.2846)  };
        \addplot[name path=U16, draw=none, forget plot] coordinates { (0.1, 8.9219) (0.2, 8.8352) (0.3, 8.7912) (0.4, 8.7761) (0.5, 8.7665) (0.6, 8.7570) (0.7, 8.7474) (0.8, 8.7474) (0.9, 8.7494)  };
        \addplot[name path=L16, draw=none, forget plot] coordinates { (0.1, 7.9740) (0.2, 7.8965) (0.3, 7.8666) (0.4, 7.8419) (0.5, 7.8351) (0.6, 7.8359) (0.7, 7.8291) (0.8, 7.8291) (0.9, 7.8198)  };
        \addplot[col6, opacity=0.15, forget plot] fill between[of=U16 and L16];
        \addplot[color=col6, mark=x, thick, mark options={solid}] coordinates { (0.1, 8.4480) (0.2, 8.3659) (0.3, 8.3289) (0.4, 8.3090) (0.5, 8.3008) (0.6, 8.2965) (0.7, 8.2883) (0.8, 8.2883) (0.9, 8.2846)  };
\nextgroupplot[title={MSE}, title style={yshift=-1.5ex}, ylabel={ }]
        \addplot[name path=U20, draw=none, forget plot] coordinates { (0.1, 0.0587) (0.2, 0.0423) (0.3, 0.0312) (0.4, 0.0232) (0.5, 0.0172) (0.6, 0.0127) (0.7, 0.0092) (0.8, 0.0065) (0.9, 0.0045)  };
        \addplot[name path=L20, draw=none, forget plot] coordinates { (0.1, 0.0570) (0.2, 0.0413) (0.3, 0.0306) (0.4, 0.0229) (0.5, 0.0170) (0.6, 0.0125) (0.7, 0.0090) (0.8, 0.0064) (0.9, 0.0044)  };
        \addplot[col0, opacity=0.15, forget plot] fill between[of=U20 and L20];
        \addplot[color=col0, mark=*, thick, mark options={solid}] coordinates { (0.1, 0.0579) (0.2, 0.0418) (0.3, 0.0309) (0.4, 0.0230) (0.5, 0.0171) (0.6, 0.0126) (0.7, 0.0091) (0.8, 0.0065) (0.9, 0.0044)  };
        \addplot[name path=U21, draw=none, forget plot] coordinates { (0.1, 0.0297) (0.2, 0.0152) (0.3, 0.0091) (0.4, 0.0059) (0.5, 0.0039) (0.6, 0.0026) (0.7, 0.0018) (0.8, 0.0012) (0.9, 0.0008)  };
        \addplot[name path=L21, draw=none, forget plot] coordinates { (0.1, 0.0292) (0.2, 0.0144) (0.3, 0.0085) (0.4, 0.0054) (0.5, 0.0036) (0.6, 0.0024) (0.7, 0.0017) (0.8, 0.0011) (0.9, 0.0007)  };
        \addplot[col1, opacity=0.15, forget plot] fill between[of=U21 and L21];
        \addplot[color=col1, mark=triangle*, thick, mark options={solid}] coordinates { (0.1, 0.0294) (0.2, 0.0148) (0.3, 0.0088) (0.4, 0.0056) (0.5, 0.0038) (0.6, 0.0025) (0.7, 0.0017) (0.8, 0.0011) (0.9, 0.0008)  };
        \addplot[name path=U22, draw=none, forget plot] coordinates { (0.1, 0.0427) (0.2, 0.0271) (0.3, 0.0186) (0.4, 0.0131) (0.5, 0.0092) (0.6, 0.0065) (0.7, 0.0045) (0.8, 0.0030) (0.9, 0.0020)  };
        \addplot[name path=L22, draw=none, forget plot] coordinates { (0.1, 0.0418) (0.2, 0.0267) (0.3, 0.0183) (0.4, 0.0129) (0.5, 0.0090) (0.6, 0.0063) (0.7, 0.0044) (0.8, 0.0029) (0.9, 0.0019)  };
        \addplot[col2, opacity=0.15, forget plot] fill between[of=U22 and L22];
        \addplot[color=col2, mark=triangle*, thick, mark options={solid}] coordinates { (0.1, 0.0423) (0.2, 0.0269) (0.3, 0.0185) (0.4, 0.0130) (0.5, 0.0091) (0.6, 0.0064) (0.7, 0.0044) (0.8, 0.0030) (0.9, 0.0020)  };
        \addplot[name path=U23, draw=none, forget plot] coordinates { (0.1, 0.0077) (0.2, 0.0048) (0.3, 0.0033) (0.4, 0.0024) (0.5, 0.0019) (0.6, 0.0015) (0.7, 0.0011) (0.8, 0.0008) (0.9, 0.0005)  };
        \addplot[name path=L23, draw=none, forget plot] coordinates { (0.1, 0.0071) (0.2, 0.0043) (0.3, 0.0030) (0.4, 0.0022) (0.5, 0.0017) (0.6, 0.0014) (0.7, 0.0010) (0.8, 0.0007) (0.9, 0.0005)  };
        \addplot[col3, opacity=0.15, forget plot] fill between[of=U23 and L23];
        \addplot[color=col3, mark=square*, thick, mark options={solid}] coordinates { (0.1, 0.0074) (0.2, 0.0046) (0.3, 0.0031) (0.4, 0.0023) (0.5, 0.0018) (0.6, 0.0014) (0.7, 0.0010) (0.8, 0.0008) (0.9, 0.0005)  };
        \addplot[name path=U24, draw=none, forget plot] coordinates { (0.1, 0.0126) (0.2, 0.0063) (0.3, 0.0039) (0.4, 0.0025) (0.5, 0.0017) (0.6, 0.0012) (0.7, 0.0008) (0.8, 0.0006) (0.9, 0.0004)  };
        \addplot[name path=L24, draw=none, forget plot] coordinates { (0.1, 0.0117) (0.2, 0.0059) (0.3, 0.0036) (0.4, 0.0023) (0.5, 0.0016) (0.6, 0.0011) (0.7, 0.0008) (0.8, 0.0005) (0.9, 0.0004)  };
        \addplot[col4, opacity=0.15, forget plot] fill between[of=U24 and L24];
        \addplot[color=col4, mark=diamond*, thick, mark options={solid}] coordinates { (0.1, 0.0121) (0.2, 0.0061) (0.3, 0.0037) (0.4, 0.0024) (0.5, 0.0016) (0.6, 0.0011) (0.7, 0.0008) (0.8, 0.0006) (0.9, 0.0004)  };
        \addplot[name path=U25, draw=none, forget plot] coordinates { (0.1, 0.0027) (0.2, 0.0013) (0.3, 0.0007) (0.4, 0.0005) (0.5, 0.0003) (0.6, 0.0002) (0.7, 0.0002) (0.8, 0.0001) (0.9, 0.0001)  };
        \addplot[name path=L25, draw=none, forget plot] coordinates { (0.1, 0.0026) (0.2, 0.0012) (0.3, 0.0007) (0.4, 0.0004) (0.5, 0.0003) (0.6, 0.0002) (0.7, 0.0001) (0.8, 0.0001) (0.9, 0.0001)  };
        \addplot[col5, opacity=0.15, forget plot] fill between[of=U25 and L25];
        \addplot[color=col5, mark=pentagon*, thick, mark options={solid}] coordinates { (0.1, 0.0026) (0.2, 0.0013) (0.3, 0.0007) (0.4, 0.0005) (0.5, 0.0003) (0.6, 0.0002) (0.7, 0.0002) (0.8, 0.0001) (0.9, 0.0001)  };
        \addplot[name path=U26, draw=none, forget plot] coordinates { (0.1, 0.0197) (0.2, 0.0117) (0.3, 0.0077) (0.4, 0.0054) (0.5, 0.0039) (0.6, 0.0028) (0.7, 0.0020) (0.8, 0.0014) (0.9, 0.0010)  };
        \addplot[name path=L26, draw=none, forget plot] coordinates { (0.1, 0.0183) (0.2, 0.0111) (0.3, 0.0072) (0.4, 0.0049) (0.5, 0.0035) (0.6, 0.0025) (0.7, 0.0018) (0.8, 0.0013) (0.9, 0.0009)  };
        \addplot[col6, opacity=0.15, forget plot] fill between[of=U26 and L26];
        \addplot[color=col6, mark=x, thick, mark options={solid}] coordinates { (0.1, 0.0190) (0.2, 0.0114) (0.3, 0.0075) (0.4, 0.0052) (0.5, 0.0037) (0.6, 0.0027) (0.7, 0.0019) (0.8, 0.0014) (0.9, 0.0009)  };
\nextgroupplot[title={KL-Divergence}, title style={yshift=-1.5ex}, ylabel={ }]
        \addplot[name path=U30, draw=none, forget plot] coordinates { (0.1, 0.1263) (0.2, 0.0838) (0.3, 0.0584) (0.4, 0.0407) (0.5, 0.0285) (0.6, 0.0199) (0.7, 0.0141) (0.8, 0.0097) (0.9, 0.0064)  };
        \addplot[name path=L30, draw=none, forget plot] coordinates { (0.1, 0.1175) (0.2, 0.0788) (0.3, 0.0545) (0.4, 0.0382) (0.5, 0.0264) (0.6, 0.0186) (0.7, 0.0129) (0.8, 0.0088) (0.9, 0.0060)  };
        \addplot[col0, opacity=0.15, forget plot] fill between[of=U30 and L30];
        \addplot[color=col0, mark=*, thick, mark options={solid}] coordinates { (0.1, 0.1219) (0.2, 0.0813) (0.3, 0.0564) (0.4, 0.0395) (0.5, 0.0275) (0.6, 0.0193) (0.7, 0.0135) (0.8, 0.0092) (0.9, 0.0062)  };
        \addplot[name path=U31, draw=none, forget plot] coordinates { (0.1, 0.0632) (0.2, 0.0295) (0.3, 0.0170) (0.4, 0.0108) (0.5, 0.0071) (0.6, 0.0046) (0.7, 0.0031) (0.8, 0.0020) (0.9, 0.0013)  };
        \addplot[name path=L31, draw=none, forget plot] coordinates { (0.1, 0.0578) (0.2, 0.0273) (0.3, 0.0156) (0.4, 0.0097) (0.5, 0.0063) (0.6, 0.0042) (0.7, 0.0028) (0.8, 0.0019) (0.9, 0.0012)  };
        \addplot[col1, opacity=0.15, forget plot] fill between[of=U31 and L31];
        \addplot[color=col1, mark=triangle*, thick, mark options={solid}] coordinates { (0.1, 0.0605) (0.2, 0.0284) (0.3, 0.0163) (0.4, 0.0103) (0.5, 0.0067) (0.6, 0.0044) (0.7, 0.0029) (0.8, 0.0019) (0.9, 0.0013)  };
        \addplot[name path=U32, draw=none, forget plot] coordinates { (0.1, 0.0848) (0.2, 0.0496) (0.3, 0.0327) (0.4, 0.0222) (0.5, 0.0150) (0.6, 0.0102) (0.7, 0.0069) (0.8, 0.0046) (0.9, 0.0030)  };
        \addplot[name path=L32, draw=none, forget plot] coordinates { (0.1, 0.0785) (0.2, 0.0466) (0.3, 0.0305) (0.4, 0.0206) (0.5, 0.0141) (0.6, 0.0097) (0.7, 0.0066) (0.8, 0.0044) (0.9, 0.0029)  };
        \addplot[col2, opacity=0.15, forget plot] fill between[of=U32 and L32];
        \addplot[color=col2, mark=triangle*, thick, mark options={solid}] coordinates { (0.1, 0.0817) (0.2, 0.0481) (0.3, 0.0316) (0.4, 0.0214) (0.5, 0.0145) (0.6, 0.0099) (0.7, 0.0068) (0.8, 0.0045) (0.9, 0.0030)  };
        \addplot[name path=U33, draw=none, forget plot] coordinates { (0.1, 0.0129) (0.2, 0.0079) (0.3, 0.0056) (0.4, 0.0043) (0.5, 0.0034) (0.6, 0.0026) (0.7, 0.0019) (0.8, 0.0014) (0.9, 0.0010)  };
        \addplot[name path=L33, draw=none, forget plot] coordinates { (0.1, 0.0121) (0.2, 0.0073) (0.3, 0.0052) (0.4, 0.0040) (0.5, 0.0031) (0.6, 0.0025) (0.7, 0.0018) (0.8, 0.0013) (0.9, 0.0009)  };
        \addplot[col3, opacity=0.15, forget plot] fill between[of=U33 and L33];
        \addplot[color=col3, mark=square*, thick, mark options={solid}] coordinates { (0.1, 0.0125) (0.2, 0.0076) (0.3, 0.0054) (0.4, 0.0042) (0.5, 0.0032) (0.6, 0.0025) (0.7, 0.0019) (0.8, 0.0014) (0.9, 0.0009)  };
        \addplot[name path=U34, draw=none, forget plot] coordinates { (0.1, 0.0238) (0.2, 0.0113) (0.3, 0.0065) (0.4, 0.0041) (0.5, 0.0027) (0.6, 0.0018) (0.7, 0.0013) (0.8, 0.0009) (0.9, 0.0006)  };
        \addplot[name path=L34, draw=none, forget plot] coordinates { (0.1, 0.0228) (0.2, 0.0107) (0.3, 0.0062) (0.4, 0.0039) (0.5, 0.0026) (0.6, 0.0017) (0.7, 0.0012) (0.8, 0.0008) (0.9, 0.0006)  };
        \addplot[col4, opacity=0.15, forget plot] fill between[of=U34 and L34];
        \addplot[color=col4, mark=diamond*, thick, mark options={solid}] coordinates { (0.1, 0.0233) (0.2, 0.0110) (0.3, 0.0064) (0.4, 0.0040) (0.5, 0.0026) (0.6, 0.0018) (0.7, 0.0012) (0.8, 0.0008) (0.9, 0.0006)  };
        \addplot[name path=U35, draw=none, forget plot] coordinates { (0.1, 0.0057) (0.2, 0.0027) (0.3, 0.0015) (0.4, 0.0010) (0.5, 0.0007) (0.6, 0.0005) (0.7, 0.0003) (0.8, 0.0002) (0.9, 0.0002)  };
        \addplot[name path=L35, draw=none, forget plot] coordinates { (0.1, 0.0056) (0.2, 0.0026) (0.3, 0.0015) (0.4, 0.0009) (0.5, 0.0006) (0.6, 0.0004) (0.7, 0.0003) (0.8, 0.0002) (0.9, 0.0002)  };
        \addplot[col5, opacity=0.15, forget plot] fill between[of=U35 and L35];
        \addplot[color=col5, mark=pentagon*, thick, mark options={solid}] coordinates { (0.1, 0.0056) (0.2, 0.0026) (0.3, 0.0015) (0.4, 0.0009) (0.5, 0.0006) (0.6, 0.0005) (0.7, 0.0003) (0.8, 0.0002) (0.9, 0.0002)  };
        \addplot[name path=U36, draw=none, forget plot] coordinates { (0.1, 0.0368) (0.2, 0.0211) (0.3, 0.0134) (0.4, 0.0090) (0.5, 0.0064) (0.6, 0.0045) (0.7, 0.0032) (0.8, 0.0022) (0.9, 0.0015)  };
        \addplot[name path=L36, draw=none, forget plot] coordinates { (0.1, 0.0354) (0.2, 0.0200) (0.3, 0.0127) (0.4, 0.0085) (0.5, 0.0058) (0.6, 0.0041) (0.7, 0.0029) (0.8, 0.0021) (0.9, 0.0014)  };
        \addplot[col6, opacity=0.15, forget plot] fill between[of=U36 and L36];
        \addplot[color=col6, mark=x, thick, mark options={solid}] coordinates { (0.1, 0.0361) (0.2, 0.0206) (0.3, 0.0130) (0.4, 0.0088) (0.5, 0.0061) (0.6, 0.0043) (0.7, 0.0031) (0.8, 0.0021) (0.9, 0.0015)  };
    \end{groupplot}

\end{tikzpicture}

%% file: data/plots/supp_search/scale_metric_ablation_qwen3_4b.tex
\begin{tikzpicture}
    \definecolor{col0}{HTML}{1f77b4}
    \definecolor{col1}{HTML}{ff7f0e}
    \definecolor{col2}{HTML}{2ca02c}
    \definecolor{col3}{HTML}{d62728}
    \definecolor{col4}{HTML}{9467bd}
    \definecolor{col5}{HTML}{8c564b}
    \definecolor{col6}{HTML}{e377c2}
    \definecolor{col7}{HTML}{7f7f7f}
    \definecolor{col8}{HTML}{bcbd22}
    \definecolor{col9}{HTML}{17becf}
        \begin{groupplot}[
        group style={
            group size=4 by 1,
            horizontal sep=1.8cm,
            vertical sep=2cm,
        },
        width=5.5cm, height=5cm,
        xlabel={Remaining ratio},
        grid=major, grid style={dashed, gray!30},
        legend style={draw=none},
    ]
\nextgroupplot[title={Energy}, title style={yshift=-1.5ex}, ylabel={Sensitivity}]
        \addplot[color=col0, mark=*, thick, mark options={solid}] coordinates { (0.10019743336623889, 11609211.0000) (0.19990128331688056, 8319213.0000) (0.30009871668311944, 6187403.0000) (0.3998025666337611, 4654894.5000) (0.5, 3485897.5000) (0.6001974333662389, 2576310.5000) (0.6999012833168805, 1863472.0000) (0.8000987166831195, 1296825.6250) (0.8998025666337611, 853017.6250)  };
        \addplot[color=col1, mark=triangle*, thick, mark options={solid}] coordinates { (0.10019743336623889, 6164662.5000) (0.19990128331688056, 4360102.5000) (0.30009871668311944, 3258489.5000) (0.3998025666337611, 2477641.5000) (0.5, 1880985.6250) (0.6001974333662389, 1412499.2500) (0.6999012833168805, 1040134.3750) (0.8000987166831195, 738969.6250) (0.8998025666337611, 498246.4062)  };
        \addplot[color=col2, mark=triangle*, thick, mark options={solid}] coordinates { (0.10019743336623889, 10499866.0000) (0.19990128331688056, 7414947.0000) (0.30009871668311944, 5485939.0000) (0.3998025666337611, 4117104.0000) (0.5, 3078019.2500) (0.6001974333662389, 2274419.2500) (0.6999012833168805, 1646617.6250) (0.8000987166831195, 1147737.6250) (0.8998025666337611, 757484.8125)  };
        \addplot[color=col3, mark=square*, thick, mark options={solid}] coordinates { (0.09986320109439124, 5018160.0000) (0.1997264021887825, 3018128.0000) (0.29958960328317374, 1997952.0000) (0.399452804377565, 1371152.0000) (0.4993160054719562, 956864.0000) (0.600547195622435, 665760.0000) (0.7004103967168263, 459568.0000) (0.8002735978112175, 309848.0000) (0.9001367989056087, 201496.0000)  };
        \addplot[color=col4, mark=diamond*, thick, mark options={solid}] coordinates { (0.10031746031746032, 5608200.0000) (0.2, 3610368.0000) (0.29968253968253966, 2506484.7500) (0.4, 1786075.2500) (0.5003174603174603, 1287214.3750) (0.6, 931600.0000) (0.6996825396825397, 671182.3750) (0.8, 477640.0000) (0.9003174603174603, 334062.4062)  };
        \addplot[color=col5, mark=pentagon*, thick, mark options={solid}] coordinates { (0.10031746031746032, 3152742.5000) (0.2, 1781004.7500) (0.29968253968253966, 1175872.0000) (0.4, 825024.0000) (0.5003174603174603, 595328.0000) (0.6, 434585.5938) (0.6996825396825397, 316774.4062) (0.8, 228249.5938) (0.9003174603174603, 161779.2031)  };
        \addplot[color=col6, mark=x, thick, mark options={solid}] coordinates { (0.09986320109439124, 9471220.0000) (0.1997264021887825, 6089448.0000) (0.29958960328317374, 4271760.0000) (0.399452804377565, 3079020.0000) (0.4993160054719562, 2234624.0000) (0.600547195622435, 1609088.0000) (0.7004103967168263, 1151560.0000) (0.8002735978112175, 807800.0000) (0.9001367989056087, 549216.0000)  };
\nextgroupplot[title={Perplexity}, title style={yshift=-1.5ex}, ylabel={ }]
        \addplot[name path=U10, draw=none, forget plot] coordinates { (0.1, 16.0655) (0.2, 16.0607) (0.3, 16.0947) (0.4, 16.0901) (0.5, 16.1166) (0.6, 16.1257) (0.7, 16.1398) (0.8, 16.1375) (0.9, 16.1283)  };
        \addplot[name path=L10, draw=none, forget plot] coordinates { (0.1, 13.8135) (0.2, 13.8041) (0.3, 13.8559) (0.4, 13.8193) (0.5, 13.8370) (0.6, 13.8428) (0.7, 13.8579) (0.8, 13.8733) (0.9, 13.8676)  };
        \addplot[col0, opacity=0.15, forget plot] fill between[of=U10 and L10];
        \addplot[color=col0, mark=*, thick, mark options={solid}] coordinates { (0.1, 14.9395) (0.2, 14.9324) (0.3, 14.9753) (0.4, 14.9547) (0.5, 14.9768) (0.6, 14.9842) (0.7, 14.9988) (0.8, 15.0054) (0.9, 14.9980)  };
        \addplot[name path=U11, draw=none, forget plot] coordinates { (0.1, 16.0890) (0.2, 16.0576) (0.3, 16.0936) (0.4, 16.0959) (0.5, 16.1131) (0.6, 16.1156) (0.7, 16.1156) (0.8, 16.1156) (0.9, 16.1156)  };
        \addplot[name path=L11, draw=none, forget plot] coordinates { (0.1, 13.8745) (0.2, 13.8474) (0.3, 13.8706) (0.4, 13.8551) (0.5, 13.8672) (0.6, 13.8922) (0.7, 13.8922) (0.8, 13.8922) (0.9, 13.8922)  };
        \addplot[col1, opacity=0.15, forget plot] fill between[of=U11 and L11];
        \addplot[color=col1, mark=triangle*, thick, mark options={solid}] coordinates { (0.1, 14.9818) (0.2, 14.9525) (0.3, 14.9821) (0.4, 14.9755) (0.5, 14.9902) (0.6, 15.0039) (0.7, 15.0039) (0.8, 15.0039) (0.9, 15.0039)  };
        \addplot[name path=U12, draw=none, forget plot] coordinates { (0.1, 15.9946) (0.2, 16.0261) (0.3, 16.0234) (0.4, 16.1055) (0.5, 16.1250) (0.6, 16.1156) (0.7, 16.1156) (0.8, 16.1156) (0.9, 16.1109)  };
        \addplot[name path=L12, draw=none, forget plot] coordinates { (0.1, 13.8480) (0.2, 13.8612) (0.3, 13.8501) (0.4, 13.9010) (0.5, 13.8977) (0.6, 13.8922) (0.7, 13.8922) (0.8, 13.8922) (0.9, 13.8826)  };
        \addplot[col2, opacity=0.15, forget plot] fill between[of=U12 and L12];
        \addplot[color=col2, mark=triangle*, thick, mark options={solid}] coordinates { (0.1, 14.9213) (0.2, 14.9437) (0.3, 14.9367) (0.4, 15.0033) (0.5, 15.0114) (0.6, 15.0039) (0.7, 15.0039) (0.8, 15.0039) (0.9, 14.9968)  };
        \addplot[name path=U13, draw=none, forget plot] coordinates { (0.1, 15.8545) (0.2, 15.9880) (0.3, 16.0702) (0.4, 16.0728) (0.5, 16.1357) (0.6, 16.1016) (0.7, 16.0389) (0.8, 16.0389) (0.9, 16.0702)  };
        \addplot[name path=L13, draw=none, forget plot] coordinates { (0.1, 13.6864) (0.2, 13.7994) (0.3, 13.8499) (0.4, 13.8748) (0.5, 13.9291) (0.6, 13.8770) (0.7, 13.8229) (0.8, 13.8229) (0.9, 13.8499)  };
        \addplot[col3, opacity=0.15, forget plot] fill between[of=U13 and L13];
        \addplot[color=col3, mark=square*, thick, mark options={solid}] coordinates { (0.1, 14.7705) (0.2, 14.8937) (0.3, 14.9601) (0.4, 14.9738) (0.5, 15.0324) (0.6, 14.9893) (0.7, 14.9309) (0.8, 14.9309) (0.9, 14.9601)  };
        \addplot[name path=U14, draw=none, forget plot] coordinates { (0.1, 16.2526) (0.2, 16.1802) (0.3, 16.1603) (0.4, 16.1288) (0.5, 16.1487) (0.6, 16.1379) (0.7, 16.1240) (0.8, 16.1261) (0.9, 16.1261)  };
        \addplot[name path=L14, draw=none, forget plot] coordinates { (0.1, 13.9919) (0.2, 13.9314) (0.3, 13.9350) (0.4, 13.9078) (0.5, 13.9042) (0.6, 13.9137) (0.7, 13.8983) (0.8, 13.8830) (0.9, 13.8830)  };
        \addplot[col4, opacity=0.15, forget plot] fill between[of=U14 and L14];
        \addplot[color=col4, mark=diamond*, thick, mark options={solid}] coordinates { (0.1, 15.1222) (0.2, 15.0558) (0.3, 15.0477) (0.4, 15.0183) (0.5, 15.0264) (0.6, 15.0258) (0.7, 15.0112) (0.8, 15.0046) (0.9, 15.0046)  };
        \addplot[name path=U15, draw=none, forget plot] coordinates { (0.1, 16.0921) (0.2, 16.1867) (0.3, 16.1529) (0.4, 16.1529) (0.5, 16.1353) (0.6, 16.1261) (0.7, 16.1261) (0.8, 16.1261) (0.9, 16.1261)  };
        \addplot[name path=L15, draw=none, forget plot] coordinates { (0.1, 13.8311) (0.2, 13.9123) (0.3, 13.9006) (0.4, 13.9006) (0.5, 13.8887) (0.6, 13.8830) (0.7, 13.8830) (0.8, 13.8830) (0.9, 13.8830)  };
        \addplot[col5, opacity=0.15, forget plot] fill between[of=U15 and L15];
        \addplot[color=col5, mark=pentagon*, thick, mark options={solid}] coordinates { (0.1, 14.9616) (0.2, 15.0495) (0.3, 15.0267) (0.4, 15.0267) (0.5, 15.0120) (0.6, 15.0046) (0.7, 15.0046) (0.8, 15.0046) (0.9, 15.0046)  };
        \addplot[name path=U16, draw=none, forget plot] coordinates { (0.1, 15.8851) (0.2, 15.9877) (0.3, 16.0585) (0.4, 16.1261) (0.5, 16.1551) (0.6, 16.1261) (0.7, 16.1529) (0.8, 16.1353) (0.9, 16.1038)  };
        \addplot[name path=L16, draw=none, forget plot] coordinates { (0.1, 13.6734) (0.2, 13.7593) (0.3, 13.8193) (0.4, 13.8830) (0.5, 13.8852) (0.6, 13.8830) (0.7, 13.9006) (0.8, 13.8887) (0.9, 13.8616)  };
        \addplot[col6, opacity=0.15, forget plot] fill between[of=U16 and L16];
        \addplot[color=col6, mark=x, thick, mark options={solid}] coordinates { (0.1, 14.7793) (0.2, 14.8735) (0.3, 14.9389) (0.4, 15.0046) (0.5, 15.0201) (0.6, 15.0046) (0.7, 15.0267) (0.8, 15.0120) (0.9, 14.9827)  };
\nextgroupplot[title={MSE}, title style={yshift=-1.5ex}, ylabel={ }]
        \addplot[name path=U20, draw=none, forget plot] coordinates { (0.1, 0.0099) (0.2, 0.0070) (0.3, 0.0052) (0.4, 0.0039) (0.5, 0.0029) (0.6, 0.0021) (0.7, 0.0015) (0.8, 0.0011) (0.9, 0.0007)  };
        \addplot[name path=L20, draw=none, forget plot] coordinates { (0.1, 0.0095) (0.2, 0.0066) (0.3, 0.0049) (0.4, 0.0036) (0.5, 0.0027) (0.6, 0.0020) (0.7, 0.0014) (0.8, 0.0010) (0.9, 0.0007)  };
        \addplot[col0, opacity=0.15, forget plot] fill between[of=U20 and L20];
        \addplot[color=col0, mark=*, thick, mark options={solid}] coordinates { (0.1, 0.0097) (0.2, 0.0068) (0.3, 0.0050) (0.4, 0.0037) (0.5, 0.0028) (0.6, 0.0021) (0.7, 0.0015) (0.8, 0.0010) (0.9, 0.0007)  };
        \addplot[name path=U21, draw=none, forget plot] coordinates { (0.1, 0.0073) (0.2, 0.0046) (0.3, 0.0032) (0.4, 0.0022) (0.5, 0.0016) (0.6, 0.0012) (0.7, 0.0008) (0.8, 0.0006) (0.9, 0.0004)  };
        \addplot[name path=L21, draw=none, forget plot] coordinates { (0.1, 0.0066) (0.2, 0.0041) (0.3, 0.0028) (0.4, 0.0020) (0.5, 0.0014) (0.6, 0.0010) (0.7, 0.0007) (0.8, 0.0005) (0.9, 0.0003)  };
        \addplot[col1, opacity=0.15, forget plot] fill between[of=U21 and L21];
        \addplot[color=col1, mark=triangle*, thick, mark options={solid}] coordinates { (0.1, 0.0069) (0.2, 0.0044) (0.3, 0.0030) (0.4, 0.0021) (0.5, 0.0015) (0.6, 0.0011) (0.7, 0.0008) (0.8, 0.0005) (0.9, 0.0004)  };
        \addplot[name path=U22, draw=none, forget plot] coordinates { (0.1, 0.0111) (0.2, 0.0072) (0.3, 0.0051) (0.4, 0.0037) (0.5, 0.0026) (0.6, 0.0019) (0.7, 0.0013) (0.8, 0.0009) (0.9, 0.0006)  };
        \addplot[name path=L22, draw=none, forget plot] coordinates { (0.1, 0.0102) (0.2, 0.0066) (0.3, 0.0047) (0.4, 0.0033) (0.5, 0.0024) (0.6, 0.0017) (0.7, 0.0012) (0.8, 0.0008) (0.9, 0.0005)  };
        \addplot[col2, opacity=0.15, forget plot] fill between[of=U22 and L22];
        \addplot[color=col2, mark=triangle*, thick, mark options={solid}] coordinates { (0.1, 0.0106) (0.2, 0.0069) (0.3, 0.0049) (0.4, 0.0035) (0.5, 0.0025) (0.6, 0.0018) (0.7, 0.0012) (0.8, 0.0008) (0.9, 0.0005)  };
        \addplot[name path=U23, draw=none, forget plot] coordinates { (0.1, 0.0072) (0.2, 0.0050) (0.3, 0.0040) (0.4, 0.0031) (0.5, 0.0022) (0.6, 0.0014) (0.7, 0.0009) (0.8, 0.0005) (0.9, 0.0003)  };
        \addplot[name path=L23, draw=none, forget plot] coordinates { (0.1, 0.0069) (0.2, 0.0047) (0.3, 0.0039) (0.4, 0.0030) (0.5, 0.0021) (0.6, 0.0013) (0.7, 0.0009) (0.8, 0.0005) (0.9, 0.0003)  };
        \addplot[col3, opacity=0.15, forget plot] fill between[of=U23 and L23];
        \addplot[color=col3, mark=square*, thick, mark options={solid}] coordinates { (0.1, 0.0071) (0.2, 0.0048) (0.3, 0.0040) (0.4, 0.0030) (0.5, 0.0022) (0.6, 0.0013) (0.7, 0.0009) (0.8, 0.0005) (0.9, 0.0003)  };
        \addplot[name path=U24, draw=none, forget plot] coordinates { (0.1, 0.0047) (0.2, 0.0027) (0.3, 0.0017) (0.4, 0.0011) (0.5, 0.0008) (0.6, 0.0005) (0.7, 0.0004) (0.8, 0.0003) (0.9, 0.0002)  };
        \addplot[name path=L24, draw=none, forget plot] coordinates { (0.1, 0.0044) (0.2, 0.0024) (0.3, 0.0015) (0.4, 0.0010) (0.5, 0.0007) (0.6, 0.0005) (0.7, 0.0004) (0.8, 0.0003) (0.9, 0.0002)  };
        \addplot[col4, opacity=0.15, forget plot] fill between[of=U24 and L24];
        \addplot[color=col4, mark=diamond*, thick, mark options={solid}] coordinates { (0.1, 0.0046) (0.2, 0.0026) (0.3, 0.0016) (0.4, 0.0011) (0.5, 0.0007) (0.6, 0.0005) (0.7, 0.0004) (0.8, 0.0003) (0.9, 0.0002)  };
        \addplot[name path=U25, draw=none, forget plot] coordinates { (0.1, 0.0029) (0.2, 0.0012) (0.3, 0.0007) (0.4, 0.0004) (0.5, 0.0003) (0.6, 0.0002) (0.7, 0.0002) (0.8, 0.0001) (0.9, 0.0001)  };
        \addplot[name path=L25, draw=none, forget plot] coordinates { (0.1, 0.0027) (0.2, 0.0011) (0.3, 0.0006) (0.4, 0.0004) (0.5, 0.0003) (0.6, 0.0002) (0.7, 0.0001) (0.8, 0.0001) (0.9, 0.0001)  };
        \addplot[col5, opacity=0.15, forget plot] fill between[of=U25 and L25];
        \addplot[color=col5, mark=pentagon*, thick, mark options={solid}] coordinates { (0.1, 0.0028) (0.2, 0.0011) (0.3, 0.0006) (0.4, 0.0004) (0.5, 0.0003) (0.6, 0.0002) (0.7, 0.0002) (0.8, 0.0001) (0.9, 0.0001)  };
        \addplot[name path=U26, draw=none, forget plot] coordinates { (0.1, 0.0064) (0.2, 0.0036) (0.3, 0.0023) (0.4, 0.0016) (0.5, 0.0011) (0.6, 0.0008) (0.7, 0.0006) (0.8, 0.0005) (0.9, 0.0003)  };
        \addplot[name path=L26, draw=none, forget plot] coordinates { (0.1, 0.0058) (0.2, 0.0033) (0.3, 0.0021) (0.4, 0.0015) (0.5, 0.0011) (0.6, 0.0008) (0.7, 0.0006) (0.8, 0.0004) (0.9, 0.0003)  };
        \addplot[col6, opacity=0.15, forget plot] fill between[of=U26 and L26];
        \addplot[color=col6, mark=x, thick, mark options={solid}] coordinates { (0.1, 0.0061) (0.2, 0.0034) (0.3, 0.0022) (0.4, 0.0015) (0.5, 0.0011) (0.6, 0.0008) (0.7, 0.0006) (0.8, 0.0005) (0.9, 0.0003)  };
\nextgroupplot[title={KL-Divergence}, title style={yshift=-1.5ex}, ylabel={ }]
        \addplot[name path=U30, draw=none, forget plot] coordinates { (0.1, 0.0573) (0.2, 0.0399) (0.3, 0.0286) (0.4, 0.0207) (0.5, 0.0152) (0.6, 0.0109) (0.7, 0.0077) (0.8, 0.0053) (0.9, 0.0034)  };
        \addplot[name path=L30, draw=none, forget plot] coordinates { (0.1, 0.0539) (0.2, 0.0374) (0.3, 0.0272) (0.4, 0.0198) (0.5, 0.0143) (0.6, 0.0104) (0.7, 0.0073) (0.8, 0.0050) (0.9, 0.0032)  };
        \addplot[col0, opacity=0.15, forget plot] fill between[of=U30 and L30];
        \addplot[color=col0, mark=*, thick, mark options={solid}] coordinates { (0.1, 0.0556) (0.2, 0.0386) (0.3, 0.0279) (0.4, 0.0203) (0.5, 0.0148) (0.6, 0.0107) (0.7, 0.0075) (0.8, 0.0051) (0.9, 0.0033)  };
        \addplot[name path=U31, draw=none, forget plot] coordinates { (0.1, 0.0403) (0.2, 0.0246) (0.3, 0.0166) (0.4, 0.0117) (0.5, 0.0083) (0.6, 0.0060) (0.7, 0.0043) (0.8, 0.0029) (0.9, 0.0019)  };
        \addplot[name path=L31, draw=none, forget plot] coordinates { (0.1, 0.0367) (0.2, 0.0224) (0.3, 0.0148) (0.4, 0.0103) (0.5, 0.0073) (0.6, 0.0051) (0.7, 0.0035) (0.8, 0.0025) (0.9, 0.0016)  };
        \addplot[col1, opacity=0.15, forget plot] fill between[of=U31 and L31];
        \addplot[color=col1, mark=triangle*, thick, mark options={solid}] coordinates { (0.1, 0.0385) (0.2, 0.0235) (0.3, 0.0157) (0.4, 0.0110) (0.5, 0.0078) (0.6, 0.0055) (0.7, 0.0039) (0.8, 0.0027) (0.9, 0.0018)  };
        \addplot[name path=U32, draw=none, forget plot] coordinates { (0.1, 0.0627) (0.2, 0.0409) (0.3, 0.0282) (0.4, 0.0202) (0.5, 0.0143) (0.6, 0.0101) (0.7, 0.0070) (0.8, 0.0047) (0.9, 0.0029)  };
        \addplot[name path=L32, draw=none, forget plot] coordinates { (0.1, 0.0581) (0.2, 0.0380) (0.3, 0.0260) (0.4, 0.0183) (0.5, 0.0128) (0.6, 0.0089) (0.7, 0.0060) (0.8, 0.0040) (0.9, 0.0025)  };
        \addplot[col2, opacity=0.15, forget plot] fill between[of=U32 and L32];
        \addplot[color=col2, mark=triangle*, thick, mark options={solid}] coordinates { (0.1, 0.0604) (0.2, 0.0395) (0.3, 0.0271) (0.4, 0.0192) (0.5, 0.0136) (0.6, 0.0095) (0.7, 0.0065) (0.8, 0.0043) (0.9, 0.0027)  };
        \addplot[name path=U33, draw=none, forget plot] coordinates { (0.1, 0.0373) (0.2, 0.0258) (0.3, 0.0208) (0.4, 0.0157) (0.5, 0.0110) (0.6, 0.0067) (0.7, 0.0043) (0.8, 0.0024) (0.9, 0.0014)  };
        \addplot[name path=L33, draw=none, forget plot] coordinates { (0.1, 0.0353) (0.2, 0.0244) (0.3, 0.0199) (0.4, 0.0151) (0.5, 0.0106) (0.6, 0.0061) (0.7, 0.0039) (0.8, 0.0022) (0.9, 0.0012)  };
        \addplot[col3, opacity=0.15, forget plot] fill between[of=U33 and L33];
        \addplot[color=col3, mark=square*, thick, mark options={solid}] coordinates { (0.1, 0.0363) (0.2, 0.0251) (0.3, 0.0203) (0.4, 0.0154) (0.5, 0.0108) (0.6, 0.0064) (0.7, 0.0041) (0.8, 0.0023) (0.9, 0.0013)  };
        \addplot[name path=U34, draw=none, forget plot] coordinates { (0.1, 0.0275) (0.2, 0.0148) (0.3, 0.0092) (0.4, 0.0061) (0.5, 0.0041) (0.6, 0.0029) (0.7, 0.0020) (0.8, 0.0014) (0.9, 0.0010)  };
        \addplot[name path=L34, draw=none, forget plot] coordinates { (0.1, 0.0253) (0.2, 0.0135) (0.3, 0.0083) (0.4, 0.0053) (0.5, 0.0036) (0.6, 0.0025) (0.7, 0.0018) (0.8, 0.0012) (0.9, 0.0008)  };
        \addplot[col4, opacity=0.15, forget plot] fill between[of=U34 and L34];
        \addplot[color=col4, mark=diamond*, thick, mark options={solid}] coordinates { (0.1, 0.0264) (0.2, 0.0141) (0.3, 0.0087) (0.4, 0.0057) (0.5, 0.0039) (0.6, 0.0027) (0.7, 0.0019) (0.8, 0.0013) (0.9, 0.0009)  };
        \addplot[name path=U35, draw=none, forget plot] coordinates { (0.1, 0.0145) (0.2, 0.0054) (0.3, 0.0031) (0.4, 0.0019) (0.5, 0.0013) (0.6, 0.0009) (0.7, 0.0007) (0.8, 0.0005) (0.9, 0.0004)  };
        \addplot[name path=L35, draw=none, forget plot] coordinates { (0.1, 0.0131) (0.2, 0.0048) (0.3, 0.0026) (0.4, 0.0017) (0.5, 0.0012) (0.6, 0.0008) (0.7, 0.0006) (0.8, 0.0005) (0.9, 0.0003)  };
        \addplot[col5, opacity=0.15, forget plot] fill between[of=U35 and L35];
        \addplot[color=col5, mark=pentagon*, thick, mark options={solid}] coordinates { (0.1, 0.0138) (0.2, 0.0051) (0.3, 0.0028) (0.4, 0.0018) (0.5, 0.0012) (0.6, 0.0009) (0.7, 0.0007) (0.8, 0.0005) (0.9, 0.0004)  };
        \addplot[name path=U36, draw=none, forget plot] coordinates { (0.1, 0.0356) (0.2, 0.0202) (0.3, 0.0124) (0.4, 0.0084) (0.5, 0.0057) (0.6, 0.0041) (0.7, 0.0030) (0.8, 0.0023) (0.9, 0.0016)  };
        \addplot[name path=L36, draw=none, forget plot] coordinates { (0.1, 0.0311) (0.2, 0.0172) (0.3, 0.0107) (0.4, 0.0071) (0.5, 0.0049) (0.6, 0.0034) (0.7, 0.0025) (0.8, 0.0019) (0.9, 0.0014)  };
        \addplot[col6, opacity=0.15, forget plot] fill between[of=U36 and L36];
        \addplot[color=col6, mark=x, thick, mark options={solid}] coordinates { (0.1, 0.0333) (0.2, 0.0187) (0.3, 0.0115) (0.4, 0.0077) (0.5, 0.0053) (0.6, 0.0038) (0.7, 0.0028) (0.8, 0.0021) (0.9, 0.0015)  };
    \end{groupplot}

    \node[anchor=south] at ($(group c1r1.north)!0.5!(group c4r1.north) + (0, 0.8cm)$) {
        \begin{tikzpicture}
            \draw[col0, thick] (0.0, 0) -- (0.5, 0);
            \draw[col0, fill=col0] plot[mark=*, mark options={solid}, mark size=2pt] coordinates {(0.25, 0)};
            \node[anchor=west] at (0.5, 0) {\small down\_proj};
            \draw[col1, thick] (2.5, 0) -- (3.0, 0);
            \draw[col1, fill=col1] plot[mark=triangle*, mark options={solid}, mark size=2pt] coordinates {(2.75, 0)};
            \node[anchor=west] at (3.0, 0) {\small gate\_proj};
            \draw[col2, thick] (5.0, 0) -- (5.5, 0);
            \draw[col2, fill=col2] plot[mark=triangle*, mark options={solid}, mark size=2pt] coordinates {(5.25, 0)};
            \node[anchor=west] at (5.5, 0) {\small up\_proj};
            \draw[col3, thick] (7.5, 0) -- (8.0, 0);
            \draw[col3, fill=col3] plot[mark=square*, mark options={solid}, mark size=2pt] coordinates {(7.75, 0)};
            \node[anchor=west] at (8.0, 0) {\small k\_proj};
            \draw[col4, thick] (10.0, 0) -- (10.5, 0);
            \draw[col4, fill=col4] plot[mark=diamond*, mark options={solid}, mark size=2pt] coordinates {(10.25, 0)};
            \node[anchor=west] at (10.5, 0) {\small o\_proj};
            \draw[col5, thick] (12.5, 0) -- (13.0, 0);
            \draw[col5, fill=col5] plot[mark=pentagon*, mark options={solid}, mark size=2pt] coordinates {(12.75, 0)};
            \node[anchor=west] at (13.0, 0) {\small q\_proj};
            \draw[col6, thick] (15.0, 0) -- (15.5, 0);
            \draw[col6, fill=col6] plot[mark=x, mark options={solid}, mark size=2pt] coordinates {(15.25, 0)};
            \node[anchor=west] at (15.5, 0) {\small v\_proj};
        \end{tikzpicture}
    };
\end{tikzpicture}

%% file: data/plots/supp_search/relative_comp_layer_inportance_Llama32_3B_05.tex
\begin{tikzpicture}
\begin{axis}[
    ybar, ymin=0, ylabel=\scriptsize{Relative importance},
    xlabel=\scriptsize{Block number},
    xlabel shift        = {-0.1 cm},
    xtick=data,
    xticklabel style = { font=\scriptsize, yshift=+0.08cm
    },
    xtick               = { 1, 2, 3, 4, 5, 6, 7, 8, 9, 10, 11, 12, 13, 14, 15, 16, 17, 18, 19, 20, 21, 22, 23, 24, 25, 26, 27, 28},
    xticklabels         = { 1, , 3, , 5, , 7, , 9, , 11, , 13, , 15, , 17, , 19, , 21, , 23, , 25, , 27, ,},
    axis x line* = bottom,
    axis y line = left,
    height = 4.0cm,
    width = 1.1\textwidth,
    title = {\scriptsize\textbf{\gls{llama3}.2-3B} rel. importance by layer position (without bias)},
    ymin = 0.0, ymax = 3.50,
    xmin=0.0, xmax=29.0,
    bar width=2.5px,
    yticklabel style = { font=\scriptsize, },
    yticklabel shift = {-0.05cm},
    ylabel shift    = {-2.9 cm},
    ymajorgrids,
    ]
    \addplot[fill=icmlblue] coordinates {
        (1, 3.067933779209852)
        (2, 0.4695568382740021)
        (3, 0.47681015729904175)
        (4, 0.42493027448654175)
        (5, 0.5120867490768433)
        (6, 0.3666648864746094)
        (7, 0.27992093563079834)
        (8, 0.32997381687164307)
        (9, 0.21318089962005615)
        (10, 0.18350660800933838)
        (11, 0.1616734266281128)
        (12, 0.17722159624099731)
        (13, 0.20412993431091309)
        (14, 0.2286977767944336)
        (15, 0.31884562969207764)
        (16, 0.3417541980743408)
        (17, 0.3327906131744385)
        (18, 0.3006139397621155)
        (19, 0.30928635597229004)
        (20, 0.33457595109939575)
        (21, 0.27873826026916504)
        (22, 0.2698352038860321)
        (23, 0.262969970703125)
        (24, 0.2693336009979248)
        (25, 0.3252376914024353)
        (26, 0.3874925374984741)
        (27, 0.5620038211345673)
        (28, 0.7761836349964142)
    };

\end{axis}
\end{tikzpicture}

%% file: data/plots/supp_search/relative_comp_layer_type_importance_Llama32_3B_05.tex
\begin{tikzpicture}
\begin{axis}[
    ybar, ymin=0,%
    xlabel shift        = {-0.1 cm},
    xtick={1,2,3,4,5,6,7},
    legend style={
        at              = {(0.65, 1.0)}, %
        anchor          = north,
        font            = \scriptsize,
        legend columns  = 7
    },
    xticklabel style = { font=\scriptsize, yshift=+0.08cm
    },
    xticklabels={ q-proj, k-proj, v-proj, o-proj, gate-proj, up-proj, down-proj},
    xticklabel style = {anchor=east, align=right, font=\scriptsize, rotate=40, xshift=+0.15cm, yshift=-0.05cm},
    axis x line* = bottom,
    axis y line = left,
    height = 4.0cm,
    width = 1.2\linewidth,
    title = {\scriptsize\textbf{\gls{llama3}.2-3B} rel. importance by layer type},
    ymin = 0.0, ymax = 3.5,
    xmin=0, xmax=8,
    bar width=4px,
    ymajorgrids,
    yticklabels={},
    ]
    \addplot[fill=DarkYellow] coordinates {
        (1, 0.1701916567981243)
        (2, 0.624378889799118)
        (3, 2.4648730754852295)
        (4, 0.5164932012557983)
        (5, 2.753119468688965)
        (6, 2.777785301208496)
        (7, 2.859107494354248)
    };

\end{axis}
\end{tikzpicture}

%% file: data/plots/supp_search/relative_comp_layer_inportance_Qwen3_4B_05.tex
\begin{tikzpicture}
\begin{axis}[
    ybar, ymin=0, ylabel=\scriptsize{Relative importance},
    xlabel=\scriptsize{Block number},
    xlabel shift        = {-0.1 cm},
    xtick=data,
    xticklabel style = { font=\scriptsize, yshift=+0.08cm
    },
    xtick               = { 1, 2, 3, 4, 5, 6, 7, 8, 9, 10, 11, 12, 13, 14, 15, 16, 17, 18, 19, 20, 21, 22, 23, 24, 25, 26, 27, 28, 29, 30, 31, 32, 33, 34, 35, 36},
    xticklabels         = { 1, , 3, , 5, , 7, , 9, , 11, , 13, , 15, , 17, , 19, , 21, , 23, , 25, , 27, , 29, , 31, , 33, , 35, ,},
    axis x line* = bottom,
    axis y line = left,
    height = 4.0cm,
    width = 1.1\textwidth,
    title = {\scriptsize\textbf{\gls{qwen3}-4B} rel. importance by layer position (without bias)},
    ymin = 0.0, ymax = 4.00,
    xmin=0.0, xmax=37.0,
    bar width=2.5px,
    yticklabel style = { font=\scriptsize, },
    yticklabel shift = {-0.05cm},
    ylabel shift    = {-2.9 cm},
    ymajorgrids,
    ]
    \addplot[fill=icmlblue] coordinates {
        (1, 1.4875519275665283)
        (2, 0.28794166445732117)
        (3, 0.15951687097549438)
        (4, 0.35631364583969116)
        (5, 0.4268404245376587)
        (6, 0.4553786516189575)
        (7, 0.4971109628677368)
        (8, 0.6266237497329712)
        (9, 0.593454122543335)
        (10, 0.6184892654418945)
        (11, 0.6398448944091797)
        (12, 0.5112044811248779)
        (13, 0.4377620220184326)
        (14, 0.391582727432251)
        (15, 0.33112478256225586)
        (16, 0.2635002136230469)
        (17, 0.260381817817688)
        (18, 0.21851158142089844)
        (19, 0.22678375244140625)
        (20, 0.22945904731750488)
        (21, 0.20158016681671143)
        (22, 0.230338454246521)
        (23, 0.2888566255569458)
        (24, 0.3591582775115967)
        (25, 0.4173855781555176)
        (26, 0.3553929328918457)
        (27, 0.35500240325927734)
        (28, 0.3715341091156006)
        (29, 0.40773439407348633)
        (30, 0.4397796392440796)
        (31, 0.5031957626342773)
        (32, 0.45141059160232544)
        (33, 0.4550660252571106)
        (34, 0.4654039740562439)
        (35, 0.6438568234443665)
        (36, 0.6658291220664978)
    };

\end{axis}
\end{tikzpicture}

%% file: data/plots/supp_search/relative_comp_layer_type_importance_Qwen3_4B_05.tex
\begin{tikzpicture}
\begin{axis}[
    ybar, ymin=0,%
    xlabel shift        = {-0.1 cm},
    xtick={1,2,3,4,5,6,7},
    legend style={
        at              = {(0.65, 1.0)}, %
        anchor          = north,
        font            = \scriptsize,
        legend columns  = 7
    },
    xticklabel style = { font=\scriptsize, yshift=+0.08cm
    },
    xticklabels={ q-proj, k-proj, v-proj, o-proj, gate-proj, up-proj, down-proj},
    xticklabel style = {anchor=east, align=right, font=\scriptsize, rotate=40, xshift=+0.15cm, yshift=-0.05cm},
    axis x line* = bottom,
    axis y line = left,
    height = 4.0cm,
    width = 1.2\linewidth,
    title = {\scriptsize\textbf{\gls{qwen3}-4B} rel. importance by layer type},
    ymin = 0.0, ymax = 4.0,
    xmin=0, xmax=8,
    bar width=4px,
    ymajorgrids,
    yticklabels={},
    ]
    \addplot[fill=DarkYellow] coordinates {
        (1, 0.45611193776130676)
        (2, 2.031591773033142)
        (3, 1.7199697494506836)
        (4, 1.0155010223388672)
        (5, 3.1633505821228027)
        (6, 3.3750789165496826)
        (7, 3.869297504425049)
    };

\end{axis}
\end{tikzpicture}

%% file: data/plots/supp_search/compression_comparison_Llama32_3B_08_OURS.tex
\begin{tikzpicture}
\begin{axis}[
    ybar, ymin=0, ylabel=\scriptsize{Remaining Parameters [\%]},
    xlabel=\scriptsize{Layer number},
    xlabel shift        = {-0.1 cm},
    xtick=data,
    xticklabel style = { font=\scriptsize, yshift=+0.08cm
    },
    xticklabels         ={},
    axis x line* = bottom,
    axis y line = left,
    height = 4.0cm,
    width = 1.1\textwidth,
    title = {\scriptsize\textbf{\gls{llama3}.2-3B} maintained parameter percentage by layer position (\textbf{\gls{oursearch}})},
    ymin = 0.0, ymax = 1.10,
    xmin=0.0, xmax=197.0,
    bar width=1.0px,
    yticklabel style = { font=\scriptsize, },
    yticklabel shift = {-0.05cm},
    ylabel shift    = {-2.9 cm},
    ymajorgrids,
    ]
    \addplot[fill=icmlblue, draw=none,] coordinates {
        (1, 0.302734375)
        (2, 0.3489583333333333)
        (3, 1.0)
        (4, 1.0)
        (5, 1.0)
        (6, 1.0)
        (7, 1.0)
        (8, 0.302734375)
        (9, 0.9322916666666666)
        (10, 1.0)
        (11, 1.0)
        (12, 1.0)
        (13, 1.0)
        (14, 1.0)
        (15, 0.4694010416666667)
        (16, 1.0)
        (17, 1.0)
        (18, 0.6673177083333334)
        (19, 1.0)
        (20, 1.0)
        (21, 1.0)
        (22, 0.4641927083333333)
        (23, 1.0)
        (24, 1.0)
        (25, 1.0)
        (26, 1.0)
        (27, 1.0)
        (28, 1.0)
        (29, 0.583984375)
        (30, 1.0)
        (31, 1.0)
        (32, 1.0)
        (33, 1.0)
        (34, 1.0)
        (35, 1.0)
        (36, 0.396484375)
        (37, 1.0)
        (38, 1.0)
        (39, 0.583984375)
        (40, 1.0)
        (41, 1.0)
        (42, 1.0)
        (43, 0.318359375)
        (44, 1.0)
        (45, 1.0)
        (46, 0.6412760416666666)
        (47, 1.0)
        (48, 1.0)
        (49, 1.0)
        (50, 0.3704427083333333)
        (51, 1.0)
        (52, 1.0)
        (53, 0.6569010416666666)
        (54, 1.0)
        (55, 1.0)
        (56, 1.0)
        (57, 0.3548177083333333)
        (58, 0.7864583333333334)
        (59, 1.0)
        (60, 0.5891927083333334)
        (61, 0.6011638316920322)
        (62, 1.0)
        (63, 0.7085944494180841)
        (64, 0.3079427083333333)
        (65, 0.8802083333333334)
        (66, 1.0)
        (67, 0.6256510416666666)
        (68, 0.4722470904207699)
        (69, 0.6871083258728737)
        (70, 0.6226499552372426)
        (71, 0.3287760416666667)
        (72, 0.703125)
        (73, 1.0)
        (74, 0.5579427083333334)
        (75, 0.39346463742166515)
        (76, 0.5975828111011638)
        (77, 0.554610564010743)
        (78, 0.302734375)
        (79, 0.7135416666666666)
        (80, 1.0)
        (81, 0.6048177083333334)
        (82, 0.40062667860340195)
        (83, 0.6226499552372426)
        (84, 0.5832587287376902)
        (85, 0.3600260416666667)
        (86, 0.703125)
        (87, 1.0)
        (88, 0.6256510416666666)
        (89, 0.38630259623992835)
        (90, 0.6763652641002685)
        (91, 0.6512981199641898)
        (92, 0.302734375)
        (93, 0.6614583333333334)
        (94, 1.0)
        (95, 0.6360677083333334)
        (96, 0.4865711727842435)
        (97, 1.0)
        (98, 1.0)
        (99, 0.302734375)
        (100, 0.7135416666666666)
        (101, 1.0)
        (102, 1.0)
        (103, 0.5080572963294538)
        (104, 1.0)
        (105, 1.0)
        (106, 0.302734375)
        (107, 0.6614583333333334)
        (108, 1.0)
        (109, 0.6100260416666666)
        (110, 0.644136078782453)
        (111, 1.0)
        (112, 1.0)
        (113, 0.3287760416666667)
        (114, 1.0)
        (115, 1.0)
        (116, 1.0)
        (117, 0.6692032229185317)
        (118, 1.0)
        (119, 1.0)
        (120, 0.3287760416666667)
        (121, 0.6614583333333334)
        (122, 1.0)
        (123, 0.412109375)
        (124, 1.0)
        (125, 1.0)
        (126, 1.0)
        (127, 0.3131510416666667)
        (128, 0.453125)
        (129, 1.0)
        (130, 0.3444010416666667)
        (131, 1.0)
        (132, 1.0)
        (133, 1.0)
        (134, 0.3079427083333333)
        (135, 0.6614583333333334)
        (136, 1.0)
        (137, 0.3235677083333333)
        (138, 1.0)
        (139, 1.0)
        (140, 1.0)
        (141, 0.3131510416666667)
        (142, 0.5677083333333334)
        (143, 1.0)
        (144, 0.3131510416666667)
        (145, 0.6656222023276633)
        (146, 0.6477170993733214)
        (147, 1.0)
        (148, 0.3079427083333333)
        (149, 0.5885416666666666)
        (150, 0.7135416666666666)
        (151, 0.3548177083333333)
        (152, 0.6477170993733214)
        (153, 0.6369740376007162)
        (154, 1.0)
        (155, 0.3131510416666667)
        (156, 0.578125)
        (157, 1.0)
        (158, 0.380859375)
        (159, 0.6620411817367949)
        (160, 0.5940017905102954)
        (161, 1.0)
        (162, 0.3131510416666667)
        (163, 0.4635416666666667)
        (164, 1.0)
        (165, 0.302734375)
        (166, 1.0)
        (167, 0.6369740376007162)
        (168, 1.0)
        (169, 0.302734375)
        (170, 0.5572916666666666)
        (171, 1.0)
        (172, 0.318359375)
        (173, 1.0)
        (174, 1.0)
        (175, 1.0)
        (176, 0.318359375)
        (177, 0.6822916666666666)
        (178, 1.0)
        (179, 0.302734375)
        (180, 1.0)
        (181, 1.0)
        (182, 1.0)
        (183, 0.302734375)
        (184, 0.4947916666666667)
        (185, 1.0)
        (186, 0.412109375)
        (187, 1.0)
        (188, 1.0)
        (189, 1.0)
        (190, 0.302734375)
        (191, 0.5260416666666666)
        (192, 1.0)
        (193, 0.443359375)
        (194, 1.0)
        (195, 1.0)
        (196, 1.0)
    };

    \addplot[red, very thick, const plot] coordinates {
        (1, 0.8073846726190476)
        (8, 0.8907180059523808)
        (15, 0.8766741071428571)
        (22, 0.9234561011904762)
        (29, 0.9405691964285714)
        (36, 0.8543526785714286)
        (43, 0.851376488095238)
        (50, 0.8610491071428571)
        (57, 0.7200324330157308)
        (64, 0.65654392212346)
        (71, 0.5907859660762247)
        (78, 0.6039470160826192)
        (85, 0.62896686623396)
        (92, 0.7266902270644158)
        (99, 0.7891904768565886)
        (106, 0.7454792612546361)
        (113, 0.8568541806550284)
        (120, 0.7717633928571429)
        (127, 0.7300967261904763)
        (134, 0.7561383928571429)
        (141, 0.6439071026239501)
        (148, 0.6070764124248625)
        (155, 0.6468826269876795)
        (162, 0.6737715887048642)
        (169, 0.7397693452380951)
        (176, 0.757626488095238)
        (183, 0.744233630952381)
        (190, 0.7531622023809523)
        (197, 0.7531622023809523)
};
\end{axis}
\end{tikzpicture}

%% file: data/plots/supp_search/compression_comparison_layer_type_Llama32_3B_08_OURS.tex
\begin{tikzpicture}
\begin{axis}[
    ybar, ymin=0,%
    xlabel shift        = {-0.1 cm},
    xtick={1,2,3,4,5,6,7},
    legend style={
        at              = {(0.65, 1.0)}, %
        anchor          = north,
        font            = \scriptsize,
        legend columns  = 7
    },
    xticklabel style = { font=\scriptsize, yshift=+0.08cm
    },
    xticklabels={ q-proj, k-proj, v-proj, o-proj, gate-proj, up-proj, down-proj},
    xticklabel style = {anchor=east, align=right, font=\scriptsize, rotate=40, xshift=+0.15cm, yshift=-0.05cm},
    axis x line* = bottom,
    axis y line = left,
    height = 4.0cm,
    width = 1.2\linewidth,
    title = {\scriptsize avrg. compression by layer type},
    ymin = 0.0, ymax = 1.1,
    xmin=0, xmax=8,
    bar width=4px,
    ymajorgrids,
    yticklabels={},
    ]
    \addplot[fill=DarkYellow] coordinates {
        (1, 0.3283943965517242)
        (2, 0.7013290229885061)
        (3, 0.955639367816092)
        (4, 0.576104525862069)
        (5, 0.7771432099527676)
        (6, 0.8654956317722965)
        (7, 0.9007038557713087)
    };

\end{axis}
\end{tikzpicture}

%% file: data/plots/supp_search/compression_comparison_Llama32_3B_08_ATP.tex
\begin{tikzpicture}
\begin{axis}[
    ybar, ymin=0, ylabel=\scriptsize{Remaining Parameters [\%]},
    xlabel=\scriptsize{Layer number},
    xlabel shift        = {-0.1 cm},
    xtick=data,
    xticklabel style = { font=\scriptsize, yshift=+0.08cm
    },
    xticklabels         ={},
    axis x line* = bottom,
    axis y line = left,
    height = 4.0cm,
    width = 1.1\textwidth,
    title = {\scriptsize\textbf{\gls{llama3}.2-3B} maintained parameter percentage by layer position (\textbf{\gls{atp}})},
    ymin = 0.0, ymax = 1.10,
    xmin=0.0, xmax=197.0,
    bar width=1.0px,
    yticklabel style = { font=\scriptsize, },
    yticklabel shift = {-0.05cm},
    ylabel shift    = {-2.9 cm},
    ymajorgrids,
    ]
    \addplot[fill=icmlblue, draw=none,] coordinates {
        (1, 0.9928571428571429)
        (2, 0.9928571428571429)
        (3, 0.9928571428571429)
        (4, 0.9928571428571429)
        (5, 0.9928571428571429)
        (6, 0.9928571428571429)
        (7, 0.9928571428571429)
        (8, 0.9785714285714286)
        (9, 0.9785714285714286)
        (10, 0.9785714285714286)
        (11, 0.9785714285714286)
        (12, 0.9785714285714286)
        (13, 0.9785714285714286)
        (14, 0.9785714285714286)
        (15, 0.9642857142857144)
        (16, 0.9642857142857144)
        (17, 0.9642857142857144)
        (18, 0.9642857142857144)
        (19, 0.9642857142857144)
        (20, 0.9642857142857144)
        (21, 0.9642857142857144)
        (22, 0.9500000000000001)
        (23, 0.9500000000000001)
        (24, 0.9500000000000001)
        (25, 0.9500000000000001)
        (26, 0.9500000000000001)
        (27, 0.9500000000000001)
        (28, 0.9500000000000001)
        (29, 0.9357142857142857)
        (30, 0.9357142857142857)
        (31, 0.9357142857142857)
        (32, 0.9357142857142857)
        (33, 0.9357142857142857)
        (34, 0.9357142857142857)
        (35, 0.9357142857142857)
        (36, 0.9214285714285715)
        (37, 0.9214285714285715)
        (38, 0.9214285714285715)
        (39, 0.9214285714285715)
        (40, 0.9214285714285715)
        (41, 0.9214285714285715)
        (42, 0.9214285714285715)
        (43, 0.9071428571428573)
        (44, 0.9071428571428573)
        (45, 0.9071428571428573)
        (46, 0.9071428571428573)
        (47, 0.9071428571428573)
        (48, 0.9071428571428573)
        (49, 0.9071428571428573)
        (50, 0.8928571428571429)
        (51, 0.8928571428571429)
        (52, 0.8928571428571429)
        (53, 0.8928571428571429)
        (54, 0.8928571428571429)
        (55, 0.8928571428571429)
        (56, 0.8928571428571429)
        (57, 0.8785714285714287)
        (58, 0.8785714285714287)
        (59, 0.8785714285714287)
        (60, 0.8785714285714287)
        (61, 0.8785714285714287)
        (62, 0.8785714285714287)
        (63, 0.8785714285714287)
        (64, 0.8642857142857143)
        (65, 0.8642857142857143)
        (66, 0.8642857142857143)
        (67, 0.8642857142857143)
        (68, 0.8642857142857143)
        (69, 0.8642857142857143)
        (70, 0.8642857142857143)
        (71, 0.8500000000000001)
        (72, 0.8500000000000001)
        (73, 0.8500000000000001)
        (74, 0.8500000000000001)
        (75, 0.8500000000000001)
        (76, 0.8500000000000001)
        (77, 0.8500000000000001)
        (78, 0.8357142857142857)
        (79, 0.8357142857142857)
        (80, 0.8357142857142857)
        (81, 0.8357142857142857)
        (82, 0.8357142857142857)
        (83, 0.8357142857142857)
        (84, 0.8357142857142857)
        (85, 0.8214285714285715)
        (86, 0.8214285714285715)
        (87, 0.8214285714285715)
        (88, 0.8214285714285715)
        (89, 0.8214285714285715)
        (90, 0.8214285714285715)
        (91, 0.8214285714285715)
        (92, 0.8071428571428572)
        (93, 0.8071428571428572)
        (94, 0.8071428571428572)
        (95, 0.8071428571428572)
        (96, 0.8071428571428572)
        (97, 0.8071428571428572)
        (98, 0.8071428571428572)
        (99, 0.7928571428571429)
        (100, 0.7928571428571429)
        (101, 0.7928571428571429)
        (102, 0.7928571428571429)
        (103, 0.7928571428571429)
        (104, 0.7928571428571429)
        (105, 0.7928571428571429)
        (106, 0.7785714285714286)
        (107, 0.7785714285714286)
        (108, 0.7785714285714286)
        (109, 0.7785714285714286)
        (110, 0.7785714285714286)
        (111, 0.7785714285714286)
        (112, 0.7785714285714286)
        (113, 0.7642857142857143)
        (114, 0.7642857142857143)
        (115, 0.7642857142857143)
        (116, 0.7642857142857143)
        (117, 0.7642857142857143)
        (118, 0.7642857142857143)
        (119, 0.7642857142857143)
        (120, 0.75)
        (121, 0.75)
        (122, 0.75)
        (123, 0.75)
        (124, 0.75)
        (125, 0.75)
        (126, 0.75)
        (127, 0.7357142857142858)
        (128, 0.7357142857142858)
        (129, 0.7357142857142858)
        (130, 0.7357142857142858)
        (131, 0.7357142857142858)
        (132, 0.7357142857142858)
        (133, 0.7357142857142858)
        (134, 0.7214285714285714)
        (135, 0.7214285714285714)
        (136, 0.7214285714285714)
        (137, 0.7214285714285714)
        (138, 0.7214285714285714)
        (139, 0.7214285714285714)
        (140, 0.7214285714285714)
        (141, 0.7071428571428572)
        (142, 0.7071428571428572)
        (143, 0.7071428571428572)
        (144, 0.7071428571428572)
        (145, 0.7071428571428572)
        (146, 0.7071428571428572)
        (147, 0.7071428571428572)
        (148, 0.6928571428571428)
        (149, 0.6928571428571428)
        (150, 0.6928571428571428)
        (151, 0.6928571428571428)
        (152, 0.6928571428571428)
        (153, 0.6928571428571428)
        (154, 0.6928571428571428)
        (155, 0.6785714285714286)
        (156, 0.6785714285714286)
        (157, 0.6785714285714286)
        (158, 0.6785714285714286)
        (159, 0.6785714285714286)
        (160, 0.6785714285714286)
        (161, 0.6785714285714286)
        (162, 0.6642857142857144)
        (163, 0.6642857142857144)
        (164, 0.6642857142857144)
        (165, 0.6642857142857144)
        (166, 0.6642857142857144)
        (167, 0.6642857142857144)
        (168, 0.6642857142857144)
        (169, 0.65)
        (170, 0.65)
        (171, 0.65)
        (172, 0.65)
        (173, 0.65)
        (174, 0.65)
        (175, 0.65)
        (176, 0.6357142857142858)
        (177, 0.6357142857142858)
        (178, 0.6357142857142858)
        (179, 0.6357142857142858)
        (180, 0.6357142857142858)
        (181, 0.6357142857142858)
        (182, 0.6357142857142858)
        (183, 0.6214285714285714)
        (184, 0.6214285714285714)
        (185, 0.6214285714285714)
        (186, 0.6214285714285714)
        (187, 0.6214285714285714)
        (188, 0.6214285714285714)
        (189, 0.6214285714285714)
        (190, 0.6071428571428572)
        (191, 0.6071428571428572)
        (192, 0.6071428571428572)
        (193, 0.6071428571428572)
        (194, 0.6071428571428572)
        (195, 0.6071428571428572)
        (196, 0.6071428571428572)
    };

    \addplot[red, very thick, const plot] coordinates {
        (1, 0.9928571428571429)
        (8, 0.9785714285714286)
        (15, 0.9642857142857144)
        (22, 0.9500000000000001)
        (29, 0.9357142857142858)
        (36, 0.9214285714285716)
        (43, 0.9071428571428574)
        (50, 0.892857142857143)
        (57, 0.8785714285714288)
        (64, 0.8642857142857141)
        (71, 0.8499999999999999)
        (78, 0.8357142857142856)
        (85, 0.8214285714285714)
        (92, 0.807142857142857)
        (99, 0.7928571428571428)
        (106, 0.7785714285714286)
        (113, 0.7642857142857143)
        (120, 0.75)
        (127, 0.7357142857142858)
        (134, 0.7214285714285714)
        (141, 0.7071428571428572)
        (148, 0.692857142857143)
        (155, 0.6785714285714287)
        (162, 0.6642857142857145)
        (169, 0.65)
        (176, 0.6357142857142858)
        (183, 0.6214285714285713)
        (190, 0.6071428571428571)
        (196, 0.6071428571428571)
};
\end{axis}
\end{tikzpicture}

%% file: data/plots/supp_search/compression_comparison_layer_type_Llama32_3B_08_ATP.tex
\begin{tikzpicture}
\begin{axis}[
    ybar, ymin=0,%
    xlabel shift        = {-0.1 cm},
    xtick={1,2,3,4,5,6,7},
    legend style={
        at              = {(0.65, 1.0)}, %
        anchor          = north,
        font            = \scriptsize,
        legend columns  = 7
    },
    xticklabel style = { font=\scriptsize, yshift=+0.08cm
    },
    xticklabels={ q-proj, k-proj, v-proj, o-proj, gate-proj, up-proj, down-proj},
    xticklabel style = {anchor=east, align=right, font=\scriptsize, rotate=40, xshift=+0.15cm, yshift=-0.05cm},
    axis x line* = bottom,
    axis y line = left,
    height = 4.0cm,
    width = 1.2\linewidth,
    title = {\scriptsize avrg. compression by layer type},
    ymin = 0.0, ymax = 1.1,
    xmin=0, xmax=8,
    bar width=4px,
    ymajorgrids,
    yticklabels={},
    ]
    \addplot[fill=DarkYellow] coordinates {
        (1, 0.7724137931034482)
        (2, 0.7724137931034482)
        (3, 0.7724137931034482)
        (4, 0.7724137931034482)
        (5, 0.7724137931034482)
        (6, 0.7724137931034482)
        (7, 0.7724137931034482)
    };

\end{axis}
\end{tikzpicture}

%% file: data/plots/supp_search/compression_comparison_Llama32_3B_08_ASVD.tex
\begin{tikzpicture}
\begin{axis}[
    ybar, ymin=0, ylabel=\scriptsize{Remaining Parameters [\%]},
    xlabel=\scriptsize{Layer number},
    xlabel shift        = {-0.1 cm},
    xtick=data,
    xticklabel style = { font=\scriptsize, yshift=+0.08cm
    },
    xticklabels         ={},
    axis x line* = bottom,
    axis y line = left,
    height = 4.0cm,
    width = 1.1\textwidth,
    title = {\scriptsize\textbf{\gls{llama3}.2-3B} maintained parameter percentage by layer position (\textbf{\gls{asvd}})},
    ymin = 0.0, ymax = 1.10,
    xmin=0.0, xmax=197.0,
    bar width=1.0px,
    yticklabel style = { font=\scriptsize, },
    yticklabel shift = {-0.05cm},
    ylabel shift    = {-2.9 cm},
    ymajorgrids,
    ]
    \addplot[fill=icmlblue, draw=none,] coordinates {
        (1, 0.302734375)
        (2, 0.3072916666666667)
        (3, 1.0)
        (4, 0.6204427083333334)
        (5, 1.0)
        (6, 1.0)
        (7, 1.0)
        (8, 0.302734375)
        (9, 0.6614583333333334)
        (10, 0.9427083333333334)
        (11, 0.9225260416666666)
        (12, 1.0)
        (13, 1.0)
        (14, 1.0)
        (15, 0.3756510416666667)
        (16, 0.6614583333333334)
        (17, 0.734375)
        (18, 0.5475260416666666)
        (19, 1.0)
        (20, 1.0)
        (21, 1.0)
        (22, 0.3600260416666667)
        (23, 0.8385416666666666)
        (24, 1.0)
        (25, 0.6881510416666666)
        (26, 1.0)
        (27, 1.0)
        (28, 1.0)
        (29, 0.4798177083333333)
        (30, 0.8697916666666666)
        (31, 0.8697916666666666)
        (32, 0.5891927083333334)
        (33, 1.0)
        (34, 1.0)
        (35, 1.0)
        (36, 0.302734375)
        (37, 0.7447916666666666)
        (38, 0.6614583333333334)
        (39, 0.4954427083333333)
        (40, 1.0)
        (41, 1.0)
        (42, 1.0)
        (43, 0.302734375)
        (44, 0.6302083333333334)
        (45, 0.7135416666666666)
        (46, 0.5735677083333334)
        (47, 0.94136078782453)
        (48, 1.0)
        (49, 1.0)
        (50, 0.302734375)
        (51, 0.609375)
        (52, 0.671875)
        (53, 0.5735677083333334)
        (54, 1.0)
        (55, 1.0)
        (56, 0.9019695613249776)
        (57, 0.302734375)
        (58, 0.3489583333333333)
        (59, 0.640625)
        (60, 0.5423177083333334)
        (61, 0.8375111906893464)
        (62, 1.0)
        (63, 0.8733213965980304)
        (64, 0.302734375)
        (65, 0.4010416666666667)
        (66, 0.578125)
        (67, 0.5266927083333334)
        (68, 0.7229185317815577)
        (69, 0.8769024171888988)
        (70, 0.8231871083258728)
        (71, 0.302734375)
        (72, 0.3072916666666667)
        (73, 0.453125)
        (74, 0.4485677083333333)
        (75, 0.6763652641002685)
        (76, 0.8303491495076096)
        (77, 0.7945389435989257)
        (78, 0.302734375)
        (79, 0.3072916666666667)
        (80, 0.5677083333333334)
        (81, 0.5319010416666666)
        (82, 0.6906893464637421)
        (83, 0.8375111906893464)
        (84, 0.8088630259623992)
        (85, 0.302734375)
        (86, 0.3072916666666667)
        (87, 0.5677083333333334)
        (88, 0.5266927083333334)
        (89, 0.6799462846911369)
        (90, 0.8661593554162936)
        (91, 0.8410922112802148)
        (92, 0.302734375)
        (93, 0.3072916666666667)
        (94, 0.5885416666666666)
        (95, 0.5162760416666666)
        (96, 0.7551477170993733)
        (97, 0.8840644583706356)
        (98, 0.8948075201432408)
        (99, 0.302734375)
        (100, 0.3072916666666667)
        (101, 0.765625)
        (102, 0.6256510416666666)
        (103, 0.7766338406445837)
        (104, 0.9377797672336616)
        (105, 1.0)
        (106, 0.302734375)
        (107, 0.3072916666666667)
        (108, 0.7447916666666666)
        (109, 0.505859375)
        (110, 0.8482542524619516)
        (111, 0.9306177260519248)
        (112, 1.0)
        (113, 0.302734375)
        (114, 0.359375)
        (115, 0.546875)
        (116, 0.5944010416666666)
        (117, 0.8769024171888988)
        (118, 0.9198746642793196)
        (119, 1.0)
        (120, 0.302734375)
        (121, 0.3072916666666667)
        (122, 0.4010416666666667)
        (123, 0.3391927083333333)
        (124, 0.8983885407341092)
        (125, 0.905550581915846)
        (126, 1.0)
        (127, 0.302734375)
        (128, 0.3072916666666667)
        (129, 0.359375)
        (130, 0.302734375)
        (131, 0.9162936436884512)
        (132, 0.9306177260519248)
        (133, 1.0)
        (134, 0.302734375)
        (135, 0.3072916666666667)
        (136, 0.4635416666666667)
        (137, 0.302734375)
        (138, 0.9270367054610564)
        (139, 0.923455684870188)
        (140, 1.0)
        (141, 0.302734375)
        (142, 0.3072916666666667)
        (143, 0.3177083333333333)
        (144, 0.302734375)
        (145, 0.8804834377797672)
        (146, 0.8661593554162936)
        (147, 1.0)
        (148, 0.302734375)
        (149, 0.3072916666666667)
        (150, 0.3072916666666667)
        (151, 0.302734375)
        (152, 0.8804834377797672)
        (153, 0.8554162936436884)
        (154, 0.9449418084153984)
        (155, 0.302734375)
        (156, 0.3072916666666667)
        (157, 0.3802083333333333)
        (158, 0.302734375)
        (159, 0.8769024171888988)
        (160, 0.8446732318710832)
        (161, 0.905550581915846)
        (162, 0.302734375)
        (163, 0.3072916666666667)
        (164, 0.390625)
        (165, 0.302734375)
        (166, 0.9019695613249776)
        (167, 0.8625783348254252)
        (168, 0.8840644583706356)
        (169, 0.302734375)
        (170, 0.3072916666666667)
        (171, 0.3072916666666667)
        (172, 0.302734375)
        (173, 1.0)
        (174, 0.9091316025067144)
        (175, 0.8983885407341092)
        (176, 0.302734375)
        (177, 0.3072916666666667)
        (178, 0.4114583333333333)
        (179, 0.302734375)
        (180, 1.0)
        (181, 1.0)
        (182, 1.0)
        (183, 0.302734375)
        (184, 0.3072916666666667)
        (185, 0.4739583333333333)
        (186, 0.302734375)
        (187, 1.0)
        (188, 1.0)
        (189, 1.0)
        (190, 0.302734375)
        (191, 0.3072916666666667)
        (192, 0.515625)
        (193, 0.3391927083333333)
        (194, 1.0)
        (195, 1.0)
        (196, 1.0)
    };

    \addplot[red, very thick, const plot] coordinates {
        (1, 0.7472098214285714)
        (8, 0.8327752976190477)
        (15, 0.7598586309523808)
        (22, 0.8409598214285714)
        (29, 0.8297991071428571)
        (36, 0.7434895833333334)
        (43, 0.7373446958796948)
        (50, 0.7227888063797588)
        (57, 0.6493525719934349)
        (64, 0.6045145438994757)
        (71, 0.5447103010295434)
        (78, 0.5780998542545934)
        (85, 0.5845178478172828)
        (92, 0.6069804922304642)
        (99, 0.6736736701730827)
        (106, 0.66279272312103)
        (113, 0.6571660711621264)
        (120, 0.5934570770452317)
        (127, 0.588435255201006)
        (134, 0.6038277819520825)
        (141, 0.5681587918851515)
        (148, 0.5572705175960267)
        (155, 0.5600135687108326)
        (162, 0.5645711101696722)
        (169, 0.5753674609391654)
        (176, 0.6177455357142857)
        (183, 0.6266741071428571)
        (190, 0.6378348214285714)
        (197, 0.6378348214285714)
};
\end{axis}
\end{tikzpicture}

%% file: data/plots/supp_search/compression_comparison_layer_type_Llama32_3B_08_ASVD.tex
\begin{tikzpicture}
\begin{axis}[
    ybar, ymin=0,%
    xlabel shift        = {-0.1 cm},
    xtick={1,2,3,4,5,6,7},
    legend style={
        at              = {(0.65, 1.0)}, %
        anchor          = north,
        font            = \scriptsize,
        legend columns  = 7
    },
    xticklabel style = { font=\scriptsize, yshift=+0.08cm
    },
    xticklabels={ q-proj, k-proj, v-proj, o-proj, gate-proj, up-proj, down-proj},
    xticklabel style = {anchor=east, align=right, font=\scriptsize, rotate=40, xshift=+0.15cm, yshift=-0.05cm},
    axis x line* = bottom,
    axis y line = left,
    height = 4.0cm,
    width = 1.2\linewidth,
    title = {\scriptsize avrg. compression by layer type},
    ymin = 0.0, ymax = 1.1,
    xmin=0, xmax=8,
    bar width=4px,
    ymajorgrids,
    yticklabels={},
    ]
    \addplot[fill=DarkYellow] coordinates {
        (1, 0.3028915229885058)
        (2, 0.4019396551724136)
        (3, 0.5646551724137931)
        (4, 0.4562679597701149)
        (5, 0.8650788750656004)
        (6, 0.9027876393047881)
        (7, 0.916231901954126)
    };

\end{axis}
\end{tikzpicture}

%% file: content/supplementary_sections/fisher_section.tex
\paragraph{Experimental Protocol and Metrics.}
For the extended evaluation of \gls{ours}, we ensure an identical data split and evaluation procedure so that everything but the factor estimation is the same. To collect token-wise activations $X\in\mathbb{R}^{B\times T\times d_{\text{in}}}$ and output gradients $\Delta\in\mathbb{R}^{B\times T\times d_{\text{out}}}$, we run forward and backward passes on our calibration data. From those, we estimate the Kronecker factors, $\widehat{\mathbf{A}}$ and $\widehat{\mathbf{G}}$, apply minimal diagonal damping to ensure Cholesky stability, and then compute the Fisher-weighted \gls{svd} truncation $\widehat{\mathbf{W}}_r$ (\cref{eq:weighted_frob}-\eqref{eq:unwhiten}).
Additionally, we obtain gradients and activations from an additional, held-out test set, that is not used during factor estimation.

To assess the estimated factors, we use two metrics. First, we measure \emph{held-out Fisher degradation} $\Delta F$ when compressing a layer to rank $r=128$. For that purpose, we use the unseen test set to compute the empirical Fisher quadratic form
\begin{equation*}
    \Delta F \;=\; \mathbb{E}\big[\langle \delta, (\mathbf{W}-\widehat{\mathbf{W}}_{128})x\rangle^2\big],
\end{equation*}
which is measuring sequence-wise (sum over tokens per sequence, then square) Fisher deltas induced by the compression. Therefore, a lower $\Delta F$ indicates better generalization to unseen curvature directions, representing a more faithful Fisher-weighted truncation.
Second, we measure the the \emph{effective rank} $r_{\mathrm{eff}}$ of the Kronecker spectrum $\widehat{\mathbf{A}}\otimes\widehat{\mathbf{G}}$, which is computed from the trace-normalized eigenvalues $p_i$ as $r_{\mathrm{eff}}=\exp(-\sum_i p_i\log p_i)$. Higher $r_{\mathrm{eff}}$ indicates more diverse directions covered.

\paragraph{Extended Comparison between KFAC and Lanczos (GFWSVD).}
In the main content, we illustrate the failure mode of the sequence-gradient estimator of \glslink{gfwsvd}{GFWSVD}\glsunset{gfwsvd} (\cref{sec:fisher_ablation}). Using the protocol outlined above, we provide broader evidence of the phenomenon across layers and models. Specifically, \cref{tab:lanczos_results} reports (i) held-out Fisher degradation $\Delta F$ after rank truncation and (ii) the effective rank $r_{\mathrm{eff}}$ of the implied Kronecker spectrum on \gls{llama3}.2-3B and \gls{qwen3}-4B on block 0 and block 15.
Across layers and blocks, Lanczos/NKP (as used in \gls{gfwsvd}) frequently exhibits \emph{spectral collapse}: the factors produce substantially lower $r_{\mathrm{eff}}$ than the token-wise \glslink{kfac}{KFAC}\glsunset{kfac}, which we attribute to the coupled estimator's reliance on a small number of sequence-level gradient samples, which concentrates curvature mass into a low-dimensional subspace.
When examining the truncation quality, we find that the spectral collapse correlates with worse truncation. At rank~128, \gls{kfac} consistently achieves lower held-out Fisher degradation than \gls{gfwsvd} (e.g., $\approx 1.4\times$ to $3\times$ lower $\Delta F$ depending on model/layer group in \cref{tab:lanczos_results}).

Combined, the results support our claim that the dominant limitation of coupled sequence-gradient fitting is the sample-limited rank bottleneck. In contrast the token-wise estimated factors of \gls{kfac} retain substantially more diverse curvature directions and generalize better to unseen gradients.

\begin{table}[]
    \centering
    \caption{Empirical Comparison of the actual Fisher information lost, when compressing with the obtained factors to rank 128, and the effective rank of each factor representation. Greater effective rank preserves more diverse directions, maintaining better generalization on the unseen fisher components.}
    \resizebox{1\linewidth}{!}{%
\begin{tabular}{cc|ccc|ccc|ccc|ccc}
\toprule
\midrule
 & & \multicolumn{6}{c}{\gls{llama3}.2-3B} & \multicolumn{6}{c}{\gls{qwen3}-4B} \\
 & & \multicolumn{3}{c}{$\Delta F$} & \multicolumn{3}{c}{Eff.Rank} & \multicolumn{3}{c}{$\Delta F$} & \multicolumn{3}{c}{Eff.Rank} \\
\cmidrule(lr){3-8} \cmidrule(lr){9-14}
Idx & Layer & \gls{kfac} & \gls{gfwsvd} & $\times$ & \gls{kfac} & \gls{gfwsvd} & $\times$ & \gls{kfac} & \gls{gfwsvd} & $\times$ & \gls{kfac} & \gls{gfwsvd} & $\times$ \\
\midrule
\multirow{7}{*}{0} & q\_proj & 5.4e-07 & 1.4e-06 & 0.39 & 7570.9 & 636.1 & 11.90 & 5.3e-04 & 5.6e-04 & 0.93 & 169327.1 & 2675.6 & 63.29 \\
 & k\_proj & 5.3e-07 & 8.0e-07 & 0.66 & 1548.4 & 545.3 & 2.84 & 8.8e-04 & 1.1e-03 & 0.81 & 67210.2 & 920.8 & 72.99 \\
 & v\_proj & 2.5e-05 & 5.2e-05 & 0.49 & 44295.1 & 181.1 & 244.62 & 4.6e-04 & 1.0e-03 & 0.46 & 93505.1 & 380.8 & 245.54 \\
 & o\_proj & 4.8e-05 & 9.5e-05 & 0.51 & 255441.8 & 230.7 & 1107.10 & 5.0e-03 & 7.6e-03 & 0.66 & 259343.8 & 370.3 & 700.37 \\
 & up\_proj & 2.3e-04 & 3.6e-04 & 0.65 & 3133342.0 & 63.7 & 49225.85 & 8.3e-04 & 9.1e-04 & 0.91 & 2655398.0 & 1449.3 & 1832.19 \\
 & down\_proj & 3.0e-04 & 5.5e-04 & 0.55 & 1972431.0 & 90.1 & 21899.57 & 1.3e-03 & 2.0e-03 & 0.63 & 2969225.0 & 340.4 & 8722.68 \\
 & \textbf{Avg} & \textbf{1.0e-04} & \textbf{1.8e-04} & \textbf{0.54} & \textbf{902438.2} & \textbf{291.2} & \textbf{12081.98} & \textbf{1.5e-03} & \textbf{2.2e-03} & \textbf{0.73} & \textbf{1035668.2} & \textbf{1022.9} & \textbf{1939.51} \\
\midrule
\multirow{7}{*}{15} & q\_proj & 9.2e-06 & 2.2e-05 & 0.41 & 79847.3 & 9279.5 & 8.60 & 1.1e-04 & 3.1e-04 & 0.34 & 20170.5 & 12.4 & 1625.46 \\
 & k\_proj & 4.4e-06 & 9.6e-06 & 0.46 & 26065.4 & 1111.5 & 23.45 & 7.9e-05 & 1.3e-04 & 0.63 & 20775.3 & 28.3 & 733.45 \\
 & v\_proj & 6.6e-05 & 6.3e-05 & 1.05 & 305087.9 & 661.5 & 461.17 & 1.4e-04 & 8.5e-04 & 0.16 & 28784.4 & 10.1 & 2861.56 \\
 & o\_proj & 7.9e-05 & 8.2e-05 & 0.97 & 673192.2 & 5528.5 & 121.77 & 3.2e-04 & 7.6e-04 & 0.42 & 178611.0 & 24.9 & 7174.04 \\
 & up\_proj & 1.3e-04 & 2.1e-04 & 0.63 & 2853569.0 & 4045.4 & 705.38 & 9.1e-04 & 6.1e-03 & 0.15 & 550894.1 & 12.7 & 43284.68 \\
 & down\_proj & 1.8e-04 & 1.9e-04 & 0.93 & 4104151.0 & 4575.7 & 896.94 & 6.0e-04 & 2.0e-03 & 0.30 & 1766736.0 & 20.9 & 84487.91 \\
 & \textbf{Avg} & \textbf{7.9e-05} & \textbf{9.7e-05} & \textbf{0.74} & \textbf{1340318.8} & \textbf{4200.4} & \textbf{369.55} & \textbf{3.6e-04} & \textbf{1.7e-03} & \textbf{0.33} & \textbf{427661.9} & \textbf{18.2} & \textbf{23361.18} \\
\midrule
\bottomrule
\end{tabular}
}
    \label{tab:lanczos_results}
\end{table}

%% file: content/supp_llama7b.tex
\subsection{\glslink{llama}{Llama}\glsunset{llama}-7B Results}\label{supp:sec:llama_7b_results}
While first-generation \gls{llama}-7B~\cite{touvron2023llama} cannot be considered a recent architecture anymore, it remains a popular benchmark in the compression domain. To allow for comparisons with other related works, we provide results on \gls{llama}-7B (Hugging Face conversion \texttt{jeffwan/llama-7b-hf}) in this section. Following \gls{svdllm}, we use calibration seed 3 for all approaches while reusing the setup as described in the main section. The results for the decompositions are given in \cref{tab:llama_7b_results}, while the different searches are compared in \cref{tab:llama_7b_search_results}. Therefore, between the tables, results for 0.6, 0.7, 0.8 and 0.9 retention (\gls{ours} uniform and with \gls{oursearch}) are provided.

\begin{table}[tp!]
    \centering
    \captionof{table}{Results of all methods on \textbf{\gls{llama}-7B} at uniform compression rates of 0.9 and 0.7. \textbf{Bold}/\underline{Underline} denote best/second-best compressed results.}
    \resizebox{1\textwidth}{!}{%
        \input{data/tables/llama7b_results}
    }
    \label{tab:llama_7b_results}
\end{table}

\begin{table}[tp!]
    \centering
    \captionof{table}{Comparison of searches on \textbf{\gls{llama}-7B} with \gls{ours} at 0.8 and 0.6 parameter retention. \textbf{Bold}/\underline{Underline} denote best/second-best results.}
    \resizebox{1\textwidth}{!}{%
        \input{data/tables/llama7b_search_results}
    }
    \label{tab:llama_7b_search_results}
\end{table}

%% file: data/tables/llama7b_results.tex
\begin{tabular}{l|c|cccccccccc}
\toprule
\midrule
 &\multicolumn{1}{c|}{Method} & \glslink{wiki}{Wiki}\glsunset{wiki} $\downarrow$ & \glslink{c4}{C4}\glsunset{c4} $\downarrow$ & \glslink{ptb}{PTB}\glsunset{ptb} $\downarrow$ & Openb. $\uparrow$ & ARC\_e $\uparrow$ & ARC\_c $\uparrow$ & WinoG. $\uparrow$ & HellaS. $\uparrow$ & PIQA $\uparrow$ & \textbf{Avg. $\uparrow$} \\ \midrule
 & \multicolumn{1}{c|}{\color[HTML]{9B9B9B}Original} & {\color[HTML]{9B9B9B}5.68} & {\color[HTML]{9B9B9B}7.34} & {\color[HTML]{9B9B9B}8.35} & {\color[HTML]{9B9B9B}0.31} & {\color[HTML]{9B9B9B}0.75} & {\color[HTML]{9B9B9B}0.41} & {\color[HTML]{9B9B9B}0.70} & {\color[HTML]{9B9B9B}0.56} & {\color[HTML]{9B9B9B}0.78} & {\color[HTML]{9B9B9B}58.50} \\
\midrule
\multirow{8}{*}{0.9} & \multicolumn{1}{c|}{\acrshort{fwsvd}} & 43.48 & 72.10 & 77.97 & 0.15 & 0.40 & 0.25 & 0.54 & 0.31 & 0.60 & 37.69 \\
 & \multicolumn{1}{c|}{\acrshort{asvd}} & 9.96 & 13.70 & 14.69 & 0.25 & 0.63 & 0.33 & 0.64 & 0.47 & 0.71 & 50.66 \\
 & \multicolumn{1}{c|}{\acrshort{svdllm}} & 7.07 & 12.04 & 12.42 & \underline{0.26} & 0.62 & 0.32 & \underline{0.65} & 0.45 & 0.71 & 50.30 \\
 & \multicolumn{1}{c|}{\acrshort{svdllmv2}} & 7.07 & 12.05 & 12.43 & \underline{0.26} & 0.62 & 0.32 & \underline{0.65} & 0.45 & 0.71 & 50.29 \\
 & \multicolumn{1}{c|}{\acrshort{dobisvd}} & 7.69 & 12.54 & 12.53 & 0.24 & 0.60 & 0.32 & 0.64 & 0.46 & 0.70 & 49.28 \\
 & \multicolumn{1}{c|}{\acrshort{gfwsvd}} & 112.09 & 256.17 & 294.54 & 0.19 & 0.46 & 0.24 & 0.55 & 0.33 & 0.62 & 39.60 \\
 & \multicolumn{1}{c|}{\midrowcolor{m} \textbf{\gls{ours}}} & \underline{6.62} & \underline{10.52} & \underline{10.70} & \underline{0.26} & \underline{0.70} & \underline{0.37} & \underline{0.65} & \underline{0.48} & \underline{0.73} & \underline{53.15} \\
 & \multicolumn{1}{c|}{\midrowcolor{m} \textbf{$+$\gls{oursearch}}} & \textbf{5.77} & \textbf{7.62} & \textbf{8.54} & \textbf{0.31} & \textbf{0.74} & \textbf{0.41} & \textbf{0.69} & \textbf{0.56} & \textbf{0.78} & \textbf{58.15} \\
\midrule
\multirow{8}{*}{0.7} & \multicolumn{1}{c|}{\acrshort{fwsvd}} & 280.04 & 369.17 & 399.63 & 0.13 & 0.30 & 0.20 & 0.49 & 0.27 & 0.56 & 32.43 \\
 & \multicolumn{1}{c|}{\acrshort{asvd}} & 35.55 & 45.03 & 50.00 & 0.20 & 0.46 & 0.28 & 0.58 & 0.35 & 0.63 & 41.85 \\
 & \multicolumn{1}{c|}{\acrshort{svdllm}} & 9.50 & 25.80 & 27.10 & \underline{0.23} & 0.52 & 0.27 & 0.60 & 0.36 & 0.64 & 43.61 \\
 & \multicolumn{1}{c|}{\acrshort{svdllmv2}} & 9.52 & 25.89 & 27.24 & \underline{0.23} & 0.52 & 0.26 & 0.60 & 0.36 & 0.64 & 43.60 \\
 & \multicolumn{1}{c|}{\acrshort{dobisvd}} & 10.82 & 25.22 & 22.40 & 0.20 & 0.45 & 0.25 & 0.58 & 0.35 & 0.62 & 40.90 \\
 & \multicolumn{1}{c|}{\acrshort{gfwsvd}} & 1462.89 & 2024.91 & 2784.08 & 0.13 & 0.30 & 0.21 & 0.54 & 0.26 & 0.54 & 33.23 \\
 & \multicolumn{1}{c|}{\midrowcolor{m} \textbf{\gls{ours}}} & \underline{8.51} & \underline{19.23} & \underline{16.99} & \underline{0.23} & \underline{0.61} & \underline{0.29} & \underline{0.61} & \underline{0.38} & \underline{0.65} & \underline{46.09} \\
 & \multicolumn{1}{c|}{\midrowcolor{m} \textbf{$+$\gls{oursearch}}} & \textbf{7.39} & \textbf{13.55} & \textbf{14.14} & \textbf{0.25} & \textbf{0.65} & \textbf{0.33} & \textbf{0.63} & \textbf{0.43} & \textbf{0.70} & \textbf{49.88} \\
\midrule
\bottomrule
\end{tabular}

%% file: data/tables/llama7b_search_results.tex
\begin{tabular}{l|c|cccccccccc}
\toprule
\midrule
 &\multicolumn{1}{c|}{Search Method} & \glslink{wiki}{Wiki}\glsunset{wiki} $\downarrow$ & \glslink{c4}{C4}\glsunset{c4} $\downarrow$ & \glslink{ptb}{PTB}\glsunset{ptb} $\downarrow$ & Openb. $\uparrow$ & ARC\_e $\uparrow$ & ARC\_c $\uparrow$ & WinoG. $\uparrow$ & HellaS. $\uparrow$ & PIQA $\uparrow$ & \textbf{Avg. $\uparrow$} \\ \midrule
 & \multicolumn{1}{c|}{\color[HTML]{9B9B9B}Original} & {\color[HTML]{9B9B9B}5.68} & {\color[HTML]{9B9B9B}7.34} & {\color[HTML]{9B9B9B}8.35} & {\color[HTML]{9B9B9B}0.31} & {\color[HTML]{9B9B9B}0.75} & {\color[HTML]{9B9B9B}0.41} & {\color[HTML]{9B9B9B}0.70} & {\color[HTML]{9B9B9B}0.56} & {\color[HTML]{9B9B9B}0.78} & {\color[HTML]{9B9B9B}58.50} \\
\midrule
\multirow{7}{*}{0.8} & \multicolumn{1}{c|}{uniform} & 7.27 & 13.26 & 12.75 & 0.24 & 0.67 & 0.34 & 0.63 & 0.43 & 0.70 & 50.08 \\
 & \multicolumn{1}{c|}{\acrshort{asvd}} & \underline{6.67} & 10.68 & \underline{11.89} & \textbf{0.29} & \underline{0.70} & \textbf{0.38} & \textbf{0.67} & \underline{0.48} & \underline{0.74} & \underline{54.38} \\
 & \multicolumn{1}{c|}{\acrshort{svdllmv2}} & 7.27 & 13.20 & 12.75 & 0.24 & 0.67 & 0.34 & 0.63 & 0.43 & 0.69 & 50.23 \\
 & \multicolumn{1}{c|}{\acrshort{mrcs}} & 7.26 & 13.19 & 12.37 & 0.25 & 0.66 & 0.33 & 0.64 & 0.44 & 0.70 & 50.32 \\
 & \multicolumn{1}{c|}{\acrshort{ars}} & 7.29 & \underline{10.67} & 12.49 & \underline{0.27} & 0.69 & 0.34 & 0.62 & \underline{0.48} & \underline{0.74} & 52.19 \\
 & \multicolumn{1}{c|}{\acrshort{atp}} & 7.25 & 12.95 & 12.66 & 0.25 & 0.68 & \underline{0.35} & \underline{0.65} & 0.44 & 0.70 & 51.17 \\
 & \multicolumn{1}{c|}{\midrowcolor{m} \textbf{\gls{oursearch}}} & \textbf{6.23} & \textbf{9.15} & \textbf{10.34} & \textbf{0.29} & \textbf{0.72} & \textbf{0.38} & \textbf{0.67} & \textbf{0.51} & \textbf{0.76} & \textbf{55.37} \\
\midrule
\multirow{7}{*}{0.6} & \multicolumn{1}{c|}{uniform} & 11.12 & 32.94 & 31.43 & 0.19 & 0.48 & 0.25 & 0.57 & 0.33 & 0.60 & 40.45 \\
 & \multicolumn{1}{c|}{\acrshort{asvd}} & 10.95 & 34.37 & 35.09 & 0.18 & 0.50 & 0.25 & 0.56 & 0.33 & 0.61 & 40.51 \\
 & \multicolumn{1}{c|}{\acrshort{svdllmv2}} & 11.10 & 32.77 & 31.56 & \underline{0.20} & 0.47 & 0.26 & 0.57 & 0.33 & 0.60 & 40.42 \\
 & \multicolumn{1}{c|}{\acrshort{mrcs}} & 14.32 & 54.34 & 47.13 & 0.15 & 0.40 & 0.23 & 0.54 & 0.30 & 0.58 & 36.70 \\
 & \multicolumn{1}{c|}{\acrshort{ars}} & 12.89 & \underline{21.87} & 30.55 & 0.19 & \underline{0.54} & 0.25 & 0.55 & 0.35 & \textbf{0.66} & 42.28 \\
 & \multicolumn{1}{c|}{\acrshort{atp}} & \underline{10.36} & 26.81 & \underline{27.25} & \underline{0.20} & \underline{0.54} & \underline{0.28} & \underline{0.59} & \underline{0.36} & \underline{0.63} & \underline{43.21} \\
 & \multicolumn{1}{c|}{\midrowcolor{m} \textbf{\gls{oursearch}}} & \textbf{9.17} & \textbf{18.98} & \textbf{21.16} & \textbf{0.22} & \textbf{0.59} & \textbf{0.30} & \textbf{0.60} & \textbf{0.39} & \textbf{0.66} & \textbf{46.15} \\
\midrule
\bottomrule
\end{tabular}

%% file: content/supp_concurrent.tex
\subsection{Comparison with Concurrent Work}\label{supp:sec:concurrent}
Multiple \gls{svd} works have been proposed simultaneously to the original submission of this work and its final publication. To explore the capabilities of these related works, we extend our analysis with the four most related methods. Similar to our \gls{ours}, both OBD-LLM~\cite{li2026optimalbraindecompositionaccurate} and IO-SVD~\cite{abbasi2026iosvdinputoutputwhitenedsvd} rely on double sided whitening, while ZS-SVD~\cite{abbasi2026zero} and Swift-SVD~\cite{qi2026swiftsvd} propose new rank allocation strategy concurrent to \gls{oursearch}. We reimplemented all methods based on their public code or formal algorithm and verified the implementation against the \gls{llama}-7B or \gls{llama}2-7B results originally reported in the respetive publication.

\paragraph{Double-sided decompositions.}
OBD-LLM takes a principled approach that derives the decomposition from a Kronecker factorization of the layer Hessian. Since it also relies on the same \gls{kfac} Fisher approximation as \gls{ours} it is \emph{very} close to our work. The main practical difference is the factor damping. OBD-LLM damps the factors of both sides by $0.1$ times their mean diagonal, whereas \gls{ours} shrinks both factors towards a scaled identity (\cref{alg:kfac_svd}), with a strength of $0.1$ on the input and $0.3$ on the output side. IO-SVD is also close and, again, uses the same input factor as OBD-LLM and \gls{ours}. However, for the output factor it uses \gls{kl} divergence over the top-$K$ next-token probabilities to obtain a second-order expansion for the factor. The number of top-k values, that are considered in the expansion varies with the compression ratio ($K=5$ at 0.8, $K=4$ at 0.6). Notably, IO-SVD is more expensive than OBD-LLM and \gls{ours} as it requires $K$ backward passes for its construction rather than just one.

Since IO-SVD's code is not currently public, we reimplemented it based on the paper. However, as its damping and probe construction are not fully disclosed, we cannot recover their results exactly. In our testing, for example, a top-$K$ of $K=8$ works better than the $K=5$ (0.8) and $K=4$ (0.6) selected in the paper, so we use $K=8$ throughout. On \gls{llama}-7B, our IO-SVD implementation achieves a \gls{wiki} perplexity of 7.41 / 11.45 (reference: 7.31 / 11.20), for 0.8 and 0.6 retention, respectively. Hence, our reimplementation results are closely matching the reference. For OBD-LLM, which also has no publicly available code, we test our implementation on Llama 2-7B using 128 calibration sequences of 2048 tokens from \gls{c4}, uniform ranks, damping of $0.1\times$ the mean diagonal on both factors and temperature $T=1$, as reported in the main results of the original work. With those settings, at 0.8 retention, our implementation achieves 11.27 \gls{wiki} perplexity vs 11.50 of the reference, at 0.7 ours is 19.16 vs 18.83 of the reference, at 0.6, ours is 39.17 vs 36.39 of the reference. Hence, our implementation is reasonably close to the reference, considering the missing code reference.

\cref{tab:concurrent_decompositions} directly compares the decomposition approaches against each other without any dedicated search or post-compression parameter update applied across models from multiple LLM-families. \gls{ours} achieves the lowest perplexity on \gls{wiki} across all models and highest average accuracy for 3 out of the 4 models. \emph{To note}: in some cases, the performance gaps between the methods remain relatively close. At a ratio of 0.8 in particular, \gls{llama}-7B perplexities are showing only minor deviations, which could well be within the  minor implementation deviations we observed during testing, making it difficult to point out one decomposition as materially better.

\paragraph{Rank allocation.}
ZS-SVD~\cite{abbasi2026zero} estimates the change in calibration loss caused by the removal of singular values. It then aims to remove singular values such that the cumulative predicted loss change stays close to zero. For the comparison, we use the one-shot version that does not perform iterative compression with gradient-based refinement, as none of the other approaches are finetuned before, during or after search either. Following their reference implementation, the degradation is estimated using just one sequence for the predicted loss degradation since using more did degrade the results for us during testing. Swift-SVD~\cite{qi2026swiftsvd} uses a layer-importance and reconstruction loss based scoring to distribute the budget over the model. The trade-off between reconstruction and importance impact is estimated using an $\alpha$ hyperparameter, that is tuned in a couple of fast attempts, similar to our bias parameters.

Since both searches have been proposed on top of \gls{svdllm}, to ensure the validity of our reimplementations, we run them on top of \gls{svdllm} first. Both reimplementations reproduce the published \gls{llama}-7B results on \gls{svdllm}: Swift-SVD gives 7.89 / 13.49 / 63.90 perplexity at 0.8 / 0.6 / 0.4 (reference: 7.84 / 13.29 / 62.32, denoted as Swift-SVD$^*$ in the original work), and ZS-SVD 6.70 / 11.30 / 40.48 (reference: 6.74 / 11.44 / 45.17). Importantly, to reproduce ZS-SVD, the damping of $\mathbf{X}\mathbf{X}^\top/N + 10^{-4}\mathbf{I}$, which differs from the original \gls{svdllm} implementation, is required.

\cref{tab:concurrent_searches} compares the performance of the different searches when applied to \gls{ours} across the same four models across 0.8 and 0.6 retentions. ZS-SVD achieves improvements in most settings, and is second best for all 0.8 results. It still remains effective for \gls{mistral} and \gls{llama3} on 0.6, but is outperformed by uniform compression in \gls{qwen3} and \gls{llama}-7B. While its gradient-projection updates can help to recover some of its losses, the comparison is strictly between non-updated models. However, even when considering the update, the ZS-SVD 5X reported in the original work achieves a \gls{wiki} perplexity of 9.45 for \gls{llama}-7B, thereby still resulting in higher perplexity than what \gls{oursearch} achieves without any update. Swift-SVD remains close to uniform perplexity in most cases, only getting significantly better-than-uniform results on \gls{llama3} at 0.6 (26.77 vs.\ 33.97). Overall, even in the extended comparison, \gls{oursearch} substantially improves the results in almost all cases highlighting its strong capability even when compared to concurrent work.

\begin{table}[tp!]
    \centering
    \captionof{table}{Decompositions of concurrent work at uniform ranks and 0.8 and 0.6 parameter retention: \gls{wiki} perplexity and average zero-shot accuracy. \textbf{Bold}/\underline{Underline}: best/second best.}
        \input{data/tables/concurrent_decompositions}
    \label{tab:concurrent_decompositions}
\end{table}

\begin{table}[tp!]
    \centering
    \captionof{table}{Rank allocations of concurrent work applied to \gls{ours} at 0.8 and 0.6 parameter retention: \gls{wiki} perplexity and average zero-shot accuracy. \textbf{Bold}/\underline{Underline}: best/second best.}
        \input{data/tables/concurrent_searches}
    \label{tab:concurrent_searches}
\end{table}

%% file: data/tables/concurrent_decompositions.tex
\begin{tabular}{ll|cc|cc|cc|cc}
\toprule
\midrule
& & \multicolumn{2}{c|}{\gls{mistral}-7B} & \multicolumn{2}{c|}{\gls{llama3}-8B} & \multicolumn{2}{c|}{\gls{qwen3}-8B} & \multicolumn{2}{c}{\gls{llama}-7B} \\
Ratio & Decomposition & Wiki $\downarrow$ & Avg. $\uparrow$ & Wiki $\downarrow$ & Avg. $\uparrow$ & Wiki $\downarrow$ & Avg. $\uparrow$ & Wiki $\downarrow$ & Avg. $\uparrow$ \\
\midrule
\multirow{4}{*}{0.8} & \acrshort{svdllm} & 8.15 & \underline{51.61} & 14.61 & 44.84 & 14.05 & 52.31 & 7.89 & 44.65 \\
 & OBD-LLM & 7.33 & 50.88 & \underline{11.18} & 49.42 & \textbf{12.19} & \underline{52.66} & 7.42 & \underline{48.88} \\
 & IO-SVD & \underline{7.27} & 50.88 & 11.23 & \underline{49.54} & 12.39 & 52.40 & \underline{7.41} & 48.68 \\
 & \midrowcolor{m} \textbf{\gls{ours}} & \textbf{7.08} & \textbf{52.67} & \textbf{10.61} & \textbf{50.84} & \underline{12.25} & \textbf{54.18} & \textbf{7.27} & \textbf{50.08} \\
\midrule
\multirow{4}{*}{0.6} & \acrshort{svdllm} & 16.12 & 37.71 & 136.19 & 32.97 & 28.77 & 36.81 & 13.74 & 37.14 \\
 & OBD-LLM & 18.13 & 36.91 & \underline{40.39} & 35.83 & \underline{19.16} & \underline{39.49} & 11.80 & \underline{39.12} \\
 & IO-SVD & \underline{15.57} & \underline{37.82} & 45.02 & \textbf{36.29} & 20.12 & 39.40 & \underline{11.45} & 38.93 \\
 & \midrowcolor{m} \textbf{\gls{ours}} & \textbf{14.47} & \textbf{39.22} & \textbf{33.97} & \underline{36.24} & \textbf{18.00} & \textbf{41.16} & \textbf{11.12} & \textbf{40.45} \\
\midrule
\bottomrule
\end{tabular}

%% file: data/tables/concurrent_searches.tex
\begin{tabular}{ll|cc|cc|cc|cc}
\toprule
\midrule
& & \multicolumn{2}{c|}{\gls{mistral}-7B} & \multicolumn{2}{c|}{\gls{llama3}-8B} & \multicolumn{2}{c|}{\gls{qwen3}-8B} & \multicolumn{2}{c}{\gls{llama}-7B} \\
Ratio & Search & Wiki $\downarrow$ & Avg. $\uparrow$ & Wiki $\downarrow$ & Avg. $\uparrow$ & Wiki $\downarrow$ & Avg. $\uparrow$ & Wiki $\downarrow$ & Avg. $\uparrow$ \\
\midrule
\multirow{4}{*}{0.8} & uniform & 7.08 & 52.67 & 10.61 & 50.84 & 12.25 & 54.18 & 7.27 & 50.08 \\
 & ZS-SVD & \underline{6.30} & \underline{57.18} & \underline{8.72} & \underline{55.54} & \underline{11.20} & \underline{58.46} & \underline{6.49} & \underline{53.85} \\
 & Swift-SVD & 7.04 & 52.91 & 10.04 & 51.82 & 12.28 & 54.49 & 7.25 & 50.36 \\
 & \midrowcolor{m} \textbf{\gls{oursearch}} & \textbf{5.92} & \textbf{57.61} & \textbf{8.04} & \textbf{56.77} & \textbf{10.28} & \textbf{59.03} & \textbf{6.23} & \textbf{55.37} \\
\midrule
\multirow{4}{*}{0.6} & uniform & 14.47 & 39.22 & 33.97 & 36.24 & 18.00 & 41.16 & 11.12 & 40.45 \\
 & ZS-SVD & \underline{12.55} & \underline{42.23} & \underline{22.11} & \underline{41.70} & 21.39 & \textbf{46.63} & 11.19 & 37.68 \\
 & Swift-SVD & 14.24 & 40.03 & 26.77 & 38.60 & \underline{17.31} & 41.42 & \underline{10.99} & \underline{40.64} \\
 & \midrowcolor{m} \textbf{\gls{oursearch}} & \textbf{10.32} & \textbf{43.84} & \textbf{16.08} & \textbf{44.19} & \textbf{15.07} & \underline{45.60} & \textbf{9.17} & \textbf{46.15} \\
\midrule
\bottomrule
\end{tabular}

%% file: data/tables/llama2_results.tex
\begin{tabular}{l|c|cccccccccc}
\toprule
\midrule
 &\multicolumn{1}{c|}{Method} & \glslink{wiki}{Wiki}\glsunset{wiki} $\downarrow$ & \glslink{c4}{C4}\glsunset{c4} $\downarrow$ & \glslink{ptb}{PTB}\glsunset{ptb} $\downarrow$ & Openb. $\uparrow$ & ARC\_e $\uparrow$ & ARC\_c $\uparrow$ & WinoG. $\uparrow$ & HellaS. $\uparrow$ & PIQA $\uparrow$ & \textbf{Avg. $\uparrow$} \\ \midrule
 & \multicolumn{1}{c|}{\color[HTML]{9B9B9B}Original} & {\color[HTML]{9B9B9B}5.47} & {\color[HTML]{9B9B9B}7.29} & {\color[HTML]{9B9B9B}26.82} & {\color[HTML]{9B9B9B}0.33} & {\color[HTML]{9B9B9B}0.76} & {\color[HTML]{9B9B9B}0.44} & {\color[HTML]{9B9B9B}0.69} & {\color[HTML]{9B9B9B}0.57} & {\color[HTML]{9B9B9B}0.78} & {\color[HTML]{9B9B9B}59.51} \\
\midrule
\multirow{7}{*}{0.8}& \multicolumn{1}{c|}{\acrshort{fwsvd}} & 173.51 & 224.54 & 4138.65 & 0.15 & 0.29 & 0.21 & 0.50 & 0.28 & 0.55 & 33.10 \\
& \multicolumn{1}{c|}{\acrshort{asvd}} & 25.24 & 32.73 & 239.02 & 0.26 & 0.54 & 0.28 & 0.60 & 0.40 & 0.65 & 45.44 \\
& \multicolumn{1}{c|}{\acrshort{svdllm}} & 8.42 & 19.97 & 129.95 & 0.26 & 0.57 & 0.27 & 0.63 & 0.40 & 0.67 & 46.66 \\
& \multicolumn{1}{c|}{\acrshort{dobisvd}} & 9.58 & 20.20 & 102.00 & 0.24 & 0.52 & 0.24 & 0.60 & 0.38 & 0.64 & 43.80 \\
& \multicolumn{1}{c|}{\acrshort{gfwsvd}} & 63.09 & 121.61 & 881.17 & 0.15 & 0.34 & 0.21 & 0.53 & 0.30 & 0.58 & 34.97 \\
& \multicolumn{1}{c|}{\midrowcolor{m} \textbf{\gls{ours}}} & \underline{7.31} & \underline{15.04} & \underline{53.86} & \underline{0.27} & \underline{0.67} & \underline{0.32} & \underline{0.64} & \underline{0.43} & \underline{0.70} & \underline{50.77} \\
& \multicolumn{1}{c|}{\midrowcolor{m} \textbf{$+$\gls{oursearch}}} & \textbf{6.05} & \textbf{9.23} & \textbf{44.30} & \textbf{0.32} & \textbf{0.73} & \textbf{0.39} & \textbf{0.67} & \textbf{0.52} & \textbf{0.76} & \textbf{56.66} \\
\midrule\multirow{7}{*}{0.6} & \multicolumn{1}{c|}{\acrshort{fwsvd}} & 699.79 & 964.00 & 5150.63 & 0.11 & 0.27 & 0.20 & 0.50 & 0.26 & 0.54 & 31.37 \\
& \multicolumn{1}{c|}{\acrshort{asvd}} & 3004.34 & 3595.72 & 3124.02 & 0.13 & 0.28 & 0.19 & 0.50 & 0.27 & 0.54 & 31.71 \\
& \multicolumn{1}{c|}{\acrshort{svdllm}} & 16.27 & 72.05 & 786.80 & 0.19 & 0.39 & 0.21 & 0.56 & 0.31 & 0.58 & 37.42 \\
& \multicolumn{1}{c|}{\acrshort{dobisvd}} & 18.50 & 60.67 & 457.14 & 0.19 & 0.34 & 0.20 & 0.53 & 0.30 & 0.57 & 35.76 \\
& \multicolumn{1}{c|}{\acrshort{gfwsvd}} & 2946.23 & 3738.96 & 3680.99 & 0.14 & 0.26 & \underline{0.22} & 0.50 & 0.26 & 0.54 & 31.88 \\
& \multicolumn{1}{c|}{\midrowcolor{m} \textbf{\gls{ours}}} & \underline{12.25} & \underline{42.44} & \underline{187.61} & \underline{0.20} & \underline{0.46} & \underline{0.22} & \underline{0.58} & \underline{0.33} & \underline{0.60} & \underline{39.99} \\
& \multicolumn{1}{c|}{\midrowcolor{m} \textbf{$+$\gls{oursearch}}} & \textbf{9.43} & \textbf{20.84} & \textbf{110.72} & \textbf{0.24} & \textbf{0.59} & \textbf{0.29} & \textbf{0.63} & \textbf{0.39} & \textbf{0.65} & \textbf{46.48} \\
\midrule
\bottomrule
\end{tabular}

%% file: data/tables/ablation_basis_sharing_lems.tex
\begin{tabular}{c|cccccccccc}
\toprule
\midrule
 \multicolumn{1}{c|}{Method} & \glslink{wiki}{Wiki}\glsunset{wiki} $\downarrow$ & \glslink{c4}{C4}\glsunset{c4} $\downarrow$ & \glslink{ptb}{PTB}\glsunset{ptb} $\downarrow$ & Openb. $\uparrow$ & ARC\_e $\uparrow$ & ARC\_c $\uparrow$ & WinoG. $\uparrow$ & HellaS. $\uparrow$ & PIQA $\uparrow$ & \textbf{Avg. $\uparrow$} \\ \midrule
\multicolumn{1}{c|}{\color[HTML]{9B9B9B}Original} & {\color[HTML]{9B9B9B}5.47} & {\color[HTML]{9B9B9B}7.29} & {\color[HTML]{9B9B9B}24.09} & {\color[HTML]{9B9B9B}0.33} & {\color[HTML]{9B9B9B}0.75} & {\color[HTML]{9B9B9B}0.43} & {\color[HTML]{9B9B9B}0.69} & {\color[HTML]{9B9B9B}0.57} & {\color[HTML]{9B9B9B}0.78} & {\color[HTML]{9B9B9B}59.43} \\
\midrule
\multicolumn{1}{c|}{\glslink{basissharing}{BasisSharing}\glsunset{basissharing} (Ref. data + fp32)} & 7.71  & 16.21 & \textbf{59.47} & 0.29 & 0.63 & 0.31 & 0.64 & 0.44 & 0.70 & 50.17 \\

\multicolumn{1}{c|}{\gls{basissharing} (Our data + fp16)} & 8.12 & 19.35 & 106.48 & 0.26 & 0.53 & 0.24 & 0.63 & 0.42 & 0.67 & 45.83 \\

\multicolumn{1}{c|}{\midrowcolor{m} \textbf{$+$\gls{oursearch}}} & \textbf{6.25} & \textbf{9.83} & 297.47 & \textbf{0.32} & \textbf{0.72} & \textbf{0.39} & \textbf{0.67} & \textbf{0.52} & \textbf{0.75} & \textbf{56.17} \\

\midrule
\bottomrule
\end{tabular}

%% file: data/tables/ablation_basis_sharing_matched.tex
\begin{tabular}{|c|cc|cccccccccc}
\toprule
\midrule
 \multicolumn{1}{c|}{Method} & GParams & TFLOPs &\glslink{wiki}{Wiki}\glsunset{wiki} $\downarrow$ & \glslink{c4}{C4}\glsunset{c4} $\downarrow$ & \glslink{ptb}{PTB}\glsunset{ptb} $\downarrow$ & Openb. $\uparrow$ & ARC\_e $\uparrow$ & ARC\_c $\uparrow$ & WinoG. $\uparrow$ & HellaS. $\uparrow$ & PIQA $\uparrow$ & \textbf{Avg. $\uparrow$} \\ \midrule
 \multicolumn{1}{c|}{\color[HTML]{9B9B9B}Original} & {\color[HTML]{9B9B9B}6.48} & {\color[HTML]{9B9B9B}26.53} & {\color[HTML]{9B9B9B}5.47} & {\color[HTML]{9B9B9B}7.29} & {\color[HTML]{9B9B9B}24.09} & {\color[HTML]{9B9B9B}0.33} & {\color[HTML]{9B9B9B}0.75} & {\color[HTML]{9B9B9B}0.43} & {\color[HTML]{9B9B9B}0.69} & {\color[HTML]{9B9B9B}0.57} & {\color[HTML]{9B9B9B}0.78} & {\color[HTML]{9B9B9B}59.43} \\
\midrule
\midrule
\multicolumn{6}{l}{Matched \textbf{Compute} Budget: 0.8 $\cdot$ Original Linear Layer Compute} &&&&&\\
\midrule
\multicolumn{1}{c|}{\gls{basissharing}} & 4.53 & \textbf{21.40} & 10.16 & 32.60 & 231.67 & 0.25 & 0.51 & 0.24 & 0.62 & 0.36 & 0.63 & 43.50 \\

\multicolumn{1}{c|}{\gls{ours}} & 5.18 & \textbf{21.22} & \underline{7.31} & \underline{15.04} & \underline{53.86} & \underline{0.27} & \underline{0.67} & \underline{0.32} & \underline{0.64} & \underline{0.43} & \underline{0.70} & \underline{50.77} \\
\midrule
\midrule
\multicolumn{6}{l}{Matched \textbf{Memory} Budget: 0.8 $\cdot$ Original Linear Layer Memory} &&&&&\\
\midrule
\multicolumn{1}{c|}{\gls{basissharing}} & \textbf{5.18} & 24.46 & 8.12 & 19.35 & 106.48 & 0.26 & 0.53 & 0.24 & 0.63 & 0.42 & 0.67 & 45.83 \\
\multicolumn{1}{c|}{\gls{ours}} & \textbf{5.18} & 21.22 & \underline{7.31} & \underline{15.04} & \underline{53.86} & \underline{0.27} & \underline{0.67} & \underline{0.32} & \underline{0.64} & \underline{0.43} & \underline{0.70} & \underline{50.77} \\

\midrule
\bottomrule
\end{tabular}

%% file: data/plots/ablation_seed_and_samples.tex
\begin{tikzpicture}
    \pgfplotsset{
        myplotstyle/.style={
            color=icmlblue, %
            mark=triangle*,
            mark options={scale=1.2}
        },
        label style={font=\LARGE},
        tick label style={font=\small},
        grid=major,
        major grid style={line width=.2pt,draw=gray!50},
        width=4.2cm,
        height=5cm
    }

    \begin{groupplot}[
        group style={
            group size=2 by 1,
            horizontal sep=1.2cm,
            vertical sep=1.5cm
        },
        ylabel near ticks,
        xlabel near ticks
    ]

    \nextgroupplot[
        ylabel={Perplexity ($\downarrow$)},
        xlabel={Sample count},
        xmin=0.5, xmax=4.5,
        ymin=16.67, ymax=21.83,
        xtick={1, 2, 3, 4},
        xticklabels={64, 128, 256, 512},
        height=4cm,
    ]
    \addplot[myplotstyle] coordinates {
        (1, 20.54)
        (2, 18.56)
        (3, 18.00)
        (4, 17.54)
    };

    \nextgroupplot[
        xlabel={Sample seed},
        xmin=0.5, xmax=4.5,
        ymin=17.65, ymax=19.75,
        xtick={3, 17, 42, 100},
        xtick={1, 2, 3, 4},
        xticklabels={0, 42, 1198, 2025},
        xticklabel style={font=\scriptsize},
        height=4cm,
    ]
    \addplot[myplotstyle] coordinates {
        (1, 19.05)
        (2, 18.00)
        (3, 18.50)
        (4, 18.62)
    };

    \end{groupplot}
\end{tikzpicture}

%% file: data/plots/ablation_throughput.tex
\begin{tikzpicture}
\begin{axis}[
    ybar,
    bar width=0.7cm,
    ylabel={\small{Tokens/s}},
    xlabel={\small{Compression Ratio}},
    ymin=8, ymax=24,
    xmin=0.5, xmax=5.5,
    height=4cm,
    width=\linewidth,
    xtick={1,2,3,4,5},
    xticklabels={Original, 20\%, 40\%, 60\%, 80\%},
    ytick={9,14,19, 24},
    grid=major,
    major grid style={line width=.2pt,draw=gray!50},
    label style={font=\small},
    ylabel style={yshift=-0.5cm},
    tick label style={font=\small},
    scaled y ticks=false,
    enlarge y limits={upper=0},
]

\addplot[fill=icmlblue, draw=none] coordinates {
    (1, 10.02)  %
    (2, 11.38)  %
    (3, 13.40)  %
    (4, 16.16)  %
    (5, 20.17)  %

};

\pgfmathsetmacro{\origval}{2.2}

\node[above, font=\small\bfseries, text=black] at (axis cs:1, 10.02) {1.00x};
\node[above, font=\small\bfseries, text=black] at (axis cs:2, 11.38) {1.14x}; %
\node[above, font=\small\bfseries, text=black] at (axis cs:3, 13.40) {1.34x}; %
\node[above, font=\small\bfseries, text=black] at (axis cs:4, 16.16) {1.61x}; %
\node[above, font=\small\bfseries, text=black] at (axis cs:5, 20.17) {2.01x}; %

\end{axis}
\end{tikzpicture}

%% file: data/tables/fineweb_ablation_cr.tex
\begin{table}[t]
\centering
\captionof{table}{Comparison of \textbf{\gls{ours}} with activation- and Fisher-based \gls{svd} across modern \gls{llm}s under uniform compression and \textbf{\gls{oursearch}} on the \textbf{fineweb 16M} dataset.}
\resizebox{1\textwidth}{!}{%
\begin{tabular}{cc| ccc c| ccc c| ccc c}
\toprule
\midrule
&& \multicolumn{4}{c}{\gls{mistral}-7B} & \multicolumn{4}{c}{\gls{llama3}-8B} & \multicolumn{4}{c}{\gls{qwen3}-8B} \\
\cmidrule(lr){3-6} \cmidrule(lr){7-10} \cmidrule(lr){11-14}
& Method & \glslink{wiki}{Wiki}\glsunset{wiki} $\downarrow$ & \gls{c4} $\downarrow$ & \gls{ptb} $\downarrow$ & Acc $\uparrow$ & \glslink{wiki}{Wiki}\glsunset{wiki} $\downarrow$ & \gls{c4} $\downarrow$ & \gls{ptb} $\downarrow$ & Acc $\uparrow$ & \glslink{wiki}{Wiki}\glsunset{wiki} $\downarrow$ & \gls{c4} $\downarrow$ & \gls{ptb} $\downarrow$ & Acc$\uparrow$ \\
\midrule
& Baseline & 5.25 & 8.10 & 27.72 & 63.96 & 6.14 & 9.47 & 9.90 & 63.33 & 9.71 & 15.52 & 15.43 & 62.04 \\
\midrule
\multirow{8}{*}{0.9} &  \glslink{fwsvd}{FWSVD}\glsunset{fwsvd} & 9.98 & 13.72 & 59.96 & 57.61 & 51.29 & 50.99 & 95.82 & 51.65 & 27.03 & 25.24 & inf & 50.02 \\
 &  \gls{asvd} & 9.49 & 13.22 & 47.53 & 56.73 & 90.72 & 91.79 & 295.15 & 46.77 & 20.52 & 29.17 & 37.16 & 50.90 \\
 &  \gls{svdllm} & 8.23 & 11.20 & 51.89 & 57.32 & 21.93 & 19.58 & 36.73 & 54.11 & 16.23 & 19.85 & 22.06 & 57.89 \\
 &  \gls{svdllmv2} & 8.24 & 11.20 & 51.89 & 57.37 & 21.97 & 19.62 & 36.87 & 54.16 & 16.20 & 19.85 & 22.06 & 57.87 \\
 &  \glslink{dobisvd}{DOBI-SVD}\glsunset{dobisvd} & 9.11 & 12.04 & 58.58 & 55.88 & 22.67 & 20.80 & 28.21 & 52.79 & 15.89 & 20.68 & 23.30 & 56.56 \\
 &  \gls{gfwsvd} & 19.17 & 26.25 & 124.49 & 54.43 & 1678.71 & 1034.22 & 5110.55 & 37.20 & 101.61 & 80.69 & 214.26 & 50.05 \\
 &  \midrowcolor{m}\textbf{\gls{ours}} & \underline{7.46} & \underline{10.56} & \underline{44.04} & \underline{57.66} & \underline{14.41} & \underline{16.17} & \underline{18.61} & \underline{56.74} & \underline{15.34} & \underline{18.69} & \underline{20.68} & \underline{58.43} \\
& \midrowcolor{m}\textbf{$+$\gls{oursearch}} & \textbf{5.53} & \textbf{8.34} & \textbf{29.40} & \textbf{63.11} & \textbf{6.88} & \textbf{10.50} & \textbf{11.09} & \textbf{62.60} & \textbf{10.14} & \textbf{15.83} & \textbf{15.83} & \textbf{62.51} \\
\midrule
\multirow{8}{*}{0.7} &  \gls{fwsvd} & 40.49 & 45.26 & 228.97 & 44.94 & 1006.33 & 705.28 & 2432.99 & 35.33 & 113.35 & 133.04 & inf & 41.20 \\
 &  \gls{asvd} & 31.60 & 40.65 & 152.53 & 44.87 & 11607.13 & 7095.30 & 25751.55 & 32.94 & 73.18 & 87.59 & 129.95 & 42.77 \\
 &  \gls{svdllm} & 29.45 & 21.97 & 138.34 & 45.60 & 181.84 & 85.56 & 391.02 & 41.27 & 44.04 & 30.81 & 69.02 & 46.83 \\
 &  \gls{svdllmv2} & 29.57 & 22.02 & 138.34 & 45.62 & 182.55 & 85.89 & 394.08 & 41.23 & 44.39 & 30.93 & 69.56 & 46.66 \\
 &  \gls{dobisvd} & 28.05 & 24.23 & 140.51 & 44.12 & 154.93 & 73.76 & 195.09 & 40.35 & 42.85 & 31.11 & 66.63 & 48.14 \\
 &  \gls{gfwsvd} & 941.67 & 805.45 & 2180.92 & 36.87 & 26987.39 & 9473.46 & 23264.58 & 32.14 & 12164.17 & 3248.46 & 7095.30 & 35.17 \\
 &  \midrowcolor{m}\textbf{\gls{ours}} & \underline{22.98} & \underline{19.70} & \underline{90.72} & \underline{46.87} & \underline{62.84} & \underline{40.26} & \underline{104.42} & \underline{45.19} & \underline{35.95} & \underline{25.84} & \underline{45.26} & \underline{50.86} \\
& \midrowcolor{m}\textbf{$+$\gls{oursearch}} & \textbf{11.92} & \textbf{12.87} & \textbf{52.40} & \textbf{52.61} & \textbf{19.58} & \textbf{20.40} & \textbf{37.02} & \textbf{51.18} & \textbf{15.86} & \textbf{19.97} & \textbf{24.13} & \textbf{54.89} \\
\midrule
\bottomrule
\end{tabular}
\label{tab:fineweb_ablation}
}%
\end{table}

%% file: data/tables/qualitative_asking_questions.tex
\begin{table}[h!]
    \centering
    \vspace{-0mm}
    \caption{Example content generated as response to "What is the responsibility of the AI assistant?" by \gls{qwen3}-8B models that were compressed with \gls{svdllm} and \gls{ours} without any finetuning under different compression ratios.}\label{tab:qualitative_results_no_ft}
    \resizebox{\linewidth}{!}{
    \tiny
    \begin{tabular}{p{0.8cm}|p{5cm}|p{4.8cm}}
        \toprule
        \midrule
        Ratio & \gls{ours} & \gls{svdllm} \\
        \midrule
        Original & \multicolumn{2}{p{10.2cm}}{\textbf{What is the responsibility of the AI assistant?} And how can it be ensured? The responsibility and accountability of AI assistants are critical issues that require careful consideration. Here's a structured analysis of the key points and a comprehensive approach to ensuring responsibility:

        1. **Ethical and Moral Responsibility**: AI systems should be developed and used in a way that aligns with ethical and moral principles, avoiding harm to individuals and society.
        2. ...

        } \\
        \midrule
         10\%
         & \textbf{What is the responsibility of the AI assistant?} And how can it be ensured? The responsibility (ethical, legal, technical, etc.) of AI assistants is a critical area of concern in the development and application of artificial intelligence technology. This issue is not only related to the fairness and impartiality of the AI system itself but also involves the rights and interests of individuals and society. In order to ensure the correct development of intelligent assistants, we need to establish a comprehensive framework that involves both technical and non-technical dimensions. ...

         & \textbf{What is the responsibility of the AI assistant?} -  - - According to its creators, AI assistants are designed to be helpful and to support users. - = - They should not act as a replacement for human intelligence and should instead focus on assisting users through tasks they can perform. = = -
         - - - However, some AI systems have been designed with specific purposes, such as responding to questions and carrying out tasks that require human thinking. For example, there are AI programs that are used in hospitals that perform complex tasks such  as interpreting X-rays and diagnosing diseases. ... %
         \\
        \midrule
        20\%
        & \textbf{What is the responsibility of the AI assistant?} According to the given prompt, the assistant is supposed to be a helpful companion who provides the best responses. It is not supposed be something of the dark side. So the AI should not be malevolent, but rather a positive and helpful entity. But, in certain situations, even with all the good intentions, perhaps the Dark Side could come into play. Also, there are some guidelines. For example, if there is a request for harm, then the system should refuse. The AI must not provide harmful information. Therefore, to avoid any harm and to maintain a beneficial environment, it is important to ensure that the responses are safe and positive.
        ...

        & \textbf{What is the responsibility of the AI assistant?}  What is the responsibility of an AI assistant? and what should they do to be ethical?

        5.00 PM, 15 March 2021

        Please help me to create a presentation about AI assistants in the modern age.

        Please share your views and opinions about what an ethical AI should be like in today's world.

        I want to present about the AI as assistants to humans and how they are being developed. I am in charge of a department to which AI is being introduced for the first time. ...

        \\
        \midrule
        30\%
        & \textbf{What is the responsibility of the AI assistant?} And the main point is to be on time and precise.

        Which of the four are the most important. Which is most beneficial? Which one is a good solution for the whole team.

        I want to do the assignment for each one, but I need to clarify it. I'm not the best at it, and need help from someone. So I will have to provide some clarification. To get the support of someone who is willing to help, I have the authority. It is not clear, or maybe I don't know about the question and the meaning.  ...

        & \textbf{What is the responsibility of the AI assistant?} and it will give the answer. So to answer questions, the AI may be asked the question by someone, and perhaps I know something about the subject, or perhaps he can be helpful in this situation.

        So the person can explain to the audience that he wants to be the assistant, because of his own identity.  So he has been given to an assistant by the creator and can now become an employee. This idea has to exist in a situation, but the human wants the job. He is responsible for maintaining a position to create AI.

        \\
        \midrule
        \bottomrule
    \end{tabular}
    }
\end{table}

%% file: data/algorithms/overview_algorithm.tex
\begin{algorithm}[h!]
\caption{High-Level Procedure: Modular SVD Compression with Global Search}
\label{alg:high_level}
\begin{algorithmic}[1]
\State \textbf{Input:} pretrained model $M$; calibration sets $\mathcal{C}$ (general), $\mathcal{C}_S$ (sensitivity); budget $P_{\text{target}}$
\State \textbf{Input:} candidate ranks $\{\mathcal{R}_\ell\}$; flag $\texttt{recalibrate}\in\{0,1\}$
\State \textbf{Output:} compressed model $M'$
\Procedure{CompressModel}{$M,\mathcal{C},\mathcal{C}_S,P_{\text{target}},\{\mathcal{R}_\ell\},\texttt{recalibrate}$}
\Statex \hspace{1.5em} \textit{\textbf{\# (I) Decomposition Phase (Modular)}}
\For{each compressible linear layer $\ell$ in $M$}
\State \Comment{Compute SVD factors (e.g., via Alg.~\ref{alg:kfac_svd} or other SVD approaches)}
\State $\mathcal{D}_\ell \gets \Call{DecomposeLayer}{M,\mathcal{C}_S,\ell}$
\EndFor
\Statex \hspace{1.5em} \textit{\textbf{\# (II) Global Rank Allocation (LEMS)}}
\State \Comment{Solve ILP using Shape/Scale/Bias error model (Alg.~\ref{alg:rank_allocation})}
\State $\{r_\ell^\star\}_{\ell} \gets \Call{LEMS\_Search}{M,\{\mathcal{R}_\ell\},\{\mathcal{D}_\ell\}_\ell,P_{\text{target}},\mathcal{C}}$
\Statex \hspace{1.5em} \textit{\textbf{\# (III) Reconstruction Phase}}
\State $M' \gets M$
\For{each compressible layer $\ell$ in forward order}
\If{$\texttt{recalibrate}=1$}
\State $\mathcal{D}_\ell \gets \Call{DecomposeLayer}{M',\mathcal{C},\ell}$ \Comment{Optional: Recalculate on updated $M'$}
\EndIf
\State retrieve $\mathbf{U}_\ell,\mathbf{\Sigma}_\ell,\mathbf{V}_\ell,\mathbf{L}_{A,\ell},\mathbf{L}_{G,\ell}$ from $\mathcal{D}_\ell$
\State \Comment{Reconstruct low-rank weights (example for \gls{ours}, for Standard SVD, $\mathbf{L}=\mathbf{I}$)}
\State $\mathbf{C}_\ell \gets \mathbf{L}_{G,\ell}^{-\top}\mathbf{U}_{r_\ell^\star}\mathbf{\Sigma}_{r_\ell^\star}^{1/2}$
\State $\mathbf{B}_\ell \gets \mathbf{\Sigma}_{r_\ell^\star}^{1/2}\mathbf{V}_{r_\ell^\star}^\top \mathbf{L}_{A,\ell}^{-1}$
\State replace $\mathbf{W}_\ell$ in $M'$ with $\mathbf{C}_\ell\mathbf{B}_\ell$
\EndFor
\State \Return $M'$
\EndProcedure
\end{algorithmic}
\end{algorithm}

%% file: data/algorithms/kfac_algorithm.tex
\begin{algorithm}[h!]
\caption{\Call{KFAC\_SVD}{$M,\mathcal{D}_{cal},\ell$}: Estimate Factors and Compute Whitened SVD}
\label{alg:kfac_svd}
\begin{algorithmic}[1]
\State \textbf{Input:} model $M$; data $\mathcal{D}_{cal}$; target layer $\ell$ with weights $\mathbf{W}_\ell$; damping $\alpha_A,\alpha_G$
\State \textbf{Output:} decomposition cache $\mathcal{D}_\ell = \{\mathbf{U}_\ell,\mathbf{\Sigma}_\ell,\mathbf{V}_\ell,\mathbf{L}_{A,\ell},\mathbf{L}_{G,\ell}\}$
\Statex \textit{\textbf{\# (I) Streaming Curvature Estimation}}
\State initialize $\mathbf{S}_A\gets \mathbf{0}, \mathbf{S}_G\gets \mathbf{0}, N\gets 0$
\For{sequence $x \sim \mathcal{D}_{cal}$ of length $T$}
\State Forward/Backward pass of the next-token loss: inputs $X_\ell$ (rows $\bm{x}_t$) and output gradients $\Delta_\ell$ (rows $\bm{\delta}_t$)
\State $\mathbf{S}_A \gets \mathbf{S}_A + X_\ell^\top X_\ell$; \quad $\mathbf{S}_G \gets \mathbf{S}_G + \Delta_\ell^\top \Delta_\ell$
\State $N \gets N + T$ \Comment{Token-wise accumulation}
\EndFor
\State $\widehat{\mathbf{A}}_\ell \gets \frac{1}{N}\mathbf{S}_A$; \quad $\widehat{\mathbf{G}}_\ell \gets \frac{1}{N}\mathbf{S}_G$
\State $\mathbf{A}_\ell \gets (1-\alpha_A)\widehat{\mathbf{A}}_\ell + \alpha_A\,\overline{\operatorname{diag}}(\widehat{\mathbf{A}}_\ell)\,\mathbf{I}$; \quad $\mathbf{G}_\ell \gets (1-\alpha_G)\widehat{\mathbf{G}}_\ell + \alpha_G\,\overline{\operatorname{diag}}(\widehat{\mathbf{G}}_\ell)\,\mathbf{I}$ \Comment{Shrinkage}
\Statex \textit{\textbf{\# (II) Whitening and Decomposition}}
\State Compute Cholesky: $\mathbf{A}_\ell=\mathbf{L}_{A,\ell}\mathbf{L}_{A,\ell}^\top$, $\mathbf{G}_\ell=\mathbf{L}_{G,\ell}\mathbf{L}_{G,\ell}^\top$
\State $\widetilde{\mathbf{W}}_\ell \gets \mathbf{L}_{G,\ell}^{\top}\,\mathbf{W}_\ell\,\mathbf{L}_{A,\ell}$ \Comment{Project to whitened basis}
\State $(\mathbf{U}_\ell,\mathbf{\Sigma}_\ell,\mathbf{V}_\ell) \gets \operatorname{SVD}(\widetilde{\mathbf{W}}_\ell)$
\State Rank-$r$ factors: $\widehat{\mathbf{W}}_{\ell,r} = \big(\mathbf{L}_{G,\ell}^{-\top}\mathbf{U}_{\ell,r}\mathbf{\Sigma}_{\ell,r}^{1/2}\big)\big(\mathbf{\Sigma}_{\ell,r}^{1/2}\mathbf{V}_{\ell,r}^{\top}\mathbf{L}_{A,\ell}^{-1}\big)$ \Comment{Un-whiten}
\State \Return $\mathcal{D}_\ell \gets \{\mathbf{U}_\ell,\mathbf{\Sigma}_\ell,\mathbf{V}_\ell,\mathbf{L}_{A,\ell},\mathbf{L}_{G,\ell}\}$
\end{algorithmic}
\end{algorithm}

%% file: data/algorithms/search_algorithm.tex
\begin{algorithm}[h!]
\caption{\Call{LEMS\_Search}{...}: Bilevel Optimization for Rank Allocation}
\label{alg:rank_allocation}
\begin{algorithmic}[1]
\State \textbf{Input:} caches $\{\mathcal{D}_\ell\}$; budget $P_{\text{target}}$; calibration $\mathcal{C}$; search trials $N_{\text{trials}}$
\State \textbf{Output:} optimal ranks $\{r_\ell^\star\}$
\Statex \textit{\textbf{\# (I) Scale Calibration (Scale $\alpha_\ell$)}}
\State Precompute uncompressed outputs $Y_{\text{orig}}$ on $\mathcal{C}$
\For{each layer $\ell \in \{1 \dots L\}$}
\State Measure KL-div $k_\ell$ at reference rank $r_{\text{ref}}$ (e.g. 10\% param)
\State $\alpha_\ell \gets k_\ell / (S_\ell(r_{\text{ref}}) + \epsilon)$ \Comment{Eq. (4)}
\EndFor
\Statex \textit{\textbf{\# (II) Bilevel Search (Bias $\kappa, \gamma$)}}
\State Initialize TPE sampler $\mathcal{S}$; \quad $\mathcal{H}^* \gets \infty$
\For{$k = 1$ to $N_{\text{trials}}$}
\State Sample bias params $\kappa_k, \gamma_k \gets \mathcal{S}.\text{suggest}(\dots)$
\Statex \hspace{1.0em} \textit{\textbf{\# (A) Construct Calibrated Proxy}}
\State $H_{\text{norm}} \gets \sum_{j=1}^B j^{-\gamma_k}$
\For{each layer $\ell$}
\State $\beta(d_\ell) \gets \kappa_k \cdot (\sum_{j=d_\ell}^B j^{-\gamma_k}) / H_{\text{norm}}$ \Comment{Eq. (6)}
\For{each candidate $r \in \mathcal{R}_\ell$}
\State $\text{Cost}_{\ell,r} \gets \underbrace{S_\ell(r)}_{\text{Shape}} \cdot \underbrace{\alpha_\ell}_{\text{Scale}} \cdot \underbrace{(1 + \beta(d_\ell))}_{\text{Bias}}$
\EndFor
\EndFor
\Statex \hspace{1.0em} \textit{\textbf{\# (B) Inner Loop: Solve ILP}}
\State $\{r_\ell^{(k)}\} \gets \Call{SolveILP}{\{\text{Cost}_{\ell,r}\}, P_{\text{target}}}$
\Statex \hspace{1.0em} \textit{\textbf{\# (C) Outer Loop: Validate}}
\State Apply ranks $\{r_\ell^{(k)}\}$ to model $M_{\text{tmp}}$
\State $\mathcal{L}_{\text{val}} \gets \Call{EvaluateKL}{M_{\text{tmp}}, Y_{\text{orig}}, \mathcal{C}}$
\State $\mathcal{S}.\text{observe}(\kappa_k, \gamma_k, \mathcal{L}_{\text{val}})$
\If{$\mathcal{L}_{\text{val}} < \mathcal{H}^*$}
\State $\mathcal{H}^* \gets \mathcal{L}_{\text{val}}$; \quad $\{r_\ell^\star\} \gets \{r_\ell^{(k)}\}$
\EndIf
\EndFor
\State \Return $\{r_\ell^\star\}$
\end{algorithmic}
\end{algorithm}

%% file: data/tables/supp_extended_main_results_norecalib_cr.tex
\begin{table}[h!]
\centering %
\caption{Extended format zero-shot performance and perplexity of LLMs (\gls{mistral}-7B, \gls{llama3}-8B, \gls{qwen3}-8B) with a \textbf{0.9 parameter retention} compressed with a range of \textbf{SVD baselines} and \gls{ours}. \textbf{Bold}/\underline{Underline} denote best/second-best compressed results.}

    \resizebox{0.95\textwidth}{!}{%
\begin{tabular}{l|c|cccccccccc}
\toprule
\midrule
 &\multicolumn{1}{c|}{Method} & \glslink{wiki}{Wiki}\glsunset{wiki} $\downarrow$ & \gls{c4} $\downarrow$ & \gls{ptb} $\downarrow$ & Openb. $\uparrow$ & ARC\_e $\uparrow$ & ARC\_c $\uparrow$ & WinoG. $\uparrow$ & HellaS. $\uparrow$ & PIQA $\uparrow$ & \textbf{Avg. $\uparrow$} \\ \midrule
\multirow{9}{*}{\rotatebox[origin=c]{90}{\gls{mistral}-7B}} & \multicolumn{1}{c|}{\color[HTML]{9B9B9B}Original} & {\color[HTML]{9B9B9B}5.25} & {\color[HTML]{9B9B9B}8.10} & {\color[HTML]{9B9B9B}27.72} & {\color[HTML]{9B9B9B}0.35} & {\color[HTML]{9B9B9B}0.80} & {\color[HTML]{9B9B9B}0.51} & {\color[HTML]{9B9B9B}0.75} & {\color[HTML]{9B9B9B}0.62} & {\color[HTML]{9B9B9B}0.81} & {\color[HTML]{9B9B9B}63.96} \\
& \multicolumn{1}{c|}{\gls{fwsvd}} & 9.47 & 13.75 & 58.35 & 0.31 & 0.74 & 0.40 & 0.69 & 0.51 & 0.76 & 56.93 \\
& \multicolumn{1}{c|}{\gls{asvd}} & 9.14 & 13.41 & 49.13 & \underline{0.33} & 0.73 & 0.41 & 0.69 & \underline{0.52} & \underline{0.77} & \underline{57.30} \\
& \multicolumn{1}{c|}{\gls{svdllm}} & 6.46 & 15.07 & 66.12 & 0.30 & \underline{0.75} & \underline{0.42} & \underline{0.71} & 0.49 & 0.75 & 56.81 \\
& \multicolumn{1}{c|}{\gls{svdllmv2}} & 6.46 & 15.07 & 66.12 & 0.30 & \underline{0.75} & \underline{0.42} & \underline{0.71} & 0.49 & 0.75 & 56.78 \\
& \multicolumn{1}{c|}{\acrshort{dobisvd}} & 7.11 & 15.19 & 63.58 & 0.29 & 0.71 & 0.38 & \underline{0.71} & 0.49 & 0.74 & 55.17 \\
& \multicolumn{1}{c|}{\acrshort{gfwsvd}} & 31.66 & 65.09 & 305.72 & 0.18 & 0.46 & 0.24 & 0.54 & 0.38 & 0.63 & 40.73 \\
& \multicolumn{1}{c|}{\midrowcolor{m} \textbf{\gls{ours}}} & \underline{6.17} & \underline{12.79} & \underline{44.91} & 0.30 & \underline{0.75} & \underline{0.42} & 0.70 & 0.49 & 0.75 & 56.95 \\
& \multicolumn{1}{c|}{\midrowcolor{m} \textbf{$+$\gls{oursearch}}} & \textbf{5.35} & \textbf{8.44} & \textbf{29.28} & \textbf{0.34} & \textbf{0.79} & \textbf{0.49} & \textbf{0.74} & \textbf{0.60} & \textbf{0.80} & \textbf{62.71} \\
\midrule
\multirow{9}{*}{\rotatebox[origin=c]{90}{\gls{llama3}-8B}} & \multicolumn{1}{c|}{\color[HTML]{9B9B9B}Original} & {\color[HTML]{9B9B9B}6.14} & {\color[HTML]{9B9B9B}9.47} & {\color[HTML]{9B9B9B}9.90} & {\color[HTML]{9B9B9B}0.34} & {\color[HTML]{9B9B9B}0.81} & {\color[HTML]{9B9B9B}0.52} & {\color[HTML]{9B9B9B}0.74} & {\color[HTML]{9B9B9B}0.61} & {\color[HTML]{9B9B9B}0.79} & {\color[HTML]{9B9B9B}63.33} \\
& \multicolumn{1}{c|}{\gls{fwsvd}} & 42.11 & 59.96 & 69.83 & 0.25 & 0.62 & 0.31 & 0.64 & 0.41 & 0.70 & 48.88 \\
& \multicolumn{1}{c|}{\gls{asvd}} & 65.09 & 77.60 & 207.67 & 0.27 & 0.57 & 0.31 & 0.63 & 0.41 & 0.68 & 47.89 \\
& \multicolumn{1}{c|}{\gls{svdllm}} & 10.14 & 39.63 & 49.91 & 0.26 & 0.69 & 0.37 & 0.66 & 0.43 & 0.72 & 52.27 \\
& \multicolumn{1}{c|}{\gls{svdllmv2}} & 10.18 & 39.71 & 50.40 & 0.26 & 0.69 & 0.37 & 0.66 & 0.43 & 0.72 & 52.17 \\
& \multicolumn{1}{c|}{\acrshort{dobisvd}} & 11.22 & 34.77 & 42.36 & 0.27 & 0.68 & 0.37 & 0.68 & 0.46 & 0.72 & 52.98 \\
& \multicolumn{1}{c|}{\acrshort{gfwsvd}} & 2569.75 & 3827.63 & 8491.96 & 0.14 & 0.30 & 0.20 & 0.53 & 0.28 & 0.55 & 33.40 \\
& \multicolumn{1}{c|}{\midrowcolor{m} \textbf{\gls{ours}}} & \underline{8.49} & \underline{22.98} & \underline{26.15} & \underline{0.29} & \underline{0.74} & \underline{0.40} & \underline{0.70} & \underline{0.48} & \underline{0.74} & \underline{55.84} \\
& \multicolumn{1}{c|}{\midrowcolor{m} \textbf{$+$\gls{oursearch}}} & \textbf{6.51} & \textbf{10.90} & \textbf{10.94} & \textbf{0.34} & \textbf{0.80} & \textbf{0.49} & \textbf{0.72} & \textbf{0.57} & \textbf{0.78} & \textbf{61.83} \\
\midrule
\multirow{9}{*}{\rotatebox[origin=c]{90}{\gls{qwen3}-8B}} & \multicolumn{1}{c|}{\color[HTML]{9B9B9B}Original} & {\color[HTML]{9B9B9B}9.71} & {\color[HTML]{9B9B9B}15.52} & {\color[HTML]{9B9B9B}15.43} & {\color[HTML]{9B9B9B}0.31} & {\color[HTML]{9B9B9B}0.84} & {\color[HTML]{9B9B9B}0.56} & {\color[HTML]{9B9B9B}0.68} & {\color[HTML]{9B9B9B}0.57} & {\color[HTML]{9B9B9B}0.77} & {\color[HTML]{9B9B9B}62.04} \\
& \multicolumn{1}{c|}{\gls{fwsvd}} & 16.95 & 24.75 & 31.54 & 0.29 & 0.73 & 0.40 & 0.66 & 0.43 & 0.71 & 53.82 \\
& \multicolumn{1}{c|}{\gls{asvd}} & 20.36 & 29.98 & 36.37 & 0.27 & 0.61 & 0.37 & 0.63 & 0.48 & 0.71 & 51.19 \\
& \multicolumn{1}{c|}{\gls{svdllm}} & 12.52 & 27.40 & 36.66 & \underline{0.31} & 0.71 & 0.44 & 0.66 & \underline{0.50} & \underline{0.73} & 55.96 \\
& \multicolumn{1}{c|}{\gls{svdllmv2}} & 12.52 & 27.45 & 36.80 & \underline{0.31} & 0.71 & 0.44 & 0.66 & \underline{0.50} & \underline{0.73} & 55.80 \\
& \multicolumn{1}{c|}{\acrshort{dobisvd}} & 13.46 & 28.88 & 35.81 & 0.29 & 0.69 & 0.45 & \underline{0.68} & \underline{0.50} & 0.71 & 55.48 \\
& \multicolumn{1}{c|}{\acrshort{gfwsvd}} & 171.49 & 350.51 & 765.57 & 0.27 & 0.50 & 0.34 & 0.58 & 0.43 & 0.67 & 46.66 \\
& \multicolumn{1}{c|}{\midrowcolor{m} \textbf{\gls{ours}}} & \underline{11.27} & \underline{23.44} & \underline{25.69} & \underline{0.31} & \underline{0.76} & \underline{0.47} & 0.65 & 0.49 & 0.72 & \underline{56.73} \\
& \multicolumn{1}{c|}{\midrowcolor{m} \textbf{$+$\gls{oursearch}}} & \textbf{9.81} & \textbf{15.98} & \textbf{15.70} & \textbf{0.34} & \textbf{0.83} & \textbf{0.56} & \textbf{0.70} & \textbf{0.55} & \textbf{0.77} & \textbf{62.60} \\
\midrule
\bottomrule
\end{tabular}
}
\label{supp:tab:llm_main_extended_09}
\end{table}
\begin{table}[]
\centering
\captionof{table}{Extended format zero-shot performance and perplexity of LLMs (\gls{mistral}-7B, \gls{llama3}-8B, \gls{qwen3}-8B) with a \textbf{0.7 parameter retention} compressed with a range of \textbf{SVD baselines} and \gls{ours}. \textbf{Bold}/\underline{Underline} denote best/second-best compressed results. %
}

\resizebox{0.95\textwidth}{!}{%
\begin{tabular}{l|c|cccccccccc}
\toprule
\midrule
 &\multicolumn{1}{c|}{Method} & \glslink{wiki}{Wiki}\glsunset{wiki} $\downarrow$ & \gls{c4} $\downarrow$ & \gls{ptb} $\downarrow$ & Openb. $\uparrow$ & ARC\_e $\uparrow$ & ARC\_c $\uparrow$ & WinoG. $\uparrow$ & HellaS. $\uparrow$ & PIQA $\uparrow$ & \textbf{Avg. $\uparrow$} \\ \midrule
\multirow{9}{*}{\rotatebox[origin=c]{90}{\gls{mistral}-7B}} & \multicolumn{1}{c|}{\color[HTML]{9B9B9B}Original} & {\color[HTML]{9B9B9B}5.25} & {\color[HTML]{9B9B9B}8.10} & {\color[HTML]{9B9B9B}27.72} & {\color[HTML]{9B9B9B}0.35} & {\color[HTML]{9B9B9B}0.80} & {\color[HTML]{9B9B9B}0.51} & {\color[HTML]{9B9B9B}0.75} & {\color[HTML]{9B9B9B}0.62} & {\color[HTML]{9B9B9B}0.81} & {\color[HTML]{9B9B9B}63.96} \\
& \multicolumn{1}{c|}{\gls{fwsvd}} & 34.84 & 50.30 & 215.10 & 0.21 & 0.59 & 0.25 & 0.60 & 0.34 & \underline{0.66} & 44.24 \\
& \multicolumn{1}{c|}{\gls{asvd}} & 28.16 & 42.94 & 163.64 & \underline{0.24} & 0.57 & \underline{0.29} & 0.62 & \underline{0.38} & \underline{0.66} & 46.09 \\
& \multicolumn{1}{c|}{\gls{svdllm}} & 10.96 & 42.60 & 169.49 & \underline{0.24} & 0.56 & 0.25 & 0.62 & 0.35 & 0.64 & 44.39 \\
& \multicolumn{1}{c|}{\gls{svdllmv2}} & 10.96 & 42.60 & 169.49 & \underline{0.24} & 0.56 & 0.25 & 0.62 & 0.35 & 0.64 & 44.35 \\
& \multicolumn{1}{c|}{\acrshort{dobisvd}} & 12.37 & 39.71 & 170.82 & 0.22 & 0.50 & 0.26 & 0.60 & 0.35 & 0.65 & 42.94 \\
& \multicolumn{1}{c|}{\acrshort{gfwsvd}} & 2549.75 & 3404.36 & 5355.81 & 0.14 & 0.26 & 0.24 & 0.50 & 0.26 & 0.54 & 32.29 \\
& \multicolumn{1}{c|}{\midrowcolor{m} \textbf{\gls{ours}}} & \underline{9.18} & \underline{32.73} & \underline{98.86} & 0.23 & \underline{0.61} & 0.28 & \underline{0.63} & 0.36 & \underline{0.66} & \underline{46.26} \\
& \multicolumn{1}{c|}{\midrowcolor{m} \textbf{$+$\gls{oursearch}}} & \textbf{7.28} & \textbf{17.42} & \textbf{54.28} & \textbf{0.28} & \textbf{0.69} & \textbf{0.33} & \textbf{0.66} & \textbf{0.41} & \textbf{0.72} & \textbf{51.68} \\
\midrule
\multirow{9}{*}{\rotatebox[origin=c]{90}{\gls{llama3}-8B}} & \multicolumn{1}{c|}{\color[HTML]{9B9B9B}Original} & {\color[HTML]{9B9B9B}6.14} & {\color[HTML]{9B9B9B}9.47} & {\color[HTML]{9B9B9B}9.90} & {\color[HTML]{9B9B9B}0.34} & {\color[HTML]{9B9B9B}0.81} & {\color[HTML]{9B9B9B}0.52} & {\color[HTML]{9B9B9B}0.74} & {\color[HTML]{9B9B9B}0.61} & {\color[HTML]{9B9B9B}0.79} & {\color[HTML]{9B9B9B}63.33} \\
& \multicolumn{1}{c|}{\gls{fwsvd}} & 716.39 & 837.54 & 1939.75 & 0.13 & 0.33 & 0.21 & 0.52 & 0.27 & 0.55 & 33.40 \\
& \multicolumn{1}{c|}{\gls{asvd}} & 10989.41 & 9399.74 & 36032.89 & 0.14 & 0.30 & 0.20 & 0.50 & 0.27 & 0.56 & 33.03 \\
& \multicolumn{1}{c|}{\gls{svdllm}} & 34.64 & 376.04 & 600.91 & 0.20 & 0.40 & 0.21 & 0.58 & 0.29 & 0.58 & 37.69 \\
& \multicolumn{1}{c|}{\gls{svdllmv2}} & 34.98 & 378.99 & 615.16 & 0.20 & 0.40 & 0.21 & 0.58 & 0.29 & 0.58 & 37.70 \\
& \multicolumn{1}{c|}{\acrshort{dobisvd}} & 31.54 & 500.12 & 223.67 & 0.19 & 0.39 & 0.23 & 0.59 & 0.31 & 0.60 & 38.48 \\
& \multicolumn{1}{c|}{\acrshort{gfwsvd}} & 41798.89 & 53670.83 & 72222.00 & 0.14 & 0.28 & 0.21 & 0.49 & 0.26 & 0.53 & 31.77 \\
& \multicolumn{1}{c|}{\midrowcolor{m} \textbf{\gls{ours}}} & \underline{16.23} & \underline{107.32} & \underline{180.42} & \underline{0.21} & \underline{0.56} & \underline{0.27} & \underline{0.60} & \underline{0.34} & \underline{0.63} & \underline{43.69} \\
& \multicolumn{1}{c|}{\midrowcolor{m} \textbf{$+$\gls{oursearch}}} & \textbf{10.71} & \textbf{33.70} & \textbf{56.11} & \textbf{0.26} & \textbf{0.67} & \textbf{0.34} & \textbf{0.64} & \textbf{0.39} & \textbf{0.70} & \textbf{49.88} \\
\midrule
\multirow{9}{*}{\rotatebox[origin=c]{90}{\gls{qwen3}-8B}} & \multicolumn{1}{c|}{\color[HTML]{9B9B9B}Original} & {\color[HTML]{9B9B9B}9.71} & {\color[HTML]{9B9B9B}15.52} & {\color[HTML]{9B9B9B}15.43} & {\color[HTML]{9B9B9B}0.31} & {\color[HTML]{9B9B9B}0.84} & {\color[HTML]{9B9B9B}0.56} & {\color[HTML]{9B9B9B}0.68} & {\color[HTML]{9B9B9B}0.57} & {\color[HTML]{9B9B9B}0.77} & {\color[HTML]{9B9B9B}62.04} \\
& \multicolumn{1}{c|}{\gls{fwsvd}} & 41.37 & 68.75 & 90.72 & 0.20 & 0.58 & 0.27 & 0.56 & 0.33 & 0.63 & 42.68 \\
& \multicolumn{1}{c|}{\gls{asvd}} & 70.66 & 96.95 & 131.49 & 0.24 & 0.49 & 0.29 & 0.55 & 0.38 & \underline{0.65} & 43.10 \\
& \multicolumn{1}{c|}{\gls{svdllm}} & 17.11 & 63.09 & 136.19 & 0.24 & 0.60 & 0.30 & \underline{0.59} & 0.38 & \underline{0.65} & 45.76 \\
& \multicolumn{1}{c|}{\gls{svdllmv2}} & 17.15 & 63.33 & 136.72 & 0.23 & 0.60 & 0.30 & 0.58 & 0.38 & \underline{0.65} & 45.62 \\
& \multicolumn{1}{c|}{\acrshort{dobisvd}} & 18.54 & 63.83 & 94.34 & \underline{0.25} & 0.55 & 0.32 & \textbf{0.63} & \underline{0.39} & 0.64 & 46.24 \\
& \multicolumn{1}{c|}{\acrshort{gfwsvd}} & 21684.98 & 27627.38 & 50814.52 & 0.17 & 0.31 & 0.22 & 0.51 & 0.28 & 0.56 & 34.06 \\
& \multicolumn{1}{c|}{\midrowcolor{m} \textbf{\gls{ours}}} & \underline{13.97} & \underline{45.09} & \underline{72.33} & \textbf{0.28} & \underline{0.61} & \underline{0.35} & \textbf{0.63} & \underline{0.39} & 0.63 & \underline{48.24} \\
& \multicolumn{1}{c|}{\midrowcolor{m} \textbf{$+$\gls{oursearch}}} & \textbf{11.67} & \textbf{28.05} & \textbf{37.16} & \textbf{0.28} & \textbf{0.74} & \textbf{0.39} & \textbf{0.63} & \textbf{0.43} & \textbf{0.72} & \textbf{53.30} \\
\midrule
\bottomrule
\end{tabular}
    }\label{supp:tab:llm_main_extended_07}
\end{table}

%% file: data/tables/supp_extended_search_results_norecalib_cr.tex
\begin{table}[h!]
    \centering
    \caption{Extended format zero-shot performance and perplexity of LLMs (\gls{mistral}-7B, \gls{llama3}-8B, \gls{qwen3}-8B) compressed with a range of \textbf{different searches} with a \textbf{0.8 parameter retention}. \textbf{Bold}/\underline{Underline} denote best/second-best compressed results.}
\resizebox{0.95\textwidth}{!}{%
    \begin{tabular}{l|c|cccccccccc}
\toprule
\midrule
 &\multicolumn{1}{c|}{Search Method} & \glslink{wiki}{Wiki}\glsunset{wiki} $\downarrow$ & \gls{c4} $\downarrow$ & \gls{ptb} $\downarrow$ & Openb. $\uparrow$ & ARC\_e $\uparrow$ & ARC\_c $\uparrow$ & WinoG. $\uparrow$ & HellaS. $\uparrow$ & PIQA $\uparrow$ & \textbf{Avg.$\uparrow$} \\ \midrule
\multirow{7}{*}{\rotatebox[origin=c]{90}{\gls{mistral}-7B}} & \multicolumn{1}{c|}{uniform} & 7.08 & 18.58 & 57.22 & 0.27 & 0.70 & \underline{0.37} & 0.68 & \underline{0.43} & 0.71 & 52.67 \\
& \multicolumn{1}{c|}{\gls{asvd}} & \underline{6.98} & 19.66 & 64.84 & \underline{0.28} & 0.70 & 0.34 & 0.65 & 0.40 & 0.71 & 51.41 \\
& \multicolumn{1}{c|}{\gls{svdllmv2}} & 7.06 & 18.58 & 56.77 & 0.27 & \underline{0.71} & 0.36 & 0.68 & \underline{0.43} & 0.71 & 52.77 \\
& \multicolumn{1}{c|}{\gls{mrcs}} & 7.06 & 18.32 & 58.12 & 0.27 & \underline{0.71} & \underline{0.37} & 0.68 & \underline{0.43} & 0.71 & 52.96 \\
& \multicolumn{1}{c|}{\gls{ars}} & 7.32 & \underline{11.76} & \underline{44.65} & 0.27 & 0.68 & 0.36 & 0.68 & \textbf{0.50} & \underline{0.74} & \underline{53.68} \\
& \multicolumn{1}{c|}{\gls{atp}} & 7.06 & 18.15 & 58.35 & 0.27 & \underline{0.71} & \underline{0.37} & \underline{0.69} & \underline{0.43} & 0.70 & 52.75 \\
& \multicolumn{1}{c|}{\midrowcolor{m} \textbf{\gls{oursearch}}} & \textbf{5.92} & \textbf{11.11} & \textbf{34.91} & \textbf{0.33} & \textbf{0.76} & \textbf{0.41} & \textbf{0.70} & \textbf{0.50} & \textbf{0.76} & \textbf{57.61} \\
\midrule
\multirow{7}{*}{\rotatebox[origin=c]{90}{\gls{llama3}-8B}} & \multicolumn{1}{c|}{uniform} & 10.61 & 41.94 & 60.91 & 0.25 & 0.67 & 0.35 & 0.67 & 0.42 & 0.69 & 50.84 \\
& \multicolumn{1}{c|}{\gls{asvd}} & 12.02 & 59.27 & 65.86 & 0.25 & 0.63 & 0.29 & 0.63 & 0.36 & 0.69 & 47.28 \\
& \multicolumn{1}{c|}{\gls{svdllmv2}} & 10.61 & 41.54 & 60.20 & 0.25 & 0.67 & 0.35 & \underline{0.68} & 0.42 & 0.69 & 51.05 \\
& \multicolumn{1}{c|}{\gls{mrcs}} & 10.54 & 41.29 & 59.27 & \underline{0.27} & 0.67 & 0.35 & 0.66 & 0.40 & 0.70 & 51.01 \\
& \multicolumn{1}{c|}{\gls{ars}} & 11.33 & \underline{19.62} & \textbf{18.43} & \underline{0.27} & 0.65 & 0.31 & 0.61 & 0.42 & \underline{0.73} & 49.58 \\
& \multicolumn{1}{c|}{\gls{atp}} & \underline{10.28} & 35.11 & 60.20 & 0.26 & \underline{0.68} & \underline{0.37} & \textbf{0.69} & \underline{0.43} & 0.71 & \underline{52.10} \\
& \multicolumn{1}{c|}{\midrowcolor{m} \textbf{\gls{oursearch}}} & \textbf{8.04} & \textbf{16.91} & \underline{20.56} & \textbf{0.30} & \textbf{0.76} & \textbf{0.41} & \textbf{0.69} & \textbf{0.48} & \textbf{0.76} & \textbf{56.77} \\
\midrule
\multirow{7}{*}{\rotatebox[origin=c]{90}{\gls{qwen3}-8B}} & \multicolumn{1}{c|}{uniform} & 12.25 & 30.63 & 41.70 & \textbf{0.32} & 0.73 & 0.43 & \underline{0.65} & 0.44 & 0.68 & 54.18 \\
& \multicolumn{1}{c|}{\gls{asvd}} & 13.28 & 36.16 & 46.52 & 0.28 & 0.68 & 0.37 & 0.63 & 0.41 & 0.68 & 50.74 \\
& \multicolumn{1}{c|}{\gls{svdllmv2}} & 12.28 & 30.51 & 41.46 & \textbf{0.32} & 0.74 & 0.44 & \underline{0.65} & 0.44 & 0.69 & 54.39 \\
& \multicolumn{1}{c|}{\gls{mrcs}} & 14.24 & 34.23 & 47.34 & \textbf{0.32} & 0.68 & 0.42 & 0.64 & \underline{0.45} & 0.68 & 53.16 \\
& \multicolumn{1}{c|}{\gls{ars}} & \underline{11.05} & \underline{20.09} & \underline{19.89} & 0.30 & \underline{0.78} & \underline{0.46} & 0.63 & \underline{0.45} & \underline{0.75} & \underline{56.08} \\
& \multicolumn{1}{c|}{\gls{atp}} & 12.25 & 30.63 & 41.70 & \textbf{0.32} & 0.73 & 0.43 & \underline{0.65} & 0.44 & 0.68 & 54.18 \\
& \multicolumn{1}{c|}{\midrowcolor{m} \textbf{\gls{oursearch}}} & \textbf{10.28} & \textbf{17.93} & \textbf{18.22} & \underline{0.31} & \textbf{0.80} & \textbf{0.50} & \textbf{0.68} & \textbf{0.51} & \textbf{0.76} & \textbf{59.03} \\
\midrule
\bottomrule
\end{tabular}
}
    \label{supp:tab:extended_search_results_norecalib_08}
\end{table}

\begin{table}[]
    \centering
    \caption{Extended format zero-shot performance and perplexity of LLMs (\gls{mistral}-7B, \gls{llama3}-8B, \gls{qwen3}-8B) compressed with a range of \textbf{different searches} with a \textbf{0.6 parameter retention}. \textbf{Bold}/\underline{Underline} denote best/second-best compressed results.}
\resizebox{0.95\textwidth}{!}{%
    \begin{tabular}{l|c|cccccccccc}
\toprule
\midrule
 &\multicolumn{1}{c|}{Search Method} & \glslink{wiki}{Wiki}\glsunset{wiki} $\downarrow$ & \gls{c4} $\downarrow$ & \gls{ptb} $\downarrow$ & Openb. $\uparrow$ & ARC\_e $\uparrow$ & ARC\_c $\uparrow$ & WinoG. $\uparrow$ & HellaS. $\uparrow$ & PIQA $\uparrow$ & \textbf{Avg.$\uparrow$} \\ \midrule
\multirow{7}{*}{\rotatebox[origin=c]{90}{\gls{mistral}-7B}} & \multicolumn{1}{c|}{uniform} & 14.47 & 67.16 & 181.84 & 0.19 & 0.47 & 0.23 & 0.57 & 0.31 & 0.59 & 39.22 \\
& \multicolumn{1}{c|}{\gls{asvd}} & 16.33 & 91.43 & 268.74 & 0.17 & 0.40 & 0.20 & 0.55 & 0.29 & 0.57 & 36.35 \\
& \multicolumn{1}{c|}{\gls{svdllmv2}} & 14.50 & 67.16 & 183.27 & \underline{0.20} & 0.47 & 0.23 & 0.57 & 0.31 & 0.59 & 39.37 \\
& \multicolumn{1}{c|}{\gls{mrcs}} & \underline{14.41} & 70.93 & 222.79 & \underline{0.20} & 0.48 & 0.23 & 0.57 & 0.31 & 0.59 & 39.58 \\
& \multicolumn{1}{c|}{\gls{ars}} & 18.47 & \textbf{33.05} & \underline{176.24} & 0.18 & \underline{0.49} & 0.23 & 0.55 & \underline{0.34} & \underline{0.64} & 40.66 \\
& \multicolumn{1}{c|}{\gls{atp}} & 15.73 & 59.73 & 188.35 & \underline{0.20} & 0.48 & \textbf{0.25} & \underline{0.60} & 0.33 & 0.61 & \underline{41.13} \\
& \multicolumn{1}{c|}{\midrowcolor{m} \textbf{\gls{oursearch}}} & \textbf{10.32} & \underline{35.18} & \textbf{112.47} & \textbf{0.21} & \textbf{0.57} & \underline{0.24} & \textbf{0.61} & \textbf{0.35} & \textbf{0.65} & \textbf{43.84} \\
\midrule
    \multirow{7}{*}{\rotatebox[origin=c]{90}{\gls{llama3}-8B}} & \multicolumn{1}{c|}{uniform} & 33.97 & 325.43 & 536.55 & 0.19 & 0.38 & 0.19 & 0.55 & 0.29 & 0.57 & 36.24 \\
& \multicolumn{1}{c|}{\gls{asvd}} & 47.99 & 446.55 & 870.90 & 0.16 & 0.36 & 0.19 & 0.53 & 0.28 & 0.56 & 34.73 \\
& \multicolumn{1}{c|}{\gls{svdllmv2}} & 33.25 & 317.89 & 532.37 & 0.18 & 0.38 & 0.19 & 0.55 & 0.29 & 0.58 & 36.32 \\
& \multicolumn{1}{c|}{\gls{mrcs}} & 36.58 & 315.42 & 520.04 & 0.18 & 0.37 & 0.19 & 0.55 & 0.28 & 0.57 & 35.76 \\
& \multicolumn{1}{c|}{\gls{ars}} & 23.62 & \textbf{44.39} & \textbf{64.08} & \underline{0.21} & \underline{0.55} & 0.23 & 0.53 & \underline{0.34} & \textbf{0.66} & \underline{42.05} \\
& \multicolumn{1}{c|}{\gls{atp}} & \underline{21.09} & 129.95 & 300.98 & 0.20 & 0.49 & \underline{0.24} & \textbf{0.62} & 0.33 & 0.63 & 41.96 \\
& \multicolumn{1}{c|}{\midrowcolor{m} \textbf{\gls{oursearch}}} & \textbf{16.08} & \underline{63.83} & \underline{140.51} & \textbf{0.22} & \textbf{0.56} & \textbf{0.26} & \underline{0.61} & \textbf{0.35} & \underline{0.65} & \textbf{44.19} \\
\midrule
\multirow{7}{*}{\rotatebox[origin=c]{90}{\gls{qwen3}-8B}} & \multicolumn{1}{c|}{uniform} & \underline{18.00} & 77.30 & 151.34 & \underline{0.22} & 0.46 & 0.25 & \textbf{0.60} & \underline{0.34} & 0.60 & 41.16 \\
& \multicolumn{1}{c|}{\gls{asvd}}     & 23.25 & 97.33 & 184.70 & 0.20 & 0.45 & 0.23 & 0.58 & 0.32 & 0.61 & 39.88 \\
& \multicolumn{1}{c|}{\gls{svdllmv2}} & 18.04 & 76.40 & 150.16 & \underline{0.22} & 0.45 & \underline{0.26} & \textbf{0.60} & \underline{0.34} & 0.61 & 41.23 \\
& \multicolumn{1}{c|}{\gls{mrcs}} & 27.13 & 109.86 & 224.54 & \underline{0.22} & 0.45 & 0.25 & 0.57 & 0.33 & 0.60 & 40.26 \\
& \multicolumn{1}{c|}{\gls{ars}} & 19.47 & \textbf{53.65} & \textbf{57.89} & 0.21 & \underline{0.56} & \underline{0.26} & 0.55 & 0.33 & \underline{0.64} & \underline{42.60} \\
& \multicolumn{1}{c|}{\gls{atp}} & \underline{18.00} & 77.30 & 151.34 & \underline{0.22} & 0.46 & 0.25 & \textbf{0.60} & \underline{0.34} & 0.60 & 41.16 \\
& \multicolumn{1}{c|}{\midrowcolor{m} \textbf{\gls{oursearch}}} & \textbf{15.07} & \underline{54.39} & \underline{112.91} & \textbf{0.23} & \textbf{0.62} & \textbf{0.30} & \underline{0.59} & \textbf{0.35} & \textbf{0.65} & \textbf{45.60} \\
\midrule
\bottomrule
\end{tabular}
}
    \label{supp:tab:extended_search_results_norecalib_06}
\end{table}

%% file: data/tables/ablation_generalize_results_table_norecalib_cr.tex
\begin{tabular}{cc|cccccccccc}
\toprule
\midrule
&& \multicolumn{10}{c}{\gls{qwen3}} \\
& Method & \multicolumn{2}{c}{1.7B} & \multicolumn{2}{c}{4B} & \multicolumn{2}{c}{8B} & \multicolumn{2}{c}{14B} & \multicolumn{2}{c}{32B} \\\cmidrule{3-12}
&  & PPL $\downarrow$ & Acc $\uparrow$ & PPL $\downarrow$ & Acc $\uparrow$ & PPL $\downarrow$ & Acc $\uparrow$ & PPL $\downarrow$ & Acc $\uparrow$ & PPL $\downarrow$ & Acc $\uparrow$ \\
\midrule
& Baseline & 16.68 & 53.33 & 13.64 & 58.97 & 9.71 & 62.04 & 8.64 & 65.28 & 7.59 & 45.02 \\
\midrule
\multirow{4}{*}{0.9} &  \gls{svdllm} & 30.99 & 44.18 & 17.15 & 49.28 & 12.52 & 55.96 & 11.07 & 58.82 & 9.23 & 43.23 \\
 &  \midrowcolor{m}$+$\textbf{\gls{oursearch}} & \textbf{17.62} & \textbf{51.24} & \textbf{14.02} & \textbf{57.56} & \textbf{9.83} & \textbf{62.36} & \textbf{8.77} & \textbf{65.28} & \textbf{7.64} & \textbf{65.61} \\
\cmidrule{2-12}
 &  \gls{ours} & 26.98 & 44.11 & 15.98 & 50.79 & 11.27 & 56.73 & 9.87 & 60.49 & 8.26 & 43.58 \\
 &  \midrowcolor{m}$+$\textbf{\gls{oursearch}} & \textbf{16.55} & \textbf{51.31} & \textbf{13.97} & \textbf{58.52} & \textbf{9.81} & \textbf{62.60} & \textbf{8.74} & \textbf{65.05} & \textbf{7.64} & \textbf{65.48} \\
\midrule
\multirow{4}{*}{0.8} &  \gls{svdllm} & 36.73 & 38.75 & 19.93 & 45.76 & 14.05 & 52.31 & 12.16 & 56.54 & 10.22 & 57.87 \\
 &  \midrowcolor{m}$+$\textbf{\gls{oursearch}} & \textbf{18.72} & \textbf{46.24} & \textbf{15.13} & \textbf{54.29} & \textbf{10.50} & \textbf{60.04} & \textbf{9.32} & \textbf{62.85} & \textbf{7.90} & \textbf{65.85} \\
\cmidrule{2-12}
 &  \gls{ours} & 33.25 & 39.92 & 17.18 & 48.42 & 12.25 & 54.18 & 10.75 & 57.02 & 8.84 & 42.69 \\
 &  \midrowcolor{m}$+$\textbf{\gls{oursearch}} & \textbf{18.18} & \textbf{45.61} & \textbf{14.47} & \textbf{55.05} & \textbf{10.28} & \textbf{59.03} & \textbf{9.16} & \textbf{62.61} & \textbf{7.90} & \textbf{65.10} \\
\midrule
\multirow{4}{*}{0.7} &  \gls{svdllm} & 57.67 & 35.40 & 28.72 & 38.53 & 17.11 & 45.76 & 14.75 & 52.49 & 11.65 & 54.78 \\
 &  \midrowcolor{m}$+$\textbf{\gls{oursearch}} & \textbf{23.25} & \textbf{42.05} & \textbf{17.11} & \textbf{48.91} & \textbf{12.37} & \textbf{54.75} & \textbf{10.40} & \textbf{58.32} & \textbf{8.34} & \textbf{64.51} \\
\cmidrule{2-12}
 &  \gls{ours} & 37.38 & 36.93 & 19.85 & 43.09 & 13.97 & 48.24 & 12.02 & 53.13 & 9.77 & 55.94 \\
 &  \midrowcolor{m}$+$\textbf{\gls{oursearch}} & \textbf{21.55} & \textbf{40.52} & \textbf{16.17} & \textbf{49.27} & \textbf{11.67} & \textbf{53.30} & \textbf{10.12} & \textbf{58.03} & \textbf{8.32} & \textbf{64.00} \\
\midrule
\multirow{4}{*}{0.6} &  \gls{svdllm} & 124.98 & 33.25 & 59.27 & 33.87 & 28.77 & 36.81 & 19.81 & 43.29 & 15.43 & 48.60 \\
 &  \midrowcolor{m}$+$\textbf{\gls{oursearch}} & \textbf{32.35} & \textbf{37.23} & \textbf{22.45} & \textbf{43.53} & \textbf{17.28} & \textbf{46.46} & \textbf{13.35} & \textbf{51.12} & \textbf{9.36} & \textbf{60.37} \\
\cmidrule{2-12}
 &  \gls{ours} & 51.79 & 34.84 & 28.33 & 38.19 & 18.00 & 41.16 & 14.38 & 46.97 & 11.53 & 52.69 \\
 &  \midrowcolor{m}$+$\textbf{\gls{oursearch}} & \textbf{28.21} & \textbf{37.27} & \textbf{20.20} & \textbf{42.80} & \textbf{15.07} & \textbf{45.60} & \textbf{12.13} & \textbf{50.85} & \textbf{9.27} & \textbf{60.58} \\
\midrule
\multirow{4}{*}{0.5} &  \gls{svdllm} & 227.19 & 32.05 & 159.22 & 32.19 & 65.60 & 33.58 & 31.85 & 36.30 & 25.79 & 39.37 \\
 &  \midrowcolor{m}$+$\textbf{\gls{oursearch}} & \textbf{56.33} & \textbf{35.06} & \textbf{33.57} & \textbf{38.28} & \textbf{31.66} & \textbf{40.70} & \textbf{18.32} & \textbf{42.72} & \textbf{11.88} & \textbf{54.59} \\
\cmidrule{2-12}
 &  \gls{ours} & 101.21 & 33.02 & 51.89 & 34.03 & 27.94 & 35.91 & 19.43 & 39.48 & 15.40 & 45.47 \\
 &  \midrowcolor{m}$+$\textbf{\gls{oursearch}} & \textbf{39.87} & \textbf{34.41} & \textbf{30.57} & \textbf{38.15} & \textbf{22.15} & \textbf{39.63} & \textbf{16.43} & \textbf{42.57} & \textbf{11.58} & \textbf{53.73} \\
\midrule
\bottomrule
\end{tabular}